\documentclass[sigconf]{acmart}
\makeatletter
\def\@ACM@checkaffil{
    \if@ACM@instpresent\else
    \ClassWarningNoLine{\@classname}{No institution present for an affiliation}%
    \fi
    \if@ACM@citypresent\else
    \ClassWarningNoLine{\@classname}{No city present for an affiliation}%
    \fi
    \if@ACM@countrypresent\else
        \ClassWarningNoLine{\@classname}{No country present for an affiliation}%
    \fi
}
\makeatother
\acmConference[]{}{}{}
\usepackage{xspace}
\newcommand{\method}{\texttt{HALO}\xspace}
\usepackage[utf8]{inputenc} 
\usepackage[T1]{fontenc}    
\usepackage{hyperref}       
\usepackage{url}            
\usepackage{booktabs}       
\usepackage{amsfonts}       
\usepackage{nicefrac}       
\usepackage{microtype}      
\usepackage{graphics}
\usepackage{graphicx}
\usepackage{eqnarray, amsmath}
\usepackage{algorithm, algorithmicx}
\usepackage{algpseudocode}
\usepackage{xcolor}
\usepackage{caption}
\usepackage{subcaption}
\usepackage{multirow}
\usepackage{tikz}
\usepackage{array, makecell}
\newcommand\mydots{{\cdot}{\cdot}{\cdot}}
\usepackage{enumitem}
\usepackage{relsize}
\usepackage{threeparttable}

\DeclareUnicodeCharacter{2217}{ }

\setlist[itemize]{leftmargin=*}
\setlist[enumerate]{leftmargin=*}
\setlist{nosep}
\setcopyright{none}
\renewcommand\footnotetextcopyrightpermission[1]{}

\title{Synthesize High-dimensional Longitudinal Electronic Health Records via Hierarchical Autoregressive Language Model}

\author{Brandon Theodorou$^{1,2}$, Cao Xiao$^{2}$,  Jimeng Sun$^{1,2*}$}
\affiliation{University of Illinois at Urbana-Champaign, 201 North Goodwin Avenue,
Urbana, IL, USA$^{1}$}
\affiliation{Medisyn Inc., Las Vegas, NV, USA$^{2}$}
\affiliation{$^{*}$ To whom correspondence should be addressed: jimeng@illinois.edu}

\begin{abstract}
Synthetic electronic health records (EHRs) that are both realistic and privacy-preserving offer alternatives to real EHRs for machine learning (ML) and statistical analysis. 
However, generating high-fidelity EHR data in its original, high-dimensional form poses challenges for existing methods. 
We propose Hierarchical Autoregressive Language mOdel (\method) for generating longitudinal, high-dimensional EHR, which preserve the statistical properties of real EHRs and can train accurate ML models without privacy concerns. \method generates a probability density function over medical codes, clinical visits, and patient records, allowing for generating realistic EHR data without requiring variable selection or aggregation.
Extensive experiments demonstrated that \method can generate high-fidelity data with high-dimensional disease code probabilities closely mirroring (above 0.9 $R^2$ correlation) real EHR data.  \method also enhances the accuracy of predictive modeling and enables downstream ML models to attain similar accuracy as models trained on genuine data.
\end{abstract}

\begin{document}

\maketitle


\section{Introduction}
The widespread adoption of electronic health record (EHR) systems has established the foundation for machine learning (ML) and artificial intelligence (AI) applications in healthcare.
The EHR data is highly complex, comprising over 10,000 unique medical codes for diagnoses, procedures, and medications, as well as thousands of lab measurements. Each patient record can include multiple visits with combinations of diagnoses, procedures, medications, and labs. These combinations create intricate relationships and complex patterns across tens of thousands of medical codes.
AI and ML techniques are used to learn and model complex patterns in EHR data, enabling applications such as clinical predictive modeling~\cite{GRAM, RETAIN}, health monitoring~\cite{MonitoringBook, REST}, computational phenotyping~\cite{DDL, che2017deep}, treatment recommendations~\cite{GAMENET, wang2018supervised, shang2019pre}, and more. 
However, the progress of AI and ML in healthcare is often impeded by the difficulty of  accessing and sharing large real EHR datasets. 
Sharing EHR data is  challenging due to privacy, security, and legal constraints. While patient de-identification can alleviate some of these concerns  by removing obvious patient identifiers such as name, address, and birth date \cite{AutomatedDeidentification, StrategiesDeidentification}, studies have shown that the risk of re-identification remains high even after thorough de-identification~\cite{CanadaReidentification, EvaluatingReidentification, SystematicReidentification}. 

Using synthetic patient data can offer a safer alternative to sharing real EHR data. Generative models can produce synthetic datasets as substitutes for real patient data~\cite{SynTEG, EVA, CorGAN, CONAN, MedGAN, MedWGAN, EHRMGAN}. 
Various methods have been proposed in the literature, including structured patient record generation \cite{MedGAN, EMR-WGAN, HGAN, SmoothGAN, MedWGAN} and longitudinal record generation \cite{SynTEG, EVA, EHRMGAN}. 

To date, existing methods cannot  generate realistic EHR data in its original, high-dimensional form.
The high dimensionality of EHR data, along with rare and sparse variables and complex relationships among variables, makes the generation task a difficult one. 
Consequently, existing approaches all concede to creating lower-dimensional data by either aggregating variables or using a subset of more common variables of a manageable size.
For example, the MedGAN method~\cite{MedGAN} modeled 615 disease categories without longitudinal information; the SynTEG model~\cite{SynTEG} aggregates codes to higher level phenotypes and then removes rare phenotypes, resulting in only 1,276 variables; the ehrMGAN approach~\cite{EHRMGAN} reduced the variable dimension to be less than 100, and EVA~\cite{EVA} models frequent co-occurrence patterns in the original EHR data as one-hot vectors, limiting its ability to generate diverse and novel co-occurrence patterns. Our supplementary information provides a table of these dimensionalities of existing methods. While these low-dimensional approaches may capture the proper statistics on a small number of variables and support narrow 
ML use cases relying solely on those variables, the resulting synthetic data is inadequate for broader applications that require high-dimensional data including comprehensive statistical analysis, patient phenotyping, billing prediction and analysis, disease staging, and comprehensive data sharing.

\begin{figure*}
    \centerline{\includegraphics[scale=0.24]{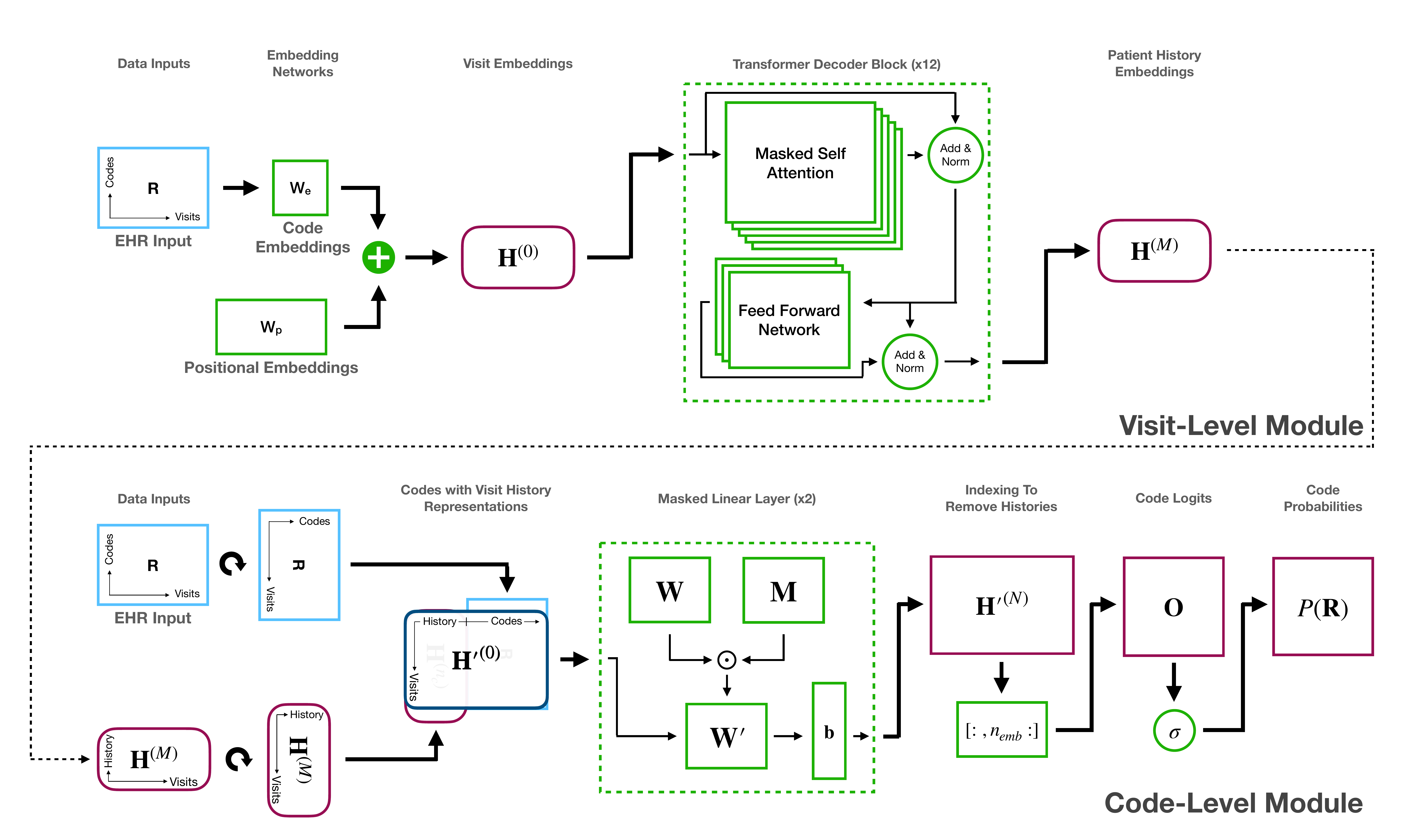}}
    \caption{\textbf{The proposed \method model.} The architecture of \method utilizing an autoregressive multi-granularity approach which analyzes at both the visit and code level to generate next code probabilities based on the history of all previous visits as generated through a stack of transformer decoder layers and the previous codes in the current visit through a series of masked linear layers.}
    \label{fig:MediSynArchitecture}
\end{figure*}

We propose an approach for generating high-dimensional EHR data in its native form: the Hierarchical Autoregressive Language Model (\method). This model, shown in Figure \ref{fig:MediSynArchitecture}, takes an autoregressive and probabilistic approach and can capture the hierarchical distribution of EHR records and their temporal relationships. Using a hierarchical approach to model binary sequences of over a million variables, \method can efficiently learn and represent complex patterns in EHR data.

\method works by utilizing a pair of modules to represent both the visit- and code-level structures of a patient record. First, it uses a coarse, visit-level module to factorize the probability along each of a patient's visits and to efficiently process and represent a patient's past medical history. It then adds fine, code-level modeling to generate each variable in a given visit based on both that past history and also the previous variables in the same visits for maximum intra-visit cohesion.

We evaluate the performance of \method by training it on a comprehensive outpatient claims dataset, as well as the MIMIC-III inpatient EHR data \cite{MIMIC}, and compare the results with a diverse set of existing synthetic EHR data generation techniques such as ~\cite{SynTEG, EVA, GPT1}. 

We evaluate the data quality based on its utility in modeling the statistical data distribution and for supporting ML models. \method can accurately synthesize high-dimensional EHR data via modeling disease code probabilities ($d\approx10,000$), disease code co-occurrence probabilities within a visit ($d\approx1,000,000$), and conditional probabilities across consecutive visits ($d\approx5,000,000$). In our experiments, we found that \method achieves a correlation coefficient of above 0.9 $R^2$ when compared to real EHR data, demonstrating its ability to generate realistic data.

In addition to generating high-fidelity and granular EHR data, we show that \method improves predictive modeling on our EHR dataset by more than $17\%$ compared to the leading baseline. We evaluate the predictive accuracy and perplexity of \method on a hold-off test set, demonstrating its superiority. Furthermore, the synthetic data generated by \method enable downstream phenotyping ML models to achieve comparable accuracy to models trained on real data, with an AUC of $0.938$ for \method data versus $0.943$ for real data.
We then demonstrate that combining real and synthetic data generated by \method can improve the accuracy of ML models even more compared to using just real EHR data. Furthermore, we show that \method generates realistic data while simultaneously protecting patients' privacy in the training data, as evaluated by a series of privacy metrics.

\section{Results}

\subsection*{Problem Formulation}
Structured EHRs are multi-level longitudinal records, where each patient is represented by a sequence of visits. Each visit is characterized by a set of medical codes, reflecting the diagnoses, procedures, and medications administered during that visit. Additional patient information, such as demographics, disease phenotype labels, lab test results, and inter-visit time, can also be included. We begin by formalizing the problem and introducing key notations that will be used throughout. \\

\noindent\textbf{EHR Data} We represent a patient record $\mathcal{R}$ as a sequence of visits over time such that 
\begin{equation}
\mathcal{R} = \mathcal{V}^{(1)}, \mathcal{V}^{(2)}, \cdots \mathcal{V}^{(T)}
\end{equation}
where each visit $\mathcal{V}^{(t)}$ contains a varying number of medical codes $m^{(t)}_1, m^{(t)}_2, \cdots, m^{(t)}_{|\mathcal{V}^{(t)}_\mathcal{C}|} \in \mathcal{C}$, lab values $l^{(t)}_1, \cdots, l^{(t)}_{|\mathcal{V}^{(t)}_\mathcal{L}|} \in \mathcal{L}$, and the inter-visit time gap $g^{(t)}$. $\mathcal{C}$ is then the set of all medical codes in our vocabulary, including diagnoses, procedures, and medications and $\mathcal{L}$ is  the set of all labs. 
Beyond the longitudinal records, a patient record also possesses some static information $\mathcal{S}$ containing demographics such as gender, race, and birth year and disease phenotype label $\mathcal{D}$ indicating major and persistent disease conditions.\\

\noindent\textbf{Matrix Representation} To allow input to \method and other machine learning models, we then convert $\mathcal{R}$, $\mathcal{S}$, and $\mathcal{D}$ into a matrix representation $\mathbf{R}$. Specifically, we build $\mathbf{R} = [\mathbf{v}_s, \mathbf{v}_l, \mathbf{v}_1, \cdots, \mathbf{v}_{T},\mathbf{v}_{e}]$, a matrix containing a sequence of the vector representations for each of the patient's $T$ visits, a preceding start visit, label visit and a succeeding end visit.

The start visit $\mathbf{v}_s$ is a one-hot vector containing a special start code added to $\mathcal{C}$ to signify the start of the record often required for certain model architectures. 

The label visit $\mathbf{v}_l$ similarly contains special codes added to $\mathcal{C}$ representing demographic and chronic disease phenotypes from $\mathcal{S}$ and $\mathcal{D}$, respectively. For example, this label visit will have codes representing the patient's gender, racial and ethnic groups, birth year, and any chronic labels. 

Each subsequent visit $\mathbf{v}_t \in \mathbb{R}^{|\mathcal{C}|}$ is then represented as a multi-hot binary vector representing medical codes, lab values, and inter-visit gaps present in that visit. To represent continuous lab values and visit gaps in a discrete form, we employ a granular discretization. This is achieved by adding multiple range codes to $\mathcal{C}$ for each lab test and for the intervals between visits.
By converting all  medical information into binary variables,  $c_t^i$ represents the presence of the $i$-th code in $\mathcal{C}$ in the $t$-th visit of the patient record $\mathcal{R}$. 

Finally, to signal the end of the patient record in $\mathbf{v}_{e}$, a special last visit code is added to $\mathcal{C}$, serving a similar purpose to a stop token in natural language generation. This not only enables generative models to learn when to terminate records but also allows for $\mathbf{R}$ to be padded through additional columns into a constant length for batch input without altering its  content.

\begin{figure*}
    \centerline{\includegraphics[scale=0.25]{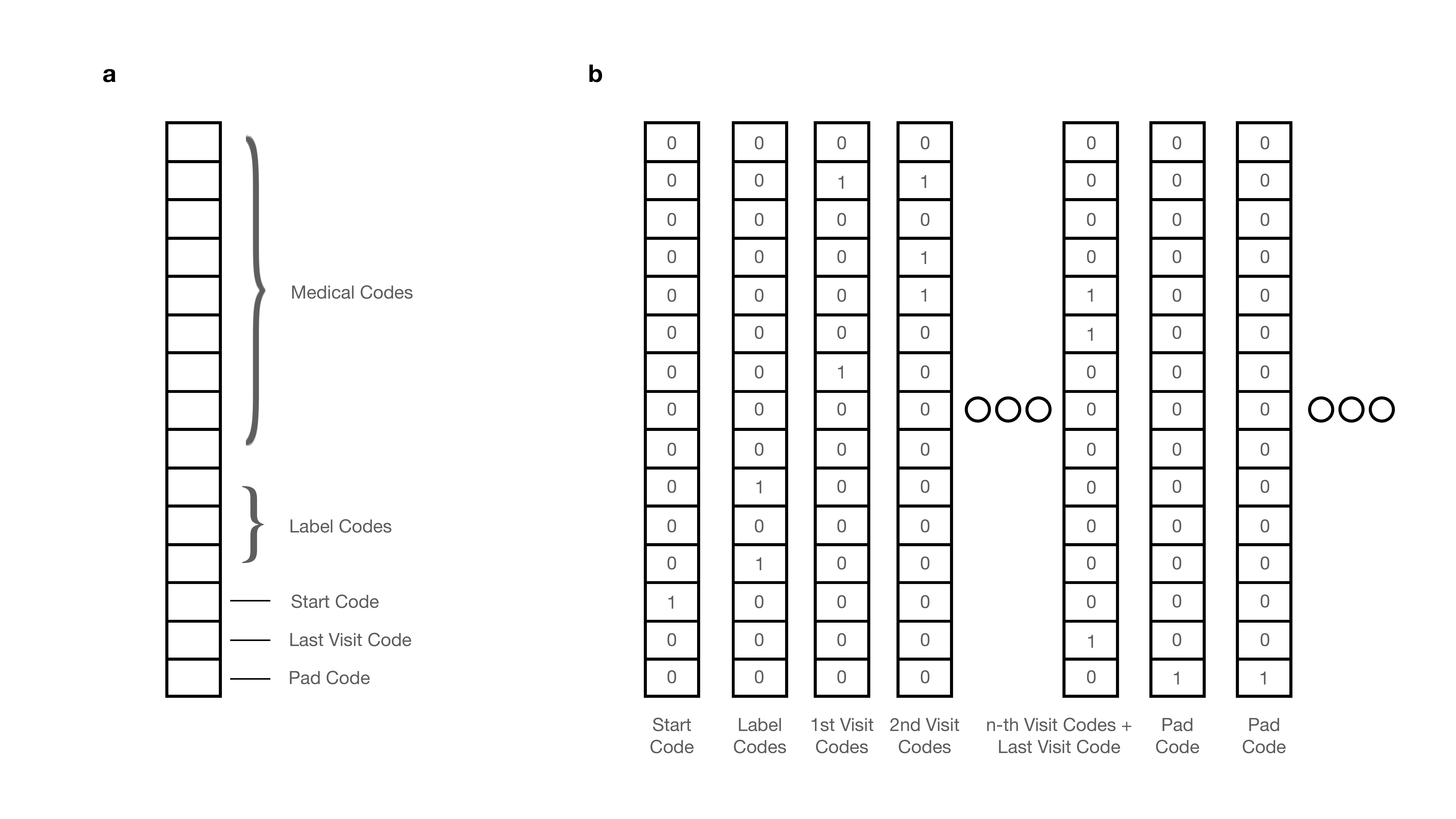}}
    \caption{\textbf{The data formatting}. \textbf{a} The visit representation. Each visit is represented as a multi-hot vector containing indices for medical codes, static label codes to cover demographics and disease phenotypes, and special codes describing the shape and temporal ordering of the patient's visit. \textbf{b} The EHR representation. An EHR is then represented as a matrix constructed as a series of temporally ordered visit vectors.}
    \label{fig:DataRepresentation}
\end{figure*}

Figure \ref{fig:DataRepresentation} depicts the format of the visit vector and the  EHR representation, and we provide a table of notations for reference in our supplementary information.\\ 

\noindent\textbf{Generation task} is to create $\mathbf{R}'$, a synthetic patient record  that is statistically similar to and offers the utility of $\mathbf{R}$ without any one-to-one mapping to a real patient. Our \method method does this by learning distribution $P(\mathbf{R})$.

\subsection*{Experimental Design}

We evaluate our method and compare it to several baselines comprising both recently proposed models and other logical autoregressive model architectures on a series of experiments on both outpatient and inpatient EHR datasets.
To maintain the fidelity of the original EHR data, our experiments focus on synthesizing original granular medical codes without aggregating or combining codes. 
Specifically, we seek to answer the following questions.
\begin{itemize}
    \item Is \method effective at modeling the underlying data distribution of electronic health records?
    \item Can \method produce a synthetic dataset that is statistically similar to real EHR data?
    \item Can \method augment real data for more accurate disease phenotyping prediction?
    \item Can \method generate realistic continuous variables  such as lab results and visit time gap?
    \item Can \method preserve patient privacy in the training?
\end{itemize}

\subsection*{Datasets and Experimental Setup}
\noindent\textbf{Datasets} 
We use two datasets for our experiments:
\begin{enumerate}
\item The outpatient EHR  is from a large real-world US claims data.
It contains 929,268 patients and binary labels for 11 chronic diseases (specific diseases and patient counts are included in the supplementary information).
This yields a final real-world outpatient EHR dataset with an average of 34.16 visits per record and 3.52 codes per visit with 9,882 unique ICD-10 codes.
\item The inpatient EHR  is from the MIMIC-III ICU stay dataset~\cite{MIMIC}. It contains 46,520 patients with 25 disease phenotype labels as defined by the MIMIC benchmark~\cite{MimicBenchmark}. This dataset has an average of 1.26 visits per record and 15.11 codes per visit with 6,841 unique ICD-9 codes. Note that this includes patients with just a single visit (and as we will show, \method's Code-Level Module allow it to be very effective on those patients).
\end{enumerate}
Both datasets share the same patient representation  as a series of visits along with chronic disease phenotype labels. We keep the ICD  codes in the data without code aggregation or removing any infrequent codes.  \\

\noindent\textbf{Experiment setup:} We use a 0.8-0.2 training-test split with an additional 0.9-0.1 training-validation split during training for both outpatient and inpatient datasets. We use the Adam optimizer with learning rate 1e-4 (which was arrived upon through experimentation). We use a batch size of 48 and train for 50 epochs, saving the model with the lowest loss on the validation set. We implement the model and train in the Python 3.6.9 coding language using the PyTorch 1.9.0+cu111 framework \cite{PyTorch} along with the scikit-learn 0.24.2 and NumPy 1.17.2 packages. Finally, all experiments are done via one NVIDIA TESLA V100 GPU with 32 GB RAM. The \method source code is publicly available on GitHub at \url{https://github.com/btheodorou99/HALO_Inpatient}. 

\subsection*{Baseline Methods}
Below we outline the baseline methods and the necessary alterations to those baselines to adapt to our problem setting. 
\begin{itemize}
    \item\method $-$ Coarse: This baseline is an ablation baseline consisting of just the coarse, visit-level granularity module of the full \method architecture. It generates each code probability based on all previous visits (grouped into a multi-hot representation) but without the fine, inter-visit modeling such that $P(c_i^i)$ is modeled by $P(c_i^i|\mathbf{v}_1, \cdots, \mathbf{v}_{t-1})$ instead of $P(c_i^i|\mathbf{v}_1, \cdots, \mathbf{v}_{t-1}, c_i^1, \cdots, c_i^{i-1})$. It consists predominantly of 12 transformer decoder blocks in the model of \cite{GPT2} augmented to support multi-hot as opposed to one-hot inputs and outputs within the embedding layer and final activation layer.
    \item GPT Model ~\cite{GPT2}: We applied the GPT model without any augmentation to support multi-hot inputs and outputs but instead with the conversion of EHRs to a fully one-hot sequential representation. However, this model had to be shrunk down to 3 blocks from 12 to fit into memory because this greatly expanded the length of the sequences.
    \item LSTM EHR Model ~\cite{lee2018natural}: is a deep, autoregressive LSTM model, adapted to generate structured patient records rather than unstructured text as it had previously been utilized, which is directly analogous to the \method $-$ Coarse model but uses LSTM blocks instead of transformer decoder blocks.
    \item SynTEG~\cite{SynTEG}: is a GAN-based model that uses a transformer and LSTM-based encoder model to generate embeddings of EHRs up to a given visit before feeding those embeddings into a conditional GAN which generates the next visit. 
    \item EVA ~\cite{EVA}: is a VAE-based model which uses a bidirectional-LSTM encoder and CNN-based decoder (using deconvolutions to expand the latent encoding to the proper temporal dimension and then masked, diluted 1D convolutions to build the records in an autoregressive manner). The only change we made was to convert the output from one-hot code combinations to multi-hot code probabilities to allow for greater representative power.
\end{itemize}

\subsection*{Evaluating EHR Language Modeling} \label{sec:modeling}
The first evaluation is conducted by predicting the probabilities and outputs of the test set. In this phase, we assess the performance of \method against two multi-hot language model baselines, namely \method $-$ Coarse and LSTM. These baselines explicitly generate a probability distribution without accessing the entire input. It's worth noting that other baseline models, such as the GAN-based SynTEG model, the VAE-based EVA model, and the GPT model, cannot be directly compared in this task,  because those methods do not make a single probability prediction for each code within the visit.

Our first evaluation aims to assess the capability of the models to predict the presence of potential medical codes, given a patient's past medical history and the previous codes from the current visit. Note that we explore different orderings of codes (such as most to least prevalent, alphanumeric, random, etc.) but find no noticeable differences, displaying the results of such an exploration in our supplementary information and settling on a random ordering throughout our experiments. This evaluation is crucial in showcasing a model's ability to learn patterns from the patient population and its potential to perform well in various patient simulation and extension applications. We show the results in Table \ref{table:TestSetResults} where we see that \method outperforms the two compared language model architectures. Upon closer examination, we observed that the LSTM baseline model struggled with the complexity and size of the outpatient EHR dataset, while our proposed model \method performed comparably to the \method $-$ Coarse ablation baseline. In contrast, in the inpatient EHR setting, where the visits are shorter but contain more codes, \method's multi-granularity approach proved to be highly effective. Specifically, the model achieved a notable 4\% reduction in binary cross-entropy (BCE) loss and a 17\% increase in F1 Score on test data when compared to the single granularity \method $-$ Coarse model. Notably, both \method models significantly outperformed the LSTM baseline in this setting. These results highlight the significant value of our multi-granularity approach in handling the complex and diverse nature of medical codes in different EHR settings.

\begin{table*}[]
\centering
\caption{Test set modeling metrics}
\begin{threeparttable}
\begin{tabular}{c|ccc|ccc}
\toprule
& \multicolumn{3}{c|}{Outpatient EHR} & \multicolumn{3}{c}{Inpatient EHR} \\
& BCE Loss 
& F1 Score & PP Per Code & BCE Loss & F1 Score & PP Per Code    \\ \midrule
LSTM    & $7.744 \times 10^{-4}$     & 0     & 660.204
    & $2.600 \times 10^{-4}$ & 0.193      & 74.565     \\
\method $-$ Coarse & $1.631 \times 10^{-4}$     & \textbf{0.829}  & 3.927
    & $2.019 \times 10^{-4}$ & 0.343  & 28.448 \\ 
\method & $\mathbf{1.624 \times 10^{-4}}$   & 0.828    & \textbf{3.903}
    & $\mathbf{1.932 \times 10^{-4}}$ & \textbf{0.414} & \textbf{24.664} \\ \bottomrule
\end{tabular}
\begin{tablenotes}[flushleft]
\item We include each of our autoregressive, predictive, and likelihood-based models. The bold value denotes the best results. Baseline methods SynTEG, EVA, and GPT are all omitted here because they either do not produce a probability distribution, peek at the outputs, or utilize a different, non-comparable data representation. \method outperforms both of the baselines, achieving up to an 4\% decrease in testset BCE loss, a 17\% increase in F1 score, and a 13\% lower perplexity per present code as compared to the leading \method $-$ Coarse baseline. Source data are provided as a Source Data file.
\end{tablenotes}
\end{threeparttable}
\label{table:TestSetResults}
\end{table*}

Additionally, we present perplexity, which evaluates the probability or likelihood of the test set as quantified by a model trained on the training set, normalized by the unit of consideration that we are interested in. In our case, this normalizing unit is the number of medical codes in a patient's medical record (or equivalently number of ones in $\mathbf{R}$). Perplexity is a metric found commonly in the wider generative modeling domain, especially on the task of natural language generation (e.g. \cite{GPT2}). We introduce it to the task of synthetic EHR generation here. Perplexity is defined mathematically by
\begin{align}
\begin{split}
    \text{PP}(D) &= \sqrt[N]{\frac{1}{P(D)}} \\
        &= \sqrt[N]{\frac{1}{P(\mathbf{R}^{(1)},\cdots,\mathbf{R}^{(|D|)})}} \\
        &= \sqrt[N]{\frac{1}{P(\mathbf{R}^{(1)})\cdots P(\mathbf{R}^{(|D|)})}}
\end{split}
\end{align}
where $D$ is the test dataset and $\mathbf{R}^{(t)}$ is the $t$-th record in $D$. In practice we calculate the values by summing their log probabilities, using the equivalent form
\begin{equation}
    \text{PP}(D) = \exp\left({-\frac{1}{N}\sum_{\mathbf{R} \in D} \log P(\mathbf{R})}\right)
\end{equation}
The normalized value then also corresponds to how many of the different normalizing units (medical codes) one would have to randomly pick between on average to achieve the same probability. The results of the perplexity evaluation are shown in Table \ref{table:TestSetResults} as well. We see similar results as with the classification evaluation with both \method and \method $-$ Coarse performing very well on the outpatient EHR dataset (with \method performing slightly better) as the LSTM baseline struggles, and \method easily outpacing both baseline methods in this likelihood evaluation for the inpatient EHR dataset, producing a 13\% lower perplexity per present code as compared to the \method $-$ Coarse architecture without the inter-visit modeling. Thus, in both of these test set evaluations, we see that \method is much more effective in terms of modeling the underlying distribution of EHRs.

\subsection*{Statistical Similarity to real EHRs} \label{sec:mirroring}
The second analysis evaluates the statistical similarity of the generated and real data. For each  methods, we generate a synthetic dataset of the same size as the training dataset. We then compare the unigram and bigram (both within the same visit and across consecutive visits) probabilities for each unique code and pair of codes within the real and synthetic datasets.\\

\begin{figure*}
    \centerline{\includegraphics[scale=0.7]{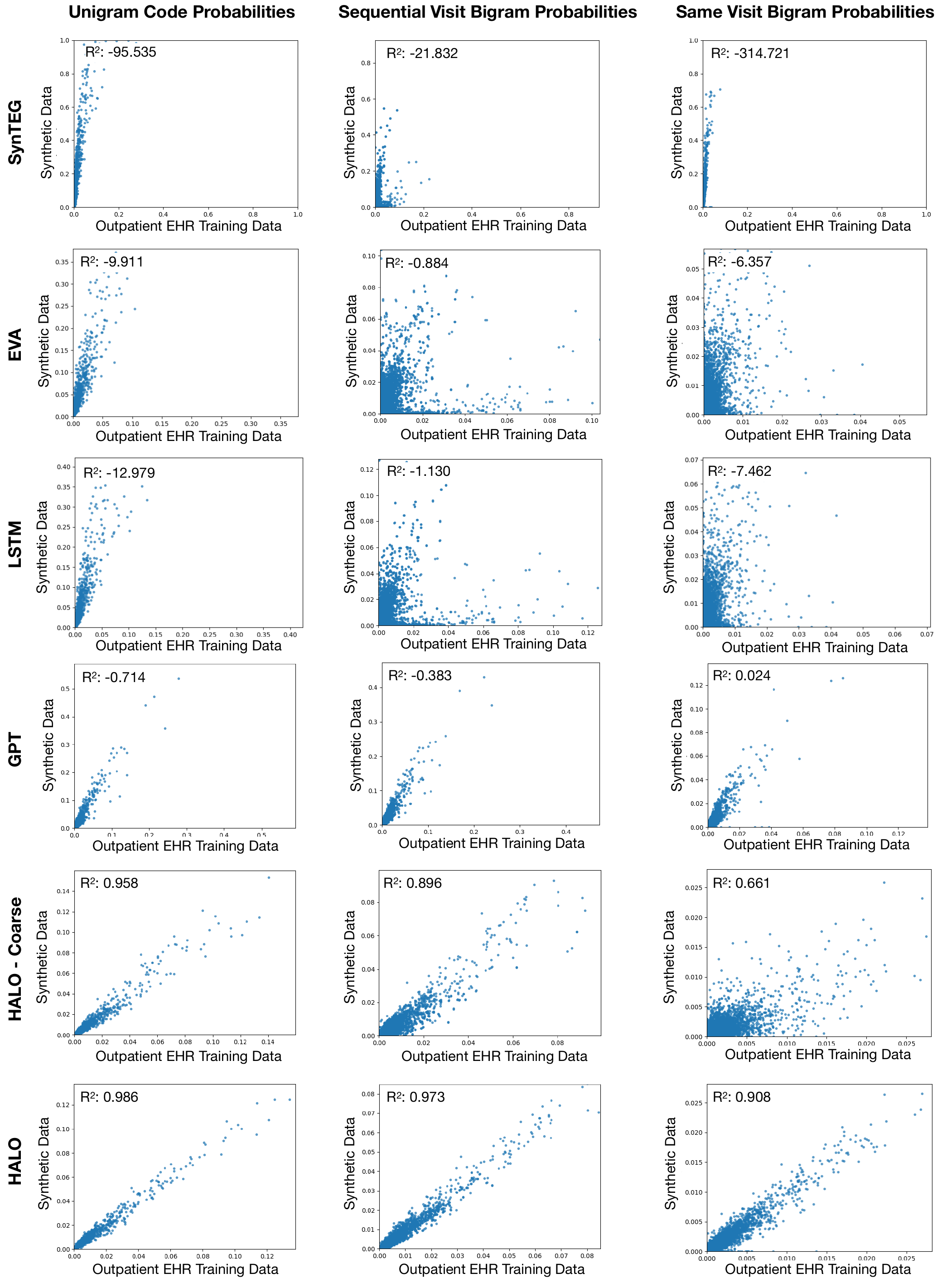}}
    \caption{\textbf{Code probability plots}. These plots show the Unigram, Sequential Visit Bigram, and Same Record Bigram probabilities for each synthetic dataset. With the exception of SynTEG, all models exhibit some correlation in the unigram and temporal bigram evaluations, but many have weak correlation or consistently yield higher synthetic probabilities due to a lack of temporal consistency and repetition across visits in the records. \method and to a lesser extent, \method $-$ Coarse perform the best in all settings, while \method is the only one that can realistically produce pairs of codes within and across visits and achieve state-of-the-art results.}
    \label{fig:codeProbs}
\end{figure*}

\noindent{\bf Statistical comparison results:} 
We evaluate the data at the visit and record level, considering approximately 10,000 individual codes and over a million bigram codes. We also compare various aggregate statistics, such as the number of visits per record, medical codes per visit, and prevalence of chronic disease labels. The code probability results are presented in Figure~\ref{fig:codeProbs}, and the aggregate statistics in Table~\ref{table:AggregateStats}.

Additionally, we provide $R^2$ values for visit-level normalized code probabilities in our high-dimensional outpatient EHR dataset and a lower-dimensional setting. The details can be found in Table~\ref{table:VisitLevel}.

Furthermore, an interactive visualization of 1000 randomly selected code-level disease prevalence comparisons between our method and real data is accessible at \href{https://vega.github.io/editor/#/url/vega/N4IgJAzgxgFgpgWwIYgFwhgF0wBwqgegIDc4BzJAOjIEtMYBXAI0poHsDp5kTykSArJQBWENgDsQAGhAATONABONHJnaT0AQXEACGuMxxFSKGtI6cAGzaYdUNghxJl4sjoASmgDIB5AOQQOhAAngbwalA6SBT6ELb0cDqKcEiWOrI0EClZdmzyOgAKycSpcOJQCpTSIADuNLL0aABsAAwtMvA0ZFjNbTJOshmuaKCYbDi9MpZwAGaYaADs7SBMNmMIaALLyt3zqABMLQC+MkgMYxA0AF5waCDiErcy9uIzXSMgSAAemR8RANYAYTY1kUdwAxAAOaHVSxIJhwSwAMQkexAAFkJCY2FIdMCGMojDoAHJwGogI4nECXMjiVIQNAAbVAdIQt3QyEUtEkMhKlgY7MOJxZSDZdxgbFIYJkEiZoDgpAMDPQACpUAg2AwspKjNUyvZ5OKdWDhSAFWVMMqQGqNVq4Jr5jJ9Xl2SBpkhSBSpPLFZa7jbNVlZGwajyzeUXXcstNTF6fRarQG7QwJk6I4b0Ml3VkKQBdU2s11fYyuJ4gFOyJCGO6M9o6OoNGC5uP3UWu4IlshlitV12Mzq7XEtZsFtt3YOh6p8gVocQMSyWGUaZlm31WsYMWBlWTVHvV9Bzhct81Ku62oMh8S4jewOLOR3lnCV-cgL7BAAUAEo86OxegvlABJTqkM6oIei4gLKqAriefocoGcATleOg3jAd6KJg16alu4g7jIe6uhAlg0BU75fMGCDfkc+beq2f4gMEgHSiA07suBS5yquCZnghSFYZuaGYPe-E4Xhj7PoRxGkcEFFUTRIr0fIlhCcB-LsrWUjDhx0HxqeOnUpqigVHcdS4SG1SYMEOCuueiA6tULwQAwf6YIoAoyAimA1HAZScZZ1k8XaSEtmIBLGegpkThZVk2QhKY-rR-muqhGqes8EhOS5blwDIbzKbq6CwZQqGVNMrj0DoAC81U6AAjHm+FPr246XjoAD8OiMkhta5gAtF8X64h+n69d1tW5rmOioJ17TDglCmukg5QSsxrFMrNubaTBa4mfAiK7k1L6MvoUqYO+fhfNApR+LiA2fp+uInUYZ1+MEV3TDdOjDZ+I60bByqjDFdyoehD55YYYKFb6xXYfAECUGVZD0NVlX7F64nNegjJkRRPU6AA1Do5EOIy42fjoBA6PsuLvjJJPDgTX242TFNUyO8l0a6VxsA4qmgbVW26XBtR7ZYACE1QzGwRnsq57kYy+OAhu+tWUG0As6EVSlCQAmjoKoWMrtVNLiWuIkJmLyPdx47agoA0nSljjpkmBo1SUsy2gcs5QrroZHEtWs-7rvzZzzsB3z7LLFB23cXbIBJcDsOg5LND5ZDXEGDDAmlWUSMwCjbuNRJdw4PosAACIu0tpGwVRf22-bXSO+HIdUgRrduxzhad5HaDR8uQsAwnQPoCl9npeImWy9luVpxDdxFSV8OI8j1VF77pflzAVd3uUcDvnXoc9-+FEHSX6BESRB9fHAXyYN+guZ3BTe0qk47myg7eHX2AEEnjhMIBOHEGRJi5MDbayQHjSmDZ6C3SYqTSagDgGgIJOA9In9oH1nqPQX6Q8PgO3fugbmvNv4XxAIyGuK0AE6BxvTSavUojLWlj1dBJCEC4ioSwhmhM6EIEQToRhXDFCsP1jodh7NfztjPsXTG1IpIH2CHfB+1QY74PjoQp26BIHow7ljRi-8eFBBQQYxQ6DIECMpgOLAQ0EHjUZkApatMwFiIsfYqxcAuhYDwc-Yemi7jsN0T-GswiBG8LpvwhmQjmEiJZgbdhnCYlhNoREgR0TYAsLieInmCBJG0RPtSa43ZgnoCgHCRw75Dis0cSA4mlFcQa1qm0Fo9cFpRmbqkAA+krfQD59B0DuMAL4019gCCGiMsZ6RMjZDgNNPwfgdBHFUYPXxdwAACnJ-jdLYL09UCFjTnzkUM6ad1xlfUGlMrISAsjTToIgL8lBnzOUZH4YkbY-D5htnHEAGznBbJ6QYPZdoHSHJfMcqmkzggTNxP7GZcyFlLOotRGQz4UD6QKbC65ChI7KhXAAJRSGkIoCpSj737mrdWtUACcAgmiQgEFSpoWwADMkIZCeF8IUYopLwotApS0WqAAWLYtVmUCAEIcZljKZBvPogAUUBAALTleiTQHg5V4oKHKgAKgASUBAAVXRC2AlqQuUkrKry-lLRBWCshKy-Y+xmV0shFS5lLRmXsu8D4c1fJ9RR2tZCFoCx9hUupVK2qrqqWQgWDKsc6BiRyoAOoACEfB4uJDoTQSIkQKu1XKiuOgU16x8Nq9wGqdBInTeiAAyjoHwSJCheE0ICOVxJtXeB0DWuVBRNB4s0HqnwmbNDEkLeW9E6a8WeAAOJypNYS31PKA3NLaAsQVCx3XMpDZCIV0IpVes5cSv1ZLUB8pXcywVrLBVMqaLeqlkrqYgFla6Xts6dAAGldUV0TTredZqj1LvJSuqlgrjYtH2E0Kla6mgLAEFug9PqAOWuXW0fYtVHUxuhB6lo4q430RTQarwvgDV1orrqmt6aK4atxAa4kNatWAl1UiXVBa-1Eu5choDzSmnXsdRK6lgqWj2tjSADliGOP+q420V1kGw02rFbay9eHXQVzxQa6dvVdWjoNa2wtgIB19pbdqtji7OOnv5aB0ZtVjYCAWG6ndCHTOSfM802lQaQPMv2LBhYDmn3xtEzrLVeL810d1QANTlXiPF5HyM0bowxpjLGK4maQ85s9K7Q2urQ4KsNoyaXOsc6lk96XUMtHvVBuTYHnUiefXcUt5a8V4iHbO4kuqO1eB0KqrwVa8WqsHXR+tja03Ej1mW9NmgAAaKWJPFetTau1DqnUurdR6wrM2rUrqaF511W73V2uswK5TdwCMZsGzoZteKdbEgm1m0dOhtX9sBOWzQ02LVpbm5l+9QrcsMrgybUT3qnOzYy-sQVoOFgLFaNCJoTS2V+fogUHw5Gh1aenUWnWuJNAVwi3i7tOg5U5rzWdkdhbaNUbxRXHwNbUdnZrTpxjrX0RfrlXWkn9aJuaEVT4LwX6tMFt1UOlnr3j0bZXd969wbhVbag3Stbb3gfNMdWD9dkOhOQhh0Jo76BtU+Am-q-HhPATarrQ2vEfbhv3Z1Y99wvhotUeF4Blz4H9jCdaOGx1Lu5ci5Qxhy9W7qVCbdQLeHromc1u7YWzQabgsC+JA7szJWGV2rg5D1XTRBVe8dyV4VmWbWQ+jVSqlWuQBaaRF4A1bbFVqorhX+7Pr6sVuZ+2xLBadCl-L5XtVYXdVqbrUmtr7h61loraq1ruaa0dv6zW+P72V0CpyxLtdtLQ2Q7h2JoHovSvlbs00qrMHi9IkdS0GfCvuML9aEv6Xq-M8J+tReq9N670PuLz4E-m+ysLHF5Ca9DKL1hpv7PqhluksBBhBnBvesXninKr1G2g9nrBFu2tFoahdlmnilOjrGWkzi9qaKauxvLpvl5ocJGuKksNesHuvkVgQStrVEsGukrmMiHncIjl4IFmdmFj4Pph1oCJRrdoWhdldlNjgQupQShm0KyvShVqyk0LttKgDoeutqIWDlSi0BfrVAIJCC7jVv5g3o1twcSC1m1p2l1j1n1rHibo2iOvmtFuml2nKtOuiLAWdnKjrHOkIf+goVJkypDoQUGhDmVgAafmusKuBrSjDqBnDrVloEbuFm1ixnWlpuwV4N3voToBXCOgzmjmzjoToOgZgfrhOhFg4e2m-qIerOfpLsvjLmvoDiIVJnnvajSo6p5ktrIZESAN4IquWrqg4Xin4KRuRnKpoN2iUXUThlBt-pllZsoZ-gEe-p5tCIyp-gyreqGsHm0dkWRhRhThquYToAatFsSH2nrDWjrBPiqiMU7m0IJgpoJqKocB6hurMaUcJtSsbIJunkKrLowegNkU9umgLlAXqocR1u3poOiKYUOtPm4Xgd7lxk0jeuKtZtIYJo+hQR4U7ocGDt9q0GGs6msf5sSAag9rEUOp2lRsxoxm2oCBjnsfFgqi3sltCRvqUeIQynZlITIf9mifgaIWhhhj5kGu6rht8SABqhgeOmqiOoSRdpFq2ogUxupkahcSVioSGrZsymKnBhqTSgwdybCZcZGvanStZmGrap6iKUiP2tOrql4JFjdk9nij4BOhRg4cqXNravahoUtgsZuk8XUdcenssTSiBoXhniKbARqkjsSGFkMeFpFizk9hqvqu4LqmqtwTzm3sSLmtEUOm6WUcoehs6tqZoX6RiWoSGsiWyaDuaW0d4Nwe4NzvroxmgQ2dTrsTzjjnmVcYJsoWGoXhftUfITyVJuKtelKmQfejlsXr2gUDabaVRqql2QKmETQa0BKkGuQTUeiSVlKpDreuBittei-sPo1vRvScxq3hPv2imuRuiCRkueBg0bBqBnekygyqWSVnyZoVhkKQwesSeakcjtsbjmdjrvVqkXKhFr4AUEUcZkybUU7oXqMi6tZiGtuh+WrEKhDgJrBuBjxvifRJsZRhWqbp+t+s4bwXsYCdRrSWeYxheYybRLgcyaMeLhflLivl8XqVntahBr2RuqKioRqaGsXu+oFq2eRpRbbpTm2Q+WxRUVflxVucOQaR6mobuc0pDsXjWl1h1mRT+nFnRQyUuVhYyjDqyjQRehhYrmoV5j5lsMEf9rWURhqtaYCE1u2kZhBb1gOrEaRrXjrqkWptOvEZmnoR2kbjoP3mWl2p+nHvBdubxZBgeRemhhqeoaicpfqSqeEb9hoR6ocBumGW0Ymj4AUM2jpWdrRq2sFpoFpkWnKp4N3jYaboCP2q1mVZ4JdvoYMR1jXkbg+RDp-jSjGmheum6tZc0l6aDjQVGuWZ7iKd0XecSD4OSfqixsSNSVFQPoUFAZTkzoce2iwbdnqmmhXHrOSXmjWoZQlgxSZbiR-gILaioZuUOdlfyrZXat-hDiiX+f5t2noWkZdp1t4LqtOoddqiSHKmVRVeiMTqpm2p2tOs2qOkuYyk0p5knlKiBpNYWW6mDltgIMbKyseQ1joN3v2iCVmXmmYVVXSfRUll2m1q4UxcIYlXPp9tlj9vllyVlTxdxiNZevNlsCGsXimkZkmZ2gUImkaijpoDdeeUlkuX9rSoJqvrDpNcGg5RoVsOnp5gRa6HoQYe1iDd1tWr5UOmds9sFgrQzaxglSpSqQeVlt9uhr9gVnIeJo7XNjJoXjxjcUpiKV1mDRDVDTDUMXDabj4Djs2gULOSkdTvmu2bqgUA+aMoJvahBt5ioThprcGjQQyi7hevaqOcVf5lAZaUZumnrNSTrlqq1mqtFYPneV4HqjzompoG+udfRhVamXmSBkVdSsoWVlBplW9TxfCTBjDk9Z6UnsXtqoFpFvsIBVHjqizp1nKkRm1iRttWNkSU1uiOVfqhbXRkuYJjhk9XZhoaNZrYXfPmrhuhNeGbaTGfmnpn2gzj4Cqimpdiju5ZYWDW2p1K2poD4sxQhU7Zza7Xln9nnTDqGpejGlBuKnifPbrvrgTtmcbmdk9rbg2uXumhwebpCQ+fPiBuxZUdfp7SxZcQKuydGocDGmrursXrRkZQxbtaWj4FzmSQMUMSzaAOA+zSujJpDrahuhDursGnncyqKuQ+rhKoJoicXlHitT5R1i-ZvU4cSIqjrA4aFV2glkUXWpaU6ZWiqt4JFsbZ2umtOiOlCaze4d7aVqGlPferShKmPV7e9ahnxaykVcuWumLY6ZtSmXmrGddbSdwYfTzvpm-XJeyeuQPSQWhprTQdSk6kXRek5f5nWT4A2bE5Wrqj-VTpJWzjWoCLaY6W2Wdh2RqiZRqeBm0NIRrmGdxbfs0tVnuZ5vZZ5qTRWtE0fXE7TVWkRj4P3ikUtbRrqlXv1riCtZDZvd2kmg1rKRVdTndQ7T42IbBp8euupdI9QxA-yp-rBvevagtsKW0c2oCO+mdgUO4KccfR1nKhNhqoxsMVs-zThsNerl5mKlZl4zQ+lrrXZoXlKq0BBgft4B1pbTWmCZFpox1qY3DTWjznHajriMSRiykeU9qkaimu3Wjk3T4PvXTimt2gAIoV7tpdrElxWZF2NaYT71opoABSeapDHpi2zqPpq2RzwjYh0O6NhpkGoyNZuThq+aUNDz0Weq7lJxo6jpDhu9g+VanBlF3a9hsBSNvgPRhGfa52LOZhXZu+GpiDKDkGg53j3zKhEGUqMawqGdf19EbDt1jNmgcdbaZGjGeqDjgjbNzjYu5Rl+nF1rwL1qOW5rIt45HmWh9E2RbObritreRFwFuxw2kWbO+mMezas6pDbJExDDRN0xr1NrHTiuoO4Oqu0OGtIpamBQ+LUB9znrc5RxnWJGlTcq8zpa+OXgyzqzeI6zxlXzFbAq6uOtbqkqotArQbIRoG66vGGpsGxeUrI60NO9mgRJVObaidMzJ9pDoj6e66wBUjIm7TgBAqF6M9CJsjOT8qfax1RtbahhHWJxDzNpsWtF7r9tjjMJtr4GEa-uK2wa0uaTHmdqYxl6R5IpvxzWL7Jtxh5tU+Z2P9gxE+cl1KbmyxLKO6bTfNY7wamNBN9KtK6G9KxeE6OueK1JXbQVcZjWWmbL+xre9V6ZPO1OltTdRTbzhaNhE+J97bPg7LRO9VK1iB2aFJWagIjGVGxRo7l73ZnpTRy2vps72zAqaGZWW6yhmGZd9EPR76WjraRGZuNHWmTpaqpuzhm9X6uIWmlTBqZGKR+m7a7g95Cnp+AqWHdKOHHqeHedn2UG8+RN+20LNprHdGD2RqOrVN-V-WtTUe6apD8lobVRedAZwqjKwZOWONIpiRMXEF5VOqsegFWx5Oprtm9qOGJ79K-LF7CukaRdt6a6W66h-Tp5P7haab5OuxEWzLKXIbHF6X6nAHNBbqDDQmSw0ILDIp1H8r3OHHttI7f7EbK6pzoG5ZyhGGFHo3Y7JpDKWG6hYxq78HrWJt3dYpZajpiOLG8rt2Kqfdnn7+Amn2kGsj6u0aaTy56h0aUbLrrocHPV53nayZZaOspT-rIAQjc7+Zi+w3VDDXL3EHQaWNvG8boeoN4NlhYd5VEd9zSZDzGqnaiaeKEWuxz2FcuISILaWbd2iatzD54ROW0mEO0hYq33y52N0aW2+nroyNOs3BLWdL42XdfDnzq3xzzS66GhrXUjtlmttqVm4qeHO6Re9bCq+xUBtLBQUB2Pm1esvg4eJl16+zcG+dMxe3l7ahI9f+mTheYtL74NuPsNxOhJrOd2I6narn3gclIGb3Qen34bUv6se6xprxZpxeVJpL7aFaPW+OHOB1J9laNhUBbVBqbWsFdeqBDhhaPWraUPMPGnY58xsjQqnjt9jKuX8mNKGPdw0ftGVhKfjWWOzHE+WfbOo+ndKqjhUdAF3RePsf3XkF+qQuz3pRRdyDL5kGb56vSPpR6he2oGPmm60IKjMeFJqZsL9dNOAn+LkTybdtjFAbTj2z+Fa6cjSw45uNApqGnxV9APdWAF7DjNmx6H69puBqXrjWUB9GPeA61de7I9jgqS9BWoXIVDlnTxmV8O49Mdp-hxI+Ycs3hdXm0WRpU492hfQNhpwFT+8C873SNFazzogF08mWUDGViDR887gQJQYiSDBr9o06M1MNNZm-y2omUQLEPlezULqE6U0hBRquxtxgkdYZGUHtTRzKZpTcI6TdndjT74tjU4-OoqlwR5KVYBinCVJ5nEbPk2gYaCIv5lu4UZWsKRYtJjmxw7F4yhuSGhILuxk4NUMlGnBIK9Z6ppWVGXXiznhbtpMBp-ADmoOswQYP8NqSNOewI7W8GUjKC9OBkhxPUVGFLAvhmygJDFIaQxCjIxgHSt5uOLg-XtSSxblpCgFeTVEmSNRSUjMB7eQbQydS4lhq7qUDiJSt5ec4MavXjPA1vTtcRSqjc2ho0gpaNrOOjPRuvXqpGVjGlaZVqbSsYIcbGeKOxqfRKFnpqyoaI-JZXMq80VBwOUHFLglxBomia-EUlRkeaqZihoAudkiVsz2Y1CuWcDuoSFq2oRaj6WsimjUbB0F6xOLwARgOpdltsO2XZteyFSTUXckaG1DwN+HXD-MjzILCFmpwRY8Q7gEJvrn0oUU3+-DVVnS2746ABAEIqEe5RhEXVxekWGwt60RFvptevDbtPwy7JIV0MNqBBsoX3Q1DRcSDMVHhUZRHcVGXgestziK7esqSAjaHlgIA6B4p64jU9hfjzpMCQMgeMHEsHSYddDGKbPTGd1fam0TCJ9XYmRlnQT44yXaU4vmjkH7DsBztL7Dljdo80ZGwrGHKK3cYSt6Iz7EHh1iCw9YUhhaQ4vCzfY9pZBS5LOnBhPbXoIc7AwVsbFDRakQ0cmR-loBcrjCmyQ6CKpDXJymE-WqRAKj6gcLtYPB-7Mdo03dQaEZey-TWuOSnpjEThQY9otEW7x6peh0ZbnMkTRyj9NWK1JNCZXpQak7iwqdzOB1FTHcssrqSgegAKCEYJ0hxYGtDWpwrUP2POeWiSF7ZLNk0g7SpokJW4n9kxinMYogMmIlt86GXO1DlnFRoYvRaGfMaVTx6VVP+m1DVB2nqoppGqmgZqo1lNzx1eqOgZGiTlNaoZoQVWAuoEKWGi5RU1mXfLp3TxBNYOPOadE6TbSTpnsy3TZtqNtY8YtsoyNDCKOD6Csgu4o4VLukd6tZneu413qbjqZ4omeC2L0ryx2z1cghXnQ4NvkqytBqskBLHALl8BuUpajpXTCx0or9oyMOhT1nrCGL3Zsh+mEjNiIsK3C2hORQYmINArZDe0xY9tLiBaqNpm0TfQZrExPq4gSW+9WCqV1Nxh4scCBJtlm0hrD57s3RXiZxMiy68OCBaFjg+XgY+ZlCtmITJiTglzsvymGQUhfWLwxkkCO7cmnKmZa1NxmFaV5g9ke4PD6q6Qw6tSVIZQN9RMDD2vPzqJK5q2UOdXHWzaK9pHsHBWTgLi-ROlBekPUCUrSmH8pC8T1ITB43bHeig2RBLdB6XvTeZARCbZ-l1wgomEaclOXHGVQHRhMkxa3dWBugqmm8VsupIiZvlXIZ0nU9KaQnX3QARlqmOudEDXSdJDNFRTPdnkXWsw54Fq0U2hnZidRg41cpvcHJRyx6h10J+PU3NHnJygVmsFeUhmug3QPEfhX1KketJVI89kqWFV1NCEeIikqWFeZtAZJ1x653KzBK7HKn2KtTweYUgvFzQNGwNqRLJY0cwM+7iti85aLwFqNnGdTwhYqD8YGXAwwDy2l7bbKqSaI5YY02lL-i4PDyt4bmk+WPF2XSplYR6NBHAeaUekUpQcy+VoIuzKEG07gwIo8Td0ebnTQs-bBEbzOCz8yS0FhOTkjgrgdSOBcPChopTsk6iEJIXXdNzM7E7FLaaaPtM5zRw2Ese7gIEpkTM6ZlmOe1TMk1i3qcdM03HftDaSTSaBRsds8QT-zqpeAHZ8BRqvqltIPis6+WJXBENxrNN0M64q9E6lXYaiW2m1OIbLMFbHsC6O6Z1CQM1pZZdaauTzPvhaFStIsJTUJn5VjGRZAqM07gnuNVSFAZaPY1MpyKL62tdsTKAgWrVtSa1Rkn+CqeoNl7Qsmcx1FMhPkAESDmR+TVkVHh4mUVnB7Iw8XmVkaXoThkHJPIdhhlAY6RsGbUl6I3ThzYOZxHwH2mu5+Ucpl5ZmmfW26yM2ZTQxYXjMCKF4d04oh1Jb3-Jk0aq3eDVpsXco9cdiS5GDE0h8zXTRqZ89GRO24E9MbM+Y15p1TLTeAmc7lQGiSUmHgSOmaVI-LJiWIE1rKahW9J8WdSHAXctmSUS-1TZAVeuZ2FZszSLS+BVqeITev21plXEwM0aJYvakmqyMdOG6FrsJRQH+Y7GV5fXIRhqp7zI8rObOfWl0KQi-6ikgfKS0hrjpgJSImwjaPGz9YHy9+M5p-gXzf5b0aTRpn9mUXvkNex4xNIWi0wPYt2VhPnB5UMWtpG2aoqjI-MixBS9eIU39HlLFyJMXcyTWzKkwXkbTeeYOfxppz3LF4jp+4xtNVSPF1VM0p4pqgCSqrUVsJjiylDSl86MpcOaTSDh6k8aRo7Uncm0nrB7nUcJZEFbNImnDyUVraNOI3rHNh6S4bpW6O6Xujn4DTni+VTbucwIFsL6ITHfYnkuyKk9yeWaBIcSHfS2lKKVaXtki2VpPVcuScl3OnhTkZMiaavPwvmO4IZobxhi+LCjUhp78POsC63qBloXTdi6pUjTlwIzHrpL640kACdnEFBL6aDJCCtOigIGTKe1PWnpjnp4Kp30prQMsvncysp4MHi9LJguskq4NSW6HQbVOAmDElyV6FaUiQgy7cWZzSLbNIVoVOsipYtfYhRjEmPNeFBbcYkoR1oB5LeCK6Xm6mvZEcQyl6c5a80Rx05m2gVYyb4B6p8FxmWaHvLtVcFmTYlNqI+SHNCK0pFe24+lHZj+HNDrmyZIBoPxIw4quVqGCGdA3dq-y5ZJElfGRI9SZzayDhG8u5R-qhKs00Q1wXJU+GejJ6O+POuNwdaMNpu0OUmaGI1ZlVieCXbjgqIS7WcJsVhYEiLzLkcTdJ3EvHBIP4nqNBJRmVSY2l0liSNq2qSSZeOkkpDdCc0+SXM1VbiKN6zeS2mpPIwaTYCjEgdIZL0kqtTcYa+iaZKgK4rFxxbQlWW3Rncro25KhfH8raLB19ekNAJZHUbQ5tS0X6ItKtQcXbKvOC4otgStLZ506hAmJFehiaFgrXQ3qBLGqjvKf14Wg3chgpTDaBcq2KueKRrknV3AziGaSzr3M7XtLgaNeQuT6hvLBZB8h6ntWjLllkN4elDZQefJe7HtLmV9Kefe1dBtoLss6PgnoweajiqM5S4vgPRt50oPUwqTWjBnVKY0XcYG4vFhPIWmcc2n9RcjKtvUKyV1-y61Ad2jSo9Uek5VdkSXXaks60hLeIZlO4KyVUN10zdNul3QNEiBBdc5sXU1JNyRS4VLyrbNowGC0cDZSGg7KsKLq71iswLmVhVW75yJ6q-zAYpZxAkbGFLaLsJOPXZ9mWs5cPHaM6w2kPlMq+YonPTz+cZ6dSt8Qv2q62pZqq-LdegFz6plVRHq98ODwgXkwv1Oc3qCZw0a-qGyKG3te-kn7PljYM-cVIZsfWiFyObmeATNXNEvoK5ctAudnyvLOiOCqUqVbeMdJf8i0prOruoOwwyZ+pRmuEk6kQFiMtBoqfxUOl6jkzaVspYRa1ncrcKSKjaGtHFR7aLN+2E4jVGs2nFgTr1wjLbJhhbE8DI+mG5puui2zLz780xSAsaxrydpJFLZaWZ2ga31UO0IVKFZuhskZiaCivUIXSOeogZ8xjaw6dDVLk4Mh0DPPVOCPKYNodJ5GOnN22-bSjaZlw4NOOynow5Jq20vZVX15b5iW0RJRFkcX0J+VBN6GkbsSvVh-C7UOpITOOw7EBYQRu7NUQ6T-ofov0P6Mru-xMqf4kGQqzmYr1C70p9RYaD6c5SqZuVTFXlKMb5RjFKbAqWOctGCLH5eaUMBdY2GC38ZBolZ3zQ0ryxNIQCNSp3dtCRhDV15IsSkyMWCSRGBVDiNpcySRzJFbYod4GZJUyiKn+50MNUw2m1p1UdZBBvgTtSOmtKgzHmQO5dSDvqUKDwdrqcVFDvSXF42WOJcGS7QikKqZG6hOzB8Qv5DV1dVA3XOHR0qQ8ER2RWxiOkzKJFCisBU1gg0aY-N1Kr4oLXCXHIaFP5XmazIsqq365jJe-QHVytpQgYnWGdBfJXzBYolturRAkodownXKaquihqhEqkkeVTtaorIs-0u16Tw8l0rlVBg9LvEQNuNZgaDj3Rg46x05ItTXmbZs5-JA6LPgszM7RZW8ya+PtkRjkkMuVeJSDFsHmyg5m5pHLEl9Rhzvrt1cqalpvU0AGTxOhio1L5XcoGLHsmuymuqgnQtlu+rwp8eRJfG7KGF6sEDi3KmXF5QS4JAAUeqxE4MNU3eG7L9uiyuinyT1PzYks1pupuBKhUcowtYY3KOGFNabT2kp2SVRd6qcNZAss4O69R3NaGaDtvXRorhtmJRuiozShVQ9NOZQgAFJBFnWdNAZLOrHF9iNPVtO-OqzhDIcDqDnWO3pQ7pssd7WzPmOyK4K+O8mq3PrnP2eautBw03ksXdSnNK1HAomrZitbr6Sac3K3N6mnR5KaeYzEAYobP5bBcN96MIYHWJXitRkeedDCKJh0N9Y+jWePi4JgIxMfAPQ2lq8zBJaY9hZh21l+VSqu7-ezcz+ec0aKeZWlL6QzCwXu5MtTDXIzwYR11GQzIpiqwVmQbq5qk10udFoVVMIMZHndg27jBbsh0blv8tug1AgQi6dYXCerIdBwQwL65bSBqIzlgRMpkidShdJouospF4kaCUbfndY210UsORlFJDj5RQ6m4e6cqNtGfXSXD0f5fOso-nV04+Yh6QmdWSXmJBmy8ld5CpraXfA1pyY4+71kOlOPkwrxkZDuh1iON0d7lpLAoL0vOw+TGstpNHKMtiWvEBUQmGI7ulxmdTqU0uNQufRcXnL9tOPFtah2bQVxyc0qxnaMWo23TVe9G9Y9WrJWP0615y5Kd4FaOQK6yiWYZpbSyLh4STyfNSZ2x9kyq3i7xfKhkyXzDrQNHmLzFVzka-7iQHKHHjxKRNBG0j3LPCap0Im5baGdKfPIwo1Lf4hKUfVw4bp1i9QF6WqWhNJps3eByY2qFzRQvOzuaCDMqp8TugaJfZaUzMs3QaXAIXNeeuzFA1IYPrzSEuozXwBMwrFZlBdQ6LFu1RrRIh3Tw6O7CeU9bAz7uTHGmrmS5WRpflH8z-KDiyNzs3FRwl3F5hGqSju6OuBugk3gMuLTlbi9Q9kbXlPU4V4qeBrtwbWC8dUlnfkykbnFecDuCS5XRZWBMcDIJEOq3RuRQkh0Xex0xtJIfqldYRhVowRRMMA08jFB96wQ4p3FzkHg0Lcqg3N37QX77uYJLsdTLTXSS7C1x2hHJLXrkx3wyUvVBqbiw2lY13gXEKeOdMfpE0t2nAypNXP2mE1NMrlZtKFRYcIzEOW+s0U+LCGiyxeDhQ9ipzTTZpMTY+lPnMl2VMZd7YIqMiIHgYSBf3JpNGhtT7T-xB233a2szJJ1ZNHWMxZGsKDby7tR-Y3Wl0R7mmVSfFD-IwqEoxGXJfaUfkRjVTS1gZlc7AsiadwNiPchwEbVBlxpE0hUoaXTtUoi3brEco6HWEby-aH8ZxVZqtVGyPwS4quW077vwdTl8ZBL6AACStW4IycpaXVK7NaT9ZLkd05ZdJTGhn640lcUqOymaV8xtEJLHDE4ofWmke9C01OcGrQbLE04he53NUT6c2ooc2cmgB2drwNVcr7WOGMrIWQBO30pGF6IE1OUWpEZeoa1PRZRVozvpqxYVLdn6sbQP72sbWK9VJZvVCmVOfLM02KdIuiaKs4mtVectUzqZNM2mXTKOMOLbspmiae-dzkJMPcujMqiw35rgzgD3FoOjGupWC6rp996AFMrjiez0Y1M+hPomjuJHabmFcyvTUEUOXBGlgfR6NC2NiN3BSqhxE2myy3ajpKx1On1K8xiFqjJFLBNspWZrlpHC2+KqYiuPWPwkhU45AVBKnOWnjLsW9SHg+NSp0oxUyDOzEHPCvfYyhjrODQTj6XGKbCNPCpvqgNaNZCWHBLTSxcgZyqndhozEyrPL5qz2zaE8vV2fMZ9nbxow2FuMJHR3XuRaR1E9UvRMPSSL1qEDOkrGI5m2B2lXSk2k0BH7cL-aAoFsWs3kZXREQwDnZlAw5a49lxUHCoXcZmUWBw+kyaPtlJDoccb9bPvaszTGSGJJamVWRdDSpVRUYqDQjI0QY-5b0YrAqpKK0wNZmazl+OkYU7bI3el4stOoVMzqQalga5NJrLddxNIlsolT1qU3mvwsaOlnWnHFVdF8kcBqSvpusdBxmV7Dn+alBZpACbFu0SfDW9fvPFDFkChrbgp3QNYJdGQa1H+gEYS6KtVMgEsBrTYnM+cJC9ZgLpiehCYkTSPwz7imeCr1X+qc+zeV1hbSpkLZWOQjMbkItKDxzNZhfFObyOzmSqyaaPMOnMGt5i0mZQZmvVpYPzdUghDG3Nnpu0b7pgWqtYv3mxmb-Oad3846RgKYqZ0N4iuuXhuxwiJeAp+caOeE2YmBMxdKDtky8ySjgR9WHWI6V5w6V0tbA8UTV1+zfDd0htp6uc1dRynuCjfOPvxzaoLHuNCffwwddK7x9YKGB7wDFTZzhLzxAJQkcRVxxXST2aJujUzbKu8UI0MuJEswIyUilGVelpzgEY6x7UdMCXS6pFQLXFMnuu9zbHkfSpakpUwpUHV5jGKOptrwNl-GcXzmA1d29jTDvEsbu4cvhmJtkqGQPIHktsq7W0kiEhrotTzR5qwr71+OoUzKwBWRioVxpu5HDviuFS-gmMUmIsLmuaYRnp1lcyHTPZY6z34Mc9XrWnN1MKr07874uPj4Ch5WYzJXAqNPLubNNj5urDVSE41SoRgz-4W7MGcK6lQqtxW2iPafVO1S35doDU4tYS8DMdL3cZtU6JEWdaLQ94Yql64o-KpxukGoMwxg8qcylNwaiGPOWZrHl6hQEZJred8Oqw1O0I9zW-cmEU9tKC32Jd2OZ86P1xV3lW69bjluZuzTP5tnkxVCx2HMVtquPGba46ghylXpb6WaNGEOYHkr3FSU6p55MJLRYRxtl3KUI+DZLqiLD6qtbyL9ryZxlgdee6SUOvHWfWkWBp7VZCrMG-pXk1ZTpTawa2UWLBgCjYWtIIFs+iaVNOmniofOlOPLEUxc6rXCiM6-vcUbfN0EdXt5SZGMStRkrMWX7xEj2xnOzo+3MTMvcR00RU6l76IOwotQ828A716qT99elOIpMcNxxKzNrZPMttYlES8mehYNs1ICUrW47AqcXiTSDE0rwVvF-SZUKMmqpER0FVMpTxp2oXPdnTK3ja3RFBl3dHy2ILPrOowcKeENKyiPscCocjYycinnGvp2GO8rQjAUJOnwmKu+txgZ+JYE-51rhHMcpcOqVID4lq7W-SIJp40ceHcY8hVAS10eSVqO9a7k6TTS3WmrBQCHgYUBAKLoQ2jjko-VjPYCJT+AkFTKekJcnVR+LfrJ2nfDidde4WFIeTF-5N5W0WaOtGGpuYGTEtaBSYwWuyF+SNUHq31dXLrun5NudDMK-tknub4xRHjG6bBle0WlWDHZ9gxbP4LXYuWuE4qwRKJdyzmzluiw9Dso46YyqHV+FjJy6oZSi3ypmxbqhN4HLZGx3cDesfUGbdVca6Es1JtEFnbIsM00pvpa5X2psaZWVoCuU1q0F0M9DHpvmKCz6Dpt5GXJWdnfBLKZBBacmJaLlHTGISMCxl95qxvEGopzN8o6bxvfW64NcZSnJ2pjIVNCMWej5zEc-w2obUEhKFmUbOf5aBKalKCy0IDVdYimJ1-QibkzTCDwa-5-XN0U7p2DuzAFPc0qMAaqjwR2FkykgcJ1PVYHEG3li0x7I2YcF9Ukj4h28DOrSuazwCZE2n2SvJxw7TrQVeyPKrgiiDHdJBrSbwkXclurdIhZFKXKnCLhSiljkTQc5cQe5-C5JfuuAES2UGKVNYeWKTUiaoViyqylVqWfpR5jTsr8aJrEC3MvF3M0Gy4H3pIMQtar+cpazE8BnrHFiViowazl5nJxSiss4sUKsrs1d-Rqe+U7ekL3MjORtegUay3lGIpB5qcRvGV0jcjE1qnWXXZbLKPLJV3T80t0rSbDdHxFSZaDyNv3iqlkAAk6yUeAcPfc0NcJxbTcE9VI8tnGPNHQcjUnoXD+Rk9NWvWfub0-7pKOs9GFbPyHWmn3zJr3LBiCQ4LEmQZ2repMvgl6pBkfrYKAPKw1CuxVA8GcH3JnAyY6J7fA07xqNcM-CS0HFe1Fr1rDjhpwx4aStBgrBpYvwU7F7mWmQ4u-NgzL5rJMZqWyCbAyw4XysjX8W0Q7S45O0+LYkImjfa9f1nDA3ZRG8Uyc3MTz0gsjLxw1E7-Mz8yjDjwHTWFNJSBF23oQb1hmPn7J91HejJWinLnrM9MTBfWGiajvPZ-L-ZcbZOknLTNVyyHvcspFdJyomTWqJOJnE4a-llNPtUWMr7lrDcyXI2bjl7epTV6Sn-VdbdElY8Hb7t6-TlR9vJtkx71aJNHdnZx32vQd1O-jJurZ33vTKwu9SOv2huY5jLmqqy5r6QyeXNonN60nAGyezzDyeD51WUVsimns7Az3RtQ-aGaGk3cRdocZYKrO+ZctVfnqeADrnVf4gqwbLBZERWkv3viuy4fdCB7L9KlLl4-KH8xEpHXI6Pjo72+-KpUVGk9e8tcsnQ1mDIg3mLkcvd6ANU5hfVQoyDFKZCJjTdL+NcmZHuelFh1xoo9v7XjAQcY+Jvnj5rtWvDUl+cDVir1KTfNFNYHWRoiOEysGDmJVHDU0kwUg0W9DTtwqX02Jx-fJ0g4coqGFgl9XiJh1YEJUDLmRJhUeNwqx+VUL1IVI8AeQKYjWKCjOxBMXqAZROsadAIAmkFoE6wvAUtQHVnrIlR28dmCqSRU8OcCwMcYbDwB1lcLUZwREbyUZzv0u2Ix3OwU6PMjuI3cdUk8YAxSaiCIaUN61vQnXacnaM79TBl9MmtPtgHY2tIdg613nY-yw1cBX2jX8vuTE28IleHxVQ9ifNol+8OsNNBdNQKAR0lI7sZjAIxm0DHX017DOhndd4JNXUclsMK5l0FuxUkmBpraSMSxEP-as3fwRtDdCJoILLYHE9SDIagssRPHLzscpvCHjGx+0G7CdUcPA1ndsM6Fl2sk2XUgxgwvMS9ETMmUTOiO8a8B0yF1H-PnEaCATN0W9sCjUg3TosKJdja54HRan2MOlWnGmkW0dwBMpRkS+hgsqLBOxzFMmMFkOZfA2URs8zaGYyF1tXEv2yCJ+KA2n5ElbRVIMTbN3WX5l+L0SQsQ6GE2JtAlEGj1w6LYGgHEVvDz1h5aUOzAsMQVIPEjR-PHdH85CqEFWnJQaJVgID3KKdEYtfKEB3DN1YKQgKlhqb7jCIPrWyipVLsa7gjoGXP4OwFWbSFjh9PRFPDSZPSe1GJoqvX12Ll-dHywAlmcQ50U4ZLTEjXIfMBS0xN63fb2lNDvfxQXscXLNGXtC0Ve3z4e0UjFs4ccfK0S8vOCyVdQfmGyX49KQnnlscsSSDC5MDjM7GYlSnbpW0lm+QYl6wWAzehN5q1XgNdRqUZuWQYmmBBjBx7-UUgmwllVMgS5TjOKjK19UBlnJhX5GJTxdV0ShwZtqHWIKUMg0VJUdRjLZ4KbVOzd4OjoNUWOnjo0cROg-5G0f+wFduqI-yJCAOLzwcpxCFnXZ85ZEdXIYIMcdW6C0Gf6QNwsGFtgMVHfc6hZwrAlz1sCxXDZgcCMwitmK815Z3ASkiafQKuI2gXWnX06-bQk3lpvOANiU0qI4SUYHKJ+lQDt0TBTl5E3EUixdF7YUKwYV7PWD6xC-Kmmp9iNSinxM92F+SxEWQ2oVZRyfOlAJp4BMP1h5WbGU1+wQ0GzHtDC7d2XTQjOXojrQGLWWj0IBcSjV9DricZSvZ7iXbExD3rEURxD+BNEU4lAHaWXzluOeFwjVwI6p3QYKNEW1Q1wpGj1rcAOXKhGoH0Ijhh0bfI-h0AucZo3u5xaI3Elo+qA8LrCWtKV2bZGwhL0XcNscQz9pE9EWkLwgwnxlNNrnZgR606Akqn1QO1Nqwe85Obgx6tjw7gQmIP5aZRJ9wQh1CI41jNokkU0BG8medBHRwPPRRUdXB3cMqQFiFF1cF6QV93pI717QKZHqkY55g82TY5FuGLC44dqRNisFUDRmio4bCCLBTIu2AxjP1+0HNH1xrvJvGSNZQ98TuJcsJCRjRaPYf3nwJ1VJRhxDAyUUFt-iCJlNZiCJomjNFbMo1QUNCVL0253UXY3JYH7ZPnu9yMXwFJMrlXC3zQ6LZWigE99Yy3GDRAjOkEpJyEQxqsgDW4yQIdZXVCNQsgqtXlCrJablslm5d1DOdxRIaVLCMGcwTYIgJNCwqZ00Wcj18VIldCKshvFbEvdsjRhXkY2BCcN2NsiMj0VF7mNPmTcOscnm7RKFLlT9xL0PfUyZm3ZKM+8-ueA35ZgXA6yL8TrYQS-DponsJuI-wsQOblkqFL0Ep2eeenKd3mQjH90TpQXEd82WKMjHsqNf0IPtalb7hswvveAwjCULI7VNxyeU6kIwDJCdCJIVg1DTfsMNUg0nNcjGcwKN57ZGiJx9ZcWlHQh0WClxA6cOiniZkI6jyhlgo03xXQCwhoWLDRVXQWixVUYGi95RfIcSQjfQvVzpQiCKqX-chrPjHekzmLsJFJmOOo0GUR0MSjEsE6YS3OoecGDzxdAvLLjKFEDC8I05ZbMVmWJTw5w02oQA5Bxb5tUELAS5AqRNg351qGxh35ZPKJQCNgaPT3DNdlUagBM7UaN2t5hqTJjg8YOJKQ6sH9B5m3D-xQ2XxwG2eRX1tmXUYJzo8w7I3gYPMJBmZ9UGLYVp8QKIHwrQ7TNqhHRdQkA0h8WwxTkYV1I4qT4x0MNJmv9flH4UTNJRdj3ztTyMX0c9UOZ2QtkaqOVBKZa4l2xFdDw9-HSM2nEg1EDAOPskhxGHcMItILGQZTGMgBIzF6gFtTNGYwJ8A1jODfnI9n5FJGQUXWMtsMIVGoJUb6LY0DgztCjxu0Q8Uxx-JanBCDnLELABjG0DlmY5CSVGQLi+1bGNN0Qo3xmVw2eNXE3UUzNOLcskiGnF0lWrSDwVQp-W8goibA6iLc9mw3yNKIn4ofyZjNKLxRGlxRLzB8D-MWF3cpMGEmLbViGTNB7F0GKjEGEzGOP1Vi9YDhVadsbfuJficMTJg0FCpcKzTsNib+KcISMCKjf9M0fFlPElybzAiEELVKlYjOdDY3JFguTUnnpF6OqBXpTxJOg3ot6ffgREMRO82AsTWcMxgwIWVJSaQ4MTbVSUiafznJFtKHNltJIaXNBxwA9ACjkkFE5fV9CB-b53XdoE59RLpX1T4lO5u46BIbs-OBs19smg4eMA5fXbInq879O30ctlaIUlGlr0cgzQiK2eCyLoM6VHmNhCbTNFhMp3CCM7VcfY-gfj38OoRPCxI88LzpW7fwQyYjTDUKDoDpV4NQt7mUS1n8TqNTHvioErjG-0YJToMC8Foirw8ZE7YL3INpyJTyzJm0f-Vw9RZWcm1ZPKSON9D06EYK9tY4wLnodh442CYc07PQRRxDBGkkzVccMwQrDLBUnEe8KcUphSJIAsBUpxE0HCUG98JeaPUUuBCdl4EmUYvFUwGOU2RY59FMKgsjrZYxLJok2OyNbwHIxrCcjvZEsSniPI-+hk5vIh8gBCaUAE21Iy+axKkxnUA8iFQcZSkR-MZNQTg6xGQYKXSJFTLTF7sK4SaFFlQRUrhwMXLMfGPodeR5y0x9iVMmp4e8FlmJJ9GATSxjy-d+wmCPogSkN9WFFRgEV8TBI3T1GyTKVs4-WXEBjI-kxNF6hEhDgmdDf2J6PllB-H5yVUZTfZjGkFQur0VQz6FXg3EPdQ5lB0YcOggH0UvSUWtIceVzVQJjFNWNFT+1J62XERAyhJKD70MoLq5fXP-g9UziFaiLdsiMBTCoQDZ5jkoaUnGNEDRUDmSdcwMMHB+ifrZAivih8MmjtMg9Cj1STgtWhVl4w0CUzji4zRDwExncBXim8oCGAj1wbNNHA-cxE4QQkT16BwmkSktWxTsJ7FXECewbSCuDPUCHf01ElchILBaiqkuiJQwr0FO2DIrMRXTKMp5BDyLJv5JGSEEQmHtGEUamW2URptdKwWiUi-ZkX+id6YTivJuHB8w+cbpBdmMCssTOQRU3aK1gvQWeTBV9d0gy5XHsK-TExyMKDfI0WV940dOH4oKMqlwcHjHWHiFy0dM2pSvnCewy5SVYIhxM42CbUNQJ3WljZxMPHD0cI9-YlMbodqMjBzQR5fRn11SnBlSYwHVNUTQF9WRlQHS5WfmNFSjo1fVOiemdRSFJqscvl2sfiDTyTJVqcMUJF3+JxLy1hqeC2oDtOXvWDIjuPDXtCG-Bb3q1PANQNNwdQ34OqTaGbwXHU-BcIlj10ZKFPXJliRPTwz07BY1eMHmW0jTR83dEELdJKItIyEZQxtLqJAUoEJBTmBIgSsNtpaEF2l10P+0CwdCLAm79BQjNEPZFiY9gkYfMdeNIMXiGNPeIl07ALPS68CbFQt-rfW1E1vPHML892XNcR1JNxENG3EuTQxXtV+0CD3atH9N9DZw9-COwkFAKFUTCzvfTUUHNg9fjgU1G-UxOGZcWThl1sl4jgTSpt0HSPsM40o5TWCv5CXEZR7QgGiHQgaRZzWS7TaE1pZ4kxtBzstUX+PcoWCQ+kHw6XfOM4ySsNXRIJDLF8h1iIJF6kvRgIr601CFg03GVE5RPUKz8AKZ5LtNYnWdB9DRUxRTZIv8a9DCTreL1NFZESbiO0IAKMIMdJSNVUX3DyuN+TDdJfMgKjc4GQfUvR1COXQmIajWWMiwEQhc2lk9UyESgidqRLMe53KSEh1RVHbDkSV-OTR1IMu9PzJoIAstQklECWT9g1ZanW+12JFwnF1KjAQ4FNS9yvDTg8xRHXwnCNYOExKcyYyLwGnQd6ckmrRYiJMPMYgcjvR+Z9aYgmJo4DdJLzxpCZ3APxgBJjI3phBB4xhZoacvDIzaGOuRWtG5P1JbtpzPsgjQo0EmSm8Jox5hec8QQjQ7p7sHWUJDesilFeIocSDVdRmFbizoI8k3ixC96-TnObY5jFjIsFG0MIIaDtNdbyq5DA3nTuyXcB7JEN9XMXLkiVUDq3u4RXYzOxcM0NqKvdj2fV2FiOg4rPQiq-IMiTxQydpORx0HIwSzQTBJZPLDMElK3WTbBLZPU8yacTjKp0pSPGBIVRfVEd8AGEt0BIC80BNa1wE+wJFSI0oDFBMznTTlCdmiFBV-x8sITEpVv8fMSlc5URVFRwTeJ9MoYIjYNCiM1BaoTaJu8B5KPi92IsQuoCcdaipI4iZ73v4TVS-wHiWYsdQgwSwwpOQtikxGPq1iuQPwFiXE0HLcTXrUckQFTlFXHhjt8ivVvEUaV8JfZBc9LHHVI9SoVsxYc5KOqwmBGKyUZ7Qy4wTEXnDrF6Te6NXOUzaGU-xe9ePC-2xyAOKr1lTbUAqSYEXJQwgOos0BqVnyDeBET0Jb4wsU7Q3maphQzq8y4gtSaCGDRwwxUbbK-84C3bS8JZhSUVB8mpb+LzJmFGDG9tcSDimsoQCPjKLIIdDbXDJjYpBzcMUHB5XQccHIpMay3gtC0GCPVF2KEdQU+RgdYcBPhI6Yt0Vaxis3Far20p-Je300sNWVMN0suPUVINt8aNKhNsvYrzky4I82vxh126Q0K7Qu2R0gp0DLKHHswQMWXmGyx2eJWFVWBfkTy88Iw9TiI8AgP2JBcQZ0wrRiI99Fi9N6MfM68C-DNCRoKbB-P5QQjZ1yQYxc1VN+VuBPLAtCX8euiioY+PjiGwY5HyJAK+sv4T4w6kzXATtoDRdk3i2uUYwpt5RAHzhYG4-RnZV9qAIyOp2JcPCFT1NbjhrR5-DZUnxOVCxP3salDE0kcxCJM0LwQ0Ap38wruCUj7ZIKAdAKEyMOPzviFUnTMqlgUxXnDzsuSPOhDHSCbFOIpPFhMF5mo15xeSUk9XM+chND1MoSE4xBjsxk43d0KdZyKMSfdJKcIKN41UanBiZXmeAILwJUJAPIZcaZfjZlv8LDHnk2iLKINQOcBLi2daEdWzbjKaVpAIKcqVJWeo7iN6OCdRpGNF9FQMa3zql8vYyR1weGKmgwtCAruP3TaUgeK8J2SaeSnYMqFAwU97U1vBzZhBdyjvJesDFQaZeo5pjxIQvSR1fy25fvV-tZcmPJpw48xZLxwMEvhwsJbI8nDTy9ZDPMzirsYfHu5exbgnlYEiqakUw4MPknZ0oC8JPIsA8ToPe4ohCplcE2CQjBjIyE1CMC56UiiyWlRMhwiGJbSPvGsiicoCwKjtS64uB1YEqtVil11D+MSlDskNIlD8Cq4tcwRIidjPC3iIUUBT5sCzIssLk8jAjihdOtPuNWWM4mJJ3KJNDyKoVYNAWIdrTVPWNN3LwlUMyOZlN+1bsTtHniKtAbwJcSrEb2zFxvVaPX583LFT1h6qbIlRsyFb0J9KcZISl+E-hGhzgScBbqXWFa+Li1TjO6e1MXM7sSuWo4PLZqOjo87TuJH4C+IdySFhUwtG44pSmxU2pvIgcrFSrEs1QAC0eEbXNLHCVrDjpaTAWNKLoJRNOExNMraUYZVFN+JTKwMvHGONIyGplapYCO-W6Vtwv7IbTP-DbA-EgCB7MUxl84f3uIUSED3GJJvNoiUDjza0U5ihOTeW3lvsmMVIpLzAFOvpsuQfNpRkAtJhCFkScISZQYdaEthLSuXKLRYiGF1Xq13VDgmbRH5F2xBiDFOSgfL+MWCT+VigxjUiSS6cZQuTgqZg0TIIFEQTRTxEtejrR801WJ3puOIzh1xB7WwQHKznX4RBUxUGdkkcJkk0RNE07SRWqDJsBsuFMmy9lx2Ka-XLhh1nSaljawfAXqEXtIsdA2RYOhPFH9zhGDYS+xdi7Um+F9aFOyvlaCU9IF0Hk6zieFxmBfPSdIC5suWjFGC+l2M2SgXA7iUCbNlv1VyjjxQJ32FMg4406IDgqExraoSGtSfOjJO4RSHWFDR0wyMq3w2ZBV188V2XG1H9VVCiVTiLs9OO7NuvCEWs1QK84JUzMnNTKxzQQk-P71vqfI3E9-yLhjLCSlFIjKV4A94h61W7B3mSiAxOH23QGRNaIAobyQCSewc4jVjdLFE-XxjR7WFBl4DQMDL17CtOZ3CgxQbQnJDT6pddiwJjynTS9TrifHVmChrdWGRVL5XniiiNUW0VmNQ1X9W2qN6WTPapcszz3pjMjbJOy45MKCoG0kpBNVbw0zMnNBr-gnYyaFMZSzENLVBCykLDGhDfPWIcivMrJw6aO0xndYijrEniw048sMKjbdKlNstHCrBFFQ0PR3tCuxDVCCxdnQWwjLiirDSNVz-TJ0xrT8QSiEpr5TkzY06yQN12JiU3sSdkvKYVwPC7A8V0Zp6wvWwsSYEiVOyNzC3YssKo+CbHKpSmMk2kEv+RvwzjGscrWJTlUQtD1QoCWLxTpsRRrFbJkanURGLGbRQut5f3USLOT8xeQ0v0s0Q+kI1GK2QM94+NaLBZYfjPF3WzOnFRTcL8ZPIK9IhY6y38x54-FmbQzsSQsdr0I-xym5AnUPMI5aS-GlkYGSumqSlHmIkWRYTcgyTJlfJO7FNx3OfVmPLerNGoGsBa9-FOZC6C5iErWNZCv2JCostB7xh+EHwMl5YuWNCCfAUtAj1r-ZPRkdBMaOsa50apoRwxEPUTLedIuLcPJysRbowsMVcfEritQdBTDL5eMe0O8ABXTiW8BEhR7mAKwK0QgdQIoiwxENTCzfDVJVSVXFTsYdZesLR6Qx6NRK5sNWrBShcsI3d1HgmdjvkBmNPXcpmMMnNCz4yPEIlJbShmPILvNKVPgFQVcbUKS2qDtQvruq-vx-qzVIqtw1pHGHQvRYGiGsPTfaaGoDpegttCRrwsup1J1DE+plQ1Zow5LU42g+7JNsnst3MpdxS+ZOMEccaUuGjVk2kgVLNkpUrpZuiA1lpZm0GivJ4xlED2EMTSV2tPx1cMgVFRhqGXnNKYhESRzlCiyihvLiud+Xbzeo77G-xGks-huJVabzAoCLScupwYlvGWmPKfwjEs8wsSoawhTjol8h0TQvbRsobbGHvlpZTcVHNMzH0m4ufixyo9OnNKDQmNDKK0OITEFHfAxQQjNoyF2CpWcCuAOp-082Ul0TM3F1FTIJMoqfK+KgeLxjj0ueyib64n308i-zab12oVPaLGOpJOGmhc4AKknhANgKjURVQLZAJS9z161DRXiT2NeJVTPU3JwjRunUTMvV5szPLzjWcDZQbJCMM6STZv+HPmNCBA8Mza5z7Kt2blAQqCSKlW5S-NpZ80kdBiz6tTgleM2cHv1dEtEkUR3jmhQUvVx1BAuhQYUyokSytGnYcWdjgBdyrnYUSblSUYgiFOTycWdblTOjkK+Zy6oqcQyK6r2okR01JgQiR1ECIMMgidQquKXBETg4ztTobVsr+ul4As4MjI4FGh+p2x+MCM0tCWHYGXfRw6T9HPjg0itE2Im2MmM3Ko6cwNnJh8HnFai17cvF1k68fQnb17yg+p4qPEjLmIJMIgqmwjJs4GivF00E2n45iuarQ8l6qQh3apEyY8rALF8t7xgrQmzWssqo8oOi4YaOQXltdosIxMPVIuS2Q45sUnai1QANYkhSyW+RnBEFvkqTi8i5OIosvqYpXKvDR8qvFuC1cG8nxKqG1UQshp9mlajLlMJDQJlUnzXZTmVgCApKGtLTH+1OVdjGlSBJV6utH-tweIB02IIWkPmLpIMV4h1Jlcb4QvxxWMyhz01qm6vy8-C6nMp4oVCM3yCnxU0mbk+SWzGFV8jZ0v9aowtC1VQTxEJiewOM7mrnxnawMNXFv8KHK3E383wNAb0LT3w9UEWG8lii6Td1JCbpLXzI3FocsMPzFu0TmpRysmmLStiFs2yLtM4tLQtSkv3RxWNtcSgvCkZF+FBWiTc9D4TvYI5FllIoUdZwkxxAQWyr2oAU6MtPCs6OMtesfY0vmyZ56ec1MJ4Qo2q5z-xBFj8bG0JNBDrnWzBr6yIU0TXsp15MWNoJdsJEhlykpKLQbod2+vGf5NCnXG0KT29WMuCYDc5PWNxDHPWglWbWJPbacGEp2aayeHrL7bHxAqQfx+MfjI4FtpemVR5UvAyJtw24-OXH1hFWTM6opM-OVNxc0B-U7R+ubJsxaupZdi8K+pNJkHoNCU8OIJi8CbGpQKqljt7DeqzHJBDf6vrL9tlCgOw9o2iCnN6w8rLPzxQbyODr-knCt109jfxUHT5VwWZ12clN8l4L2bj9INqs56tRrXi93PSqpPLn0w9MhyV20dow90K4Gl2rISen08o06DzOzCX9OqtIMLDCkRjNdmN8y87IwprM5bqcL4MKA22RrHfBLCQjLxQdzRNoq6USyMusxATNmQCFGYjn1I4nUX0URJ7Q+5UkE9WAATXqWqzRtio5Wu7HJY12UqjzdwA27TfrzJLQx8Jg0JYHGJh1bGtZj189mN5cDMrFNvNJFHnKeYBy5nhWM2eNY1B1lccQzTxrMYfQ5Vm2WrWjUzvVOiCa-S9WoqVSG-2kBcDIoPAfImBXjxnpVC6o2CcBs6DAQ9D6-pTKSBxBMOVjRLfOXLxZMw1j1550xwN5VeLQEtTwTfCNhDlN0NQQ0Ig8b7QEVK6n-hZxZyULMAFS8UMyCKguyBN07vONR1cTm7Fhqdy2Gji2ezQvJ3jiTxChLpvFkkyrjXFjbcHGrJcaYdtcZ3RH6l2NLjBKu5TyabwHBEpu2JWWJvm-qyWJm6ptIFJZhSym25fXN+uuSxW0NXn9jKu7vFSjOj7ALxgufGzC4FwhVDwLdial0uwqS24tCbP7WeQg5Lyz6TaaIFV206o9Ybty5jTvepuiI3fLiSaaOsICp3oQKi2RmleOCxi7J+rFhULJgvPXqK0eeaeU0Nje4BsY4QmVAspJ0Cu6us0Cyv429tiC+Xs8JtsBvP48zurOXfbW8Z8IAEamJ0yRsLsLmpdaDSQ-KbtSK-sjCFkEyIV4LEHVw2b548tvm1QO+Bcq95Z0LPlNxS8XAK7FD6EgKYF0lcgPvqr6pH1XJJcVH3552qf6LLlUcDPmBJ06sdksweLLwmOEINFlGIJEDOHwFCSkg8TtMiHC8W0ZKcNqhMVkkx3079t5LdrJ5Wmn304SVpesQJ0tKDeK-lMmKetm4k+3amJSAJYYM9tWXSqJfi5GD1CGM10J+hssLiqJzOlMJXdit7F2pVQN7EJPHV2Nf+fHoANMROLSKV-GrJp5KmmISlaZ3EgE08TA7T6QVRqs9tgazm1FnpNqU8u0y79CB5-sD6net1L7zqSiAfFjt4yKLmCtQq8XWVDWC3qMNR6+zpD5V2jmTNJjYT1rhI3MMoRitBKXY2cqPKLArO1d5UnqrzKq6+qw41cNXSzEDq7zmXYzKdsuwqymO7EBksK5DP8oT1KlpRsJaF5zPp0S24icbkerjtfSY2ClVo78umyLqz6pVgeytxFTGKEcwwtcnEYjPZCmvaC8SoSlzhErOWrK6cOOn2JfKcEQ7UYxKzQwb0ZVKjAJP5HLEGs4W6gloI7Q-9z9a0G8xQ5QlMuvtIto40ZNaCB4jwO8UmGbwN9cW0SwgKZOrb0I+adRRhsJcZGQL0WIThFYiyc-W-QnXY8cNajT7QpUcJDD2xdDvHZJqJoThUtKv0TKHSm9UQcsHfTVhDof48sXxwqOePzfZJ8TfsAIsFVfCoLTNPPqdw0xSW22l10TBT6dOCGZkvJNXLen0JcQVVCKU2cSHoKAeh75mv9VSCrOFUiguj14DT-cEY1x-CEUjV7vQ7UIJxgBWHvk7lCo20oNa+HOrOG2uODHI5hW0torRFs3szbb8u2eKqoFmnnFkz7OTanZbd+GZq8BQ3IZPBrSjIawnYGUtWh4K-W91SL9m2ePlaF1GaztuM2s5LJJ7JesIeRVoc1RUaZtvWCrQxYcWYSdRgve0PE4iewsSsZ4hR0kmkhOu7HTItBkUemim2pOWitmC+QZcxYVOkR2wqveOwDiWCf82hobBnUYuKAU-Trg9+qmPuaQSXUUSkcJRMWsnwkdOEYac0rcZm1sS66iR8BaJLgmqzoFRaWEpqQ8vgwwG2oKOT0NcXQzaIxmuY2SF1SvOLeMUyE5urSwaQfDDrpolnUjdLKO0Jn6FB6kItDtBALRUZDiabypl9cGJvTd7B4ZUuLdOgMV5ZEGZRLLHWLHnhLopk4IlEyTvbuXO8Ne-HAi82cGwv7K0aScnDQtgba3gbRCdnSszJyUAlT0wI+jHvS5tN9iJSMBThN+UIUi5mL7QdPdD+xxR04S2FAx9UVhCHCf4f24ueRphAI6amNrh978UAnbSoSr-hdSQq5MLzGcicjDuYkU9AoadtyuiWE5bSWtBN43dcyivkrKBOyxIY0R7J3dRMstLpGGyMhQ9z23MiJarbxsv24HregMoaqqrJqqhLwyrdt9z5PDNy95IyU8WPL5sA5P6GN-BTABDmzOqo-HFIgrgDrwDXtpqGKUar0a7KycIW7CEGVOx8Fp5c7qPjc-LBIpxUyAGVp5VUZ8Ij1RsxPVddHUMyxBUJAu1DyDYUzPVJIEUwCepILXXTAxTF6cH1CwrIveg2UQ6SkwJTS0J5yAyshPuqAF9JHdRAHmgsYIxG5QsanmIwCHoKTcs3apkdtkWcvE7VbSTQDuZkXG1KlpqqJf1g93RRhnEMsuZD1kxLKQsh3dpyATrs7LBy2pE600STME6cK+rU-G8UTfhk7XBJZm8lNXRrD8koCJnGLFaJ7BvC7l2jjphzgFfhosI2VM2rH07sfjQrROpybodGerPFSXEK1K0NaAntM0lMtFqNAkMIwswAtjJ-OwRSdsKNbIakHVQkFWQDQTMy0HzxqDkjcD9g2TL5x9FcEloxVqGfOmGaSXUfk73q56iFicBEPMW76hNfInUaxyzmIxXwiVtKnc3OtB19ieqFV2ZlDA5k47BWIsh+FqlaQjb6A4rhrRxJShPL4aVkuUrWShGkHtTq6MTjz9M+OO-X-sBeVtGd7GQABhzSSuQEEmgoXWiYHbD7M1VsStKjiP9iy9WVg7Ueva8ZF0QMpjB9M8cCDIB0iGNr34Ib8wCTUxEM2Vn0Krpv0KqVIYsYoHjfCVYgm5I0S+TxMkcfQQlKFk2GeWTk855KRm1PE6gylJE7NmRlfKXFhawwYixJQi4G1dTfia2BKQqKG1HVto5BlQyKPiVs9XuNb2OSyJtlzWgtBk0XnF2VuwVPKmntbPIv5KdblaebvPoFsH5gg0hYxfg6D3iXwfELeFJmhHDVahdv9L8wpbqem8a5XwNa6tTiQeVeoTvPnIUXRyK9k6OI1tuSTW52YRELW92dTJPZnOKk8xOcMXciHW-2fD0Vmsaaql6UFDs9Sw+Qy1NIgXDYbtMu4s7COsk0JVt5qIC-mugt4W0gXgsKBSjgh4MBetEVrXOV8MdIQfNxxWnsjSxLC6Jg9AdVlE+2Yr4LO++Pl8TO0PwyzszWiybvNQlHC2piGMDW1T4e0CVrwVZytzIsS+hsyvS7qElXFoTtBUVsON+cYPTmyc1HHsWaCxoWbfnhvV6ypC72CyxQT6IfLuCVaqE8TPFL+gtU996pxOYe7sBVuqY1LmYSsd7e5QpnSDK0N2SGKnomHz30UvWvg2b4lI-CFVitE3tP7rlc-uQXIlM-vqlEwhRWD8lG0P27HjO+Cx04puH7vHawIy+x8Ab42jE7QrupNX3okc6RVNqfquRSRGQupXFICp+27NetKhMaU4EqRAAb8CGi44Pi62MkdL1SxST5XGJ8jK-mgNvhNmXJDhVO1guTbOaLGWKM3AMaTQgxp5m5xQxlvygVBkp6IOxW5onwXGpMekxfUOI9RstLMJWjGJYxFfejfCmLChxFnRi0coEzRs7EImzBBqbMbQ5sx0l7Yp3cbEYwFWa8S4HgmpOeyMJZyXO0EZZyUQJH8vTXwwMM0Ie244JhEHz2TwzLBQwVLMF8mblx2GR2+pO9B9otzDGW0n3zDUo2eIa6U-ikdKmUveOWVq8ZowxbKqjOnCEhKsVhTlL8JFSXxVuzHi3yxCxhczdQDPYmMUcJrziGosOUag6D2SZJUxlRpGfgpd6Ib2sXM-alc2HRjHDoX8CdUTVyAZsxyilOCK3CQnZJbmmt2+5f3DSjGsNXNCfjCmsdTEHwGnU6TXo6LQoCPESMT+pC7N5g9NINKlAob+xNCEIlXYVyqD07K06RkfadBmmGMuignNomvI4iY5YQbdyaA1fJqxj71xJhmlVRUZKmgWROlCikyvPcjkj+0pmjhakJpn6IdX1KdjsttgS42cMvCJIqU8Os9EIdTLF3dXOs8JGpT-XueFWmAxsnayrkkVxpWr6l+pCFzGoGfskel0gUeKYdMLzYX8vKjAHrCFWDuKV7xGVRL5lpWMbWlPU69jtRb2GJNy6EY6-J+mVRzVhb0-WG7RElxsHTp4m58BqYmXjSxlM2D9ghq1mnDQ0pXQE-lqt0BWIWcgf9svE7ShoHR0dtipl4RKYY2oZhhObwnUB+OIj9jbK9FlM2NChU48YxP3wtKCBoJQWa+0X3xHq4hXrE4Ts6cdW2sfu1AIJ1DAnGRcUa4pHEHc2cX-lJJDxBAcuzfQjKl9oiKnEhnrBpALwWIU8I3P+pD9b6RP0to2ya0wXNb1Fn1TaQkmJJfFq6alQtBdOhxJmurjs-FGGEMiq5JRbAf-5cPGlt-4yJxexQHSlipWuksVjBSYZRq-6iKmSpsuv-M7Ox5RTpKKXuunRQhw1MjWB4yYOqL5R2DFmCAB4+ca99FCpnHQmyVKqSrDWEVzRopkuNnzoBSnbzAxorezF5FHmz1i1RDMKdfu9QfQfEk7-DY6gcJZM+ht1dupIIigEhSK0NQxt-QfKO86cYkAtX6tHNf7qHlPMnYic8fHTM7jRs9FVoJuXsi1pV0GsfPrdV8FN21+DU8MJpb6BNMZrtxTrrhTMJzqH0mUUhqwLRjJuHTMmXZi+dxTrJyGmJK7Jl51jmKUmTXY2norwhG0YNPwjk3+UK1jUJ1hBYntCGEp5upzdJXstzzRQhgN6hGpLZKpsBsFKsQX5PRhISJXfTNOyEAEjei2rWsSqgyzdZtHFM3S0rKorSmnQfGE6jJWtPyFM2jebGWmRgeL+cyGl7p-NmjbQpTdie8L1OHT8UKznGIrRV0FKwCatjAI6igc2fc79MjF+KoVdPG-bw0JOQ2bpHTLB2tHmsh0lq4hPED1bAEjtDTQOOarYqU4NyhIi7mptdom0n-Xh2wYcDKjBYTYKU1gqyI2gskNJee5rgrjkN6Pz6nURJHSkXLVvCLi9MJD4yNZp0V4XVJlFD1E70UAnbzZ8nUfzjlXfXSnFHFIaX-k7YZpzrdFxJbdfQOrcjDyeR2plNXUlQ5bNOx+1pWJUfA81RDIckpSaxnxRCr2T7lr5iW5kaNJu5h3JJa1A6rusI3DPO2r7DWVG0Z4erNrp4zguPjObETbQPCzNdmhgb2WQiz426IU0boxIlxUPow7mIBiuMzoRDIuv8w8tjWziZDMAahRDeW8otdH-jEFdGtQOZj0TJtbFGmzya++uvN488dQmOcTq2X2Z52bW8Jl8bLQEkhFCRLavGEmZi+Yi2AdzLxDIkzdWG+F8aN8jJCetXtKs3vHfYhaxi1p6KvoyRCyhGGINaWZGk4Kg7PlQD50AJaoNPeiTiEwsrvkH7m5j53QUEFKYvBNrFr803EcOc5R+LiuN7Zlbc5MtINShZ3uPITr17IzcxkA9Dy2MdF8ujA7TcttXn80iQuW9Y+cbdY4kv07Xkho1d7B0W5xmGnHx3IPCHgSqU3D+lPXdBtTv0GnxVSfWNEBdSOJozWPyaRKV9tN1K4GnGTLkzdiFwVbjs3C3vnzUNR62Gmh1XG0QbiaSyT2DcmTVTaMrkw+JiFzJVDwVDrJRpT12bJFRcskw0NxTx2m5zyiwt+8FllXsDmgjJak5O1Fb231Wpqf8yjtlhzzGtTHrESsUhUKhp5aWXot2GzsXNEODj15DKmihZ7wkTt1hfyq9WXG7DI-lEp5qszsK7Vkvn8ASErg9U-9cjxMpQOVCi+1YJyR00JAFJMbTtUxxtBnc6MGwkeNbSZ5URtKWkLF2SRJAljeMYOnD3tW7sUBcqqNYwunPXKK5uUgFCafOlTsqy6VjrSRe8SwGnfQu4I297cgbRjbm++CsMO94qLm3pjNz1gqp5y8+oUVAvJ6v00IdB6dHUiwlbrTs0+LXkcI-iN0vzQABYNsbRE2r4Z1xmcI3EykzcKfM-XMFrwRsl-d8kOhykLbVDzK8UZ8Pmty8fQjMy+RPpqsyBmyhPEIZ+eGX1jBNzdeP0x3fvk6SwSPrEAFAZGWhu42paofg7+UBTDDChh510V510KMrvQeXQ2gnbcNxrH7KUZ8ktaWS1kpayPCOIpvCaT06Ob2XcI25RzsjKgpaFGfS1DzCFZd1vqCXWLMbXAP1hTut0E2pTxbDHBhIzBY4y8qiPa1FanQZY6OZePelNNSXhYpRK3MExz1zmbxMJK8I6-cdJZjha27QHq23MskeLRw4Hi7gqvmGNocoBv3nQFDlGdJOE+1k2l8tRDwg0r5D8RWxQnDnKeP+9vYgWbcRT-i8Byd3VwBnzeNQ2SURUX4+B3fXVFLXorOg61VtVY2voGOQca7LUWZfYoPOcLUmxytTKOYpypwS5dlJYJIMwWZQOMF4A4Q3oMJDd+59pGQ5F9hxFFfJ7w2l82Ex88FUNuaeMNDAKSAB-ucYT-yiTn96Wmjgdf7DorgVZ4sFGo7HKnchwxlTW2nZdF2d8+tBjpqN3fmZoM2EN3c2hZ4ZNAGWg8AdCaRtd4T2xDLSEo2G9FvlxMkBXAXK7IyBYuJ1yMFExu+ZgCUMLFRN0dkd0Ehbd-20HWx8NZCI-aFTmV0GDnb1-dZha5ZPDMlYcd7lRx+gbo6mBymuYMQKtSohx+yXsZ2bSypmtUI5176ycHSnSQra2xBdTdYsb0EIheom210cQT9RGhRU3rq6Ju4OZOtcpQJUpAXL2qDC5KkNssSWmr+PmYz+Yhxv5gaos6NUaMUkoGnSQ3KcBOQ8WHcwpoBmk0yalgkyOlTneaN6E62qSeT92+qRvJmCPDYYKlE581Vx1yeeUkdtgnpkLx39wijyiGKoXTAoy0U1meoNyV3SYZfN-MkeKHKZPVGHtCAmryLzpL3ZmaxdVVDfRJdN2W5W5o5hsKbp7fGIib8xU7PVL30bU1M4Os4OIUMwFs9xouPB8tclNK1ptyO8eY-PPcpE0OmcJWV9K45UNYHLC7EJwTXKjUpZGOUwD9iSINNXLLeumMd07S8yqdcLCqytO4OcHm1pOcmhvo0cl10oijPC+-bF3xKOKANAmcs2OfT9ZSYvzwC7hTtEbGkL0SV8pYCKNUS5ZJeNTMSSe681gIhddSRMEBdMfR0lshPVHzVNeoyRH1SF5vbJmoYzEyWixvFaLir56D0IcqcXJyv-4XmQ8WTP7RqQw-OzVehnyomGGbh3T1u+HSoO6u0gqtZkKOlDcDbDbDGlnt0WXYw8OrMjE5r152HlrN1HMHIsu6ie-FUVQiK20tHdBJMh1C64oqIKZQxzopAmNbNISgJmqHekuNdk+7gn1e+wvdg3FTs1XdWFGJlDvZyOpKQLRvedbfo5DW0yMdn7k8+cHxCj+yphdIRBYxzny0TeleSdWjyWcjBlOuYk4fk6Tlk49rq6ajRl5F1BzbjzsQlBVLbeFroZWGZkUzm06h8URJgTpdOuGSsPOvvx3cXfwMzEiXw4NNnUBC3uas6dRW9SPiDmXvCJ2mwjV7L7LhURvGsNIQUXQspRfJ7eRMRiqOz2AVokIMqYVqKpTAou36nSr2JU3jZMHNvyD204lSCJCg0ImfNMOjdejwEQ-XAp1HBaw+FvvwzOtWMgnNoK8mugiOaO8aqHNlKcVb+U907RkUaWm5wWPLkkcVG6+hOufMMWgYCs0DVfaFWA03FlsuAngLaB+A5gwf0kd0onAXeVt6q54WIv5hh12NSKm45eNejn6Ujj5RMZRVE0gqzOK2QqhZRzmIjlFKzT39Ucsdh53z0IaqYQtHrItytGCnbkjtCM5OE1RRavy+ZfBv5CqbbGIJMMoOh1QkuQpm6GSzs+mMCC6e1nTw9drpmMDtsPUo1ciaMNb5PPrc1naCRUMIm+EhadvO3RewqisCnMyATiUCqc48tyDwLN4hlvwOaDjw1oOHdPO9S5f3X0xBbYu11QpzlUl8I1cGC3PpoMMEIxp8qbqSO9js7cO7o9xIe3fBcE3uXlZautscxJV0CbgUwINbesbtT-e0KTRRkUe-RkVeDiOhwHm9YwbyizJRmB3aO9Fks4dcHemJG4qTVnKdRujdmHcJu8ZvxH-z23wDOXBzEkubfm9Y0dZQiLOmFQtlp-j-OAhq1YtOhsCzjmux81m7LOz8Uta-XsBQMvfja2C2dQTfomjn+i-yxtHfAGfDVB3MmcBE1tJZnPtDKvcbL87Scfz10G7R7SETqpxAJPVXcviE6nnHijJTcY-cL7rDUDzbppk1FjxZ-jAXZcSdoOhGxh1r2Rx8EoKW3lSnGqeu0uHse9XQcZN8mFpYLuS5xleWZq4x78xNXvEuLBhmb68SLlPg1ZymOwmvSeZhDMNY4u8NJC7lWyKvHn2XQVt5unG-m-y4aLTaPqpPANlXTJkuOk24rddgptqPzbR-EmuGMmFgIT0QSSX3oGWrOdJZoNotDyYRsKMdL4VpTxmAPtHRmpL19HDJdHGWOGVemiAxOW2K8mufQOMLunImmP7Vz3QhdSn5cjHOy8eHGZBvIyj3AKlOg8akVWwdmMwqzVaQsvOVAaOdPBFJq743QFTHZQIsddXX4UfoFQ05VIrL6FgWfIcuwp11qqcRiV3ay2vCL90jOUlK6m8+LpOY7uH2XatuJGMUUV5fyayXpl6F2smnb9VY3i0uiDY2bSv-6h4NRPgt5-hTJZ0MZm1OyziM3EJzWCWKhvxWXgNHorMh26m8VnGrQZuqY4hTfqnL-hWrKbCRHWq1RFC+dkW30GRWZvAjIWd6bLMrm9et+V+xJP7vThBer0L+1haCUrCbib5Oi26UyDykBDHeXQAhDHu9cmua6P8w1eyQ+IWx1u7Craem8zNXjqjo1ewEMIvJ8hDrKoBO2rveQK4KizsLrtYyglA4iE54t97tQo1EhbEls+1z1LUJ6uurht3KssTfbYalvCNbP8utfsICfh37fJGU0XU06z-1KF7HvbH3zkltncLV64w0L3jwqwmmNOyNfTIvJT6LSWOkeIfScFtc7XMXkoxJXKEgR7NnP4kUkTPBbcxlys7BzWz3bmHvCOusuh7pu-CeLphr4vYeH9bNEetEAl38Pc2bS-QDGMNUq2nFnbeVliVihNCayLT6JjWR8yVmrKu5VMmdjRHwNIkeUCkHsGUD2g+UcVDLY2BM0d3RBnYKvzBUO1J8dZCbp0HN0JTMfNKCARU5l3KG6h0haUE2UIPSW00CHU015hw8acTFLqu7kr0vy32VRTKK3y0ytLK3q0irbZqqt10UO9hysciCfPYrDFmpTNHCNRdfT14wc9+vOk1Hml8jG8jZ0D1dsCy5grcLnjZ9wINat92NfZY7xlDUnIkVeFOKluR6dToQ8rFuczBp7Z5bODPUVwO9ouIBvzXTlQiOC9dBU3HolKOol5WnSYYrK+kslti7cXIFVpGHUYzyTtXoa1CMc9JNfSnTNgHKoUqNnNC+yRS14F3GFSx-M1R-XFclR+JVucDV-FpWAP7i-xmQZTx-nQjEERcTgQInSYCsOIjOAcqvusMOG+dOQTT62QClGuynOUJnLC2A6fanVXqoWnBhtHeGJiYJf33ThT8YfM4+qSQIlK43E-DM335xSuxZyhOvdKjO9xE+XSPWFBCWDeFnWpKqT+7sr1SiMZrQNB2PF-vuHg5Qez0Oj6SgvfRRBKWB11gzl86WjDMe7wnCUQaWcYi6duJS5rn3mWblj+7qVPCJ8f2IncmST07QkWK2mrQgYi2X6Fd2Zp8Ek-+AnrFb4tx3yMqoHwqxy-35+rae6AXRTANvJ0fWsKjYRuwgeU3eAZTp48+CQc4StBS28m+29udjCd9mKAZjSRdts+kkPjXqHEUK0alixx2qQZbQ54hZR4hyLKnLi1bUBeooawtUXbpcU1BbGkoNcdXdFgdXiJX3wwU6bnGAvsJxXO-TFmVTVOmPAAzNHx9UV0WnpTTfj0sbd60bx4FYqr2q9NYIhSIPj0QQpWUjm91A6rVdsFU+NIHslF9QSEIpPNlLaBP82F48Eyr6kGvlHrW05ny5KMqvtBCMwEG4B80-67rOccc95yKDnBMoVycC5g0aPzpmMH6unTlgXXQeBfgGFXuvSx+MOBt77jIfrBcXYp3nFYA2E2GTi-4ceOY38kdUHNgdmS5p2YeTuON5J+vPki2UMVAbx1t2e2b+X6vcKjVswa-KVwmfEeP6fUwsSH98tSf2Iczf2YnwdViaBEDM+Yxy3V3gDkNNzfk0zOihrQWiX4rhPFfuiOa0l58fKlGjSSWgn+1kEO0KOn9dBTxVCWZ69ltXu+2pDlwj8dh2gJ3Z583mWya4YMfKjyp4f-Lr9Wwsi7TUCg1iAObWgsdQIpG2W9hwTpaR+kemj3Ot8m8J-jc47PRlWhCaBs8KI732P9ULIXyW3OAJYi3z4IBLCpwLaDv0r50RK2bmfk06xAotsiokgaUkuwxQhiE-2HUJmmX41XHEIw2zlEtpAGwG0UTUl010Gbul6izhVkIqqToI19XOcwj3og1FRyid2E2I+USpMcpUB+OgA4q8nHViyXRqq1ZxfioLBYiKiUE8Zp3qkhO2p84InmOmEnc0PtT2i5Ux+G7mgzelny040hElQCd1d+ofFVCxpzkGPhV2Ox026ym-Dnyevx9EfGA5Mz5gki4xXrEbuloUKZRbuwNE4IjGGWKZ0yLW-R1+cBf2yMtvT3uP9jxMmskzQxcwtkOxw4Y+HnuuQ6E-AnbiHQKf36+U8XiwErSBuR5Rd8v8RSI4nHT+HyTo4dmgbm+qAG+Pjy5kwHEqEX-SguZfAkYYFhwBhwTs8hi2VKVU3JiN-Tasd-QBSqNROuTdW0id6Hnwlkn0iNtlj46pheYmGyTIZaTTIjZEzITHx9KnomywTNU9I5HWJU4xy2wQtQlGWA1MkjVgnQq9wXorwhjsAeDzOhwG3+AKmXExMgeGB1X2knIy4IfaEiw8fAusrgm8sLd3JqjpCTQKFx16p5Q-sRf1vcNuny4mzwa8RrHIw7gG5+c7Vfmz3wgWuMRuB1uiEWuTERwFQMiwNaE1cOanbuRALZu1f0HUL1i+BDHnq+dwJKoQxAbQITBNoHImTOJkjvK+114eqxwnM3wKqMVhRbQdzFaoo9WkmWDnqubNyk+47zrcjhgeKfnxTiAAxMIuxCI6LEgykH9V5OAmWnqQeGkwtn2CcaHQEot70ueveBTIkzHLexF04WzqBFyPCyFEcKmmIlBQZQxfVL+131wGHlDIwKHFvOoJ1uUZOwPi+s1omEIOECqgOFmv62neuKwk8+jxyIVEnuOLflicHLQNqIWAMwb7FQcL7HTyKLgaw4fQHo8mH1cnrgWe5Phg0xMjlsQWX7QAqXwCSfFM+rWUfkXaDeGRLE+GiQikoRqFu64dWkI0s30qgpCzEkLFNI0KQNev50ziQG1tiVNDNk-hSncBLHuoNUWfI3eizExMjAIuO3Xk9fmY22ShHGeSnaMnRll+qKx1BJqT1BauDbsBuU7sjPQX+CPyooHJwVS6hCVSy-Fi+HAnUiuTne0u7wTYm8j+ISrHzkGdgVQKjh6sTo2BCoKRfKeTh0yJAj0yLQh6ORxWJwH0wtkA8xoiErkoiqzHfklexyckaDMOpZUycm6DsoqCmnIwrwDqadVoQbkS3YIHXSB7UX-2nUSVCzDhcafJRFqPewRwiV3BeQAlB8U+l7YObFn0fHH3op4nncLbHEk2qHmsVGA+YM+1x6CIIGwJ0ls4BOEoozxlS0fUxN4P4kBmXo0kc7xHlsWARuOFogwIgEjsYvgDUGOqw7uFr0qO4r2sy8GzXUgj3Nm4R1bYFcGOoxklXm3jkieyGWie6zgREoGVZmspF-4cpESMerA1Q-0X4hpt24eoXXRWA8QO2GBwY+51yQO9aHEERJGpw+CSvEOV1S0CmgJB7yWD0PgPMiVsgeuiK0taHs2YMNcztaL4Jz+UB0kGgrE0mVbGwwy8jMssGHYsDQLT2roDAU6zHPqGmjUYDTDP8Y8xtetckGGH8gNi9oQJmkiTkq29B7egVGsEeKBYIHlgjGjqh2oavWn25iQxBKx2AObgMACDvRuEEkzRiu7Gi4Ir10OeQSxkkFhGuMtiQeKwgv4cZ3ogT6xu+esBp4NpC5yATWQO5PXNUSTGzMt4QzWpnSzWi1B0Bp0z0B6BW44DnBjkrHGGhCkTRmnWE6gYNHRAtdnkhdE0bKnwJse2WFdo1WBIkWgI4YmHl0mZ3mbObvV-4oVxSIngCrwFOGHcPqGewrdBg2cvwOu9VVIkRE0k0wqx1UIVTcMSBAS4Imz7YAQWp4rBlbWryl3KsHT5epFxTUj4KJ2pILLOq-D+wI1BOuqgOxqPAikYBNGh2acXp8-SWeWX719KuvUC4VRRVOy7DVOhSWKOIDWVYOjwo+Qjl30q+HAIB2E2ECKjlc03G70mhl9cvaHqoTAPQ2vbzJoQQ3eMlUz1UgRSIadW1q+55Xt6md38wHFwbItdH2B0U1oqzeBawvUEDgjIG6wmgBlhxMwJwnLDyk9XTm6yzwZy9CwRUiASC8DRGvYomWyIrDktwRoXw8kR3bQXoVEqAZhLc73WUULhR-w+NDVa7UQrWB3mrWfrU9OPYLrSaZkYsGMJHK-TxdWWYiuIfZAKkQUSA6r4J9qcS3qwoawMs0xERh4jmEOYOww6OGhG0vdzZWGvgAq7OxRs+UXCKthEBodygk29+yGmNfyhBK+RyOBQzyOPzzV+JtHfAGoK8ANxgeYvkInQxXG1Sg1GwoI1FXkO0wbapsw3UeIzjUwPDlErknJyBOEpydU1IOCxkchBwheeCAWt2pYIYYyuiauBkROy4E0EkozktqPqB5w4+AXo8h3ohs6x52vgj52AQmySEuXDQlS0sBU3mDG8TwIweOGn2KREK2HVTpGKH0g2NaQw+K70GoKJy24HYXhUIbx4s5QQjejzSYBAdX8GJZ35s2DEiWUPU64Jil-SkRVZBUg312T1D00wqG0qpGzwoC-RtQbmCCyOxFVul11bcklHw8oDXfBWbT60SjT9wbtCgczGgnYRZAR8hTgz2psW76zCT7679QH6vjUGWy+xfkG5X9udRA6iioSAOE81gs3+GnmU3xfQMXWyUJk28AMlCsIKOGRCvoRoOPoOT0F4Pf+H1FIIiZgYYBTzkiBmT3E93EMcEikNCGymaK7CNKEdQzAGqwLvw4UKWISdnwua3SCw7ahCwBNyL2y7GUSyiSkI1UMfymi2C8H4mv4hSRBicxx7Qb6xp8zP3Pe8ZHZ+VGEpGlaBM+gkjDw1sgSW4-xdq-njxIcDwZhsKQHM2Fh3CUCKv04MOY+YzFY+xsV+K0rHowxGTG6rOGjwHH3kh1NUPOJhRkYcLwkC71nyCjzXf49zBSC0tQ8kFMn7YPsMyefNVChY7GgkK0S9EZ12V8KMPTUpPCMILdy22ICW3hBhS4BPnlzC-UMoG5nXjOTmQIBgPi9efYPBiiS2iRul0DIWtQMuqzw8ok8XmRJQJ9ORxGouY72TuinBySTSg7sppz3e0rHeyIHXVQeFjohYIPDWW0l4CYoKQYVkmsom4mcKiCT+4EhjVBHDHEBQ2EFwEFFCwDwjxYamHm8KtWyhW3wpmimDsS1M19cTISsIGMIC285x4sH4ldG6GG2MWxgtSAiLuAGUOIykVBNuVOgzcg-GNwuIEOBjbBT4q1Byq5QndaVQiCehMla4et18mc3HV+UHh6OpDnTYBZWRU0m2E+kjh1oHHVxKEkTNOWYOSEOYOJSlbVDaurnOEk-GqsRiM6Y65DeI2FA6CP5ikhfMyZ+y233B8tUPBStWPB0rkOi+kWXSJEmORp+GNcx8nlG8DCiiHVleMcxh3GNsM2yTKAdhrYOmomAL9oEB2jy8s24a8eV4ays01+qsxsEwjSIeNrT1QHjnBIpWnTI3XGSarolZyaulDeopXGKBVDBYo9BmKcCyHQXYj5mjhCSOQnS6K6Y1SE5rTsUyKSyE6HzyEK70KERuDKhunWXcjMm1oA1SO64xEqEqxA-kO0Nf4rD0RWcrEFwSpBlUnm1m6PmwZezI1Ncu6BYiuxgX2MrCieV4xieCIjV6QcQWupOmy2Kok0C+T1UUuTiNBa6TCsOtFXQmXgNu1N3k0YKLCyXYhRkqQTwGtSO44ZXWzkszg5wNVHFoRSNH+8ShnoXokSC4OQHinThGqT6N6cKaWgIEHwzSnPwj25kI8MCqC6SfVwcmFOCcmlKQk+Zt0dQ-mmnmfW09SMAzuI7ogQKcwTMUZxXco5+2yksO2sC5eTeOTYQ+O8kNaRIUMdhcskne4rENBkf0B4vyMZorNWLRE5xAs+tn3ORhWNsjJWCcHiQ1wVA3ns5E2XCeaFXCDgxBoOwPkS3pS2ixaWRSeZHGOGkTUSvASuqNMNJUY0xGk71QJeQL1uU-iW2G5TBIOqWziB6W1lIPeHzsuHnHWePWfWgAi7O2J2lmhUjkYPBVQCpBSr4w2kVRolCky1OAoU9Fmw6Vck6g76EoA5ckYsctDlQk0GcuZuBHk02S1RECStcuqIhRd0MxBuUO9ax3ELo4dyVy+DymM6Pkix24XFowWEFwJL2gRTkJDy5znsMR4yqiV9G24ORzGe7uRzUrzE8AzCVjwPsIUhPAxdOZzi+syGxbE1SxIevhTdMDyXH0XplZmpXE8uEML5OAeC+wqCih2cqNXaTrF5KlKmQm4tQFyZ2AxEOPyUhuT3yo+T1gmSUjPqxkQ0hRaCFct12T+gQJs2g+BzYNtVuYvUH0hYKMMhYvW8BcSXrmknD9mwNw4B00Uxoc3QxRQRBc6NZxLYAmCkYesI0KUBCPaWliwskJDnIckJ8eHxAIEh3Gy6velfyU-TVSlYN6RC8Vs66RGpyWPlckxxFN6JXDaxAmR2sy5Ba4s3CGs20kRaRpl206p21Qgyl1SIsMr+T0Ta4ezBUMw4OBmHkO36+tBBOcmL+RLaJrh+OCUelwK3m8G10qI8XBKhQMF86aG-G+yM0uHzmO40elNc542JUVtiwUw9F20smMbwRO3SGqtzrQGoK4qOu3ya7SInM9F2KakTTaUwoOH6rhgo0JtAi2jSP2i69yhaKDG1IsLQV2cvFv81cS2Rdpi7cJxD+I93CSeU4yl623FskcBT-8ZRkDIP8JPs9zn3mHfUz2l4gWyqTS0wOHhZu4IlPmFdl3OV00soPUm42EZxyGwv1bKSuNdAkzzd2ovkZmJWMyBeVTpR8ZUyciZX5EEBAk83mKYqkZAXezBnn8PYgMB9kn3ItxEII20IA8aqVuaX-wnBPkLS+wZg6BRmWpMP5UORuXzhaPFjRGMEk30MI3gGag2VGAdQkhgFRtOdaBAqRK20u2L2KCYfyIxEfx3E27S96GtlXsmUMKizaGjwo8ilCYpBKxaKzKxAZVYhrbxDKyuKEGCRzMC-V2wEwuRD8a1lIqQ1S3qRoJTGaSMoovB0VBrWIom9gxYRYvSSICV3-MdOmSxFXgxyiLSt01WJN+B5GS8JOIJKFOObRoW0v26RBjkQUPAKJqiNcnIQUYoKlA++Xi4YP5TN6GOg9W0xBGkWIK628YMsoM-2EuhYjXusQP2GPUxtau5VjUUWGzQ48KOU8dx4+aiWNRg0h-wKuhTs2+zkiMLB0wFnGF4w8mfsQsw+6Ab0PG5wh6ifhFAIJ1yzxvgTw2GcxMiZsjuuZkPMma2L7QG2NYu22J0wu2JjI+2N9mvyROxNBP4S-WmIR0EnRRJh00mG6HekomTakv5TsKQaQJE2unB8nOLIWSElQUGGHmIwhOIEYBGTK7b0wMEuJQKOBnLQeBgPoJ+OCMn8ghY+QR04GzWoKI0gJouxh6O5oLEudhFtIkVHI+eiJKKkyWYUdYicRiRUbaPZ0lwjj1QSt6Ee+nnnDcN2SFOK+WxGimAphZQlEysyVjyis29RGv0GWfqI2SyM0sEeqHUsWOFLQ-MmZYryjqm2v13YXRKrm1dhawTnjHEAWMwxtEWWhLgNh4Vl12wRfVsuWcgbIsIXq8I-w58VXkWIIqDWCIJWV0dYlYJJGLuA4515yK+K2IFig9MoyKFm3UKzMzPj6hOTx5uk2Ide7qLmS0MxqJpgjqJbvERm-qKaJP4wE4B1Eiw7gH2AwzgVQ8WkzhaNhnWfi3Tux0QIE3eNVSyhCBsGNAlQ4hMpcE+B1sxamm0Jkw26nUPkhDjTcG-4VG+0GP1EmND-wcoLL0IMkyqn7AP4NhwMKpHQZWNwTdWTrGOu2IzaG1EODUARI6Yz5ht4tGjQogvzBGvyiy4BAjAw743+oeqBzQhSl2I7AJwsznhGJCtSwxpZ1H+oZzcmYyW5CrDUey9PQ4acC0YGlehCUSC1r013SyGNmg7RUqPe4xriYE-7y9cXx3nu9oWmYjzxcq0nQAKs7hJS0iwsE26LnSaakawzSyCsrojV0oRhSK-7370uYVksaJ2m+2W28Wrr2YBk7TJqPHGT4VikrETAws+EVTaR+GPgk4IVfybqD1orbR4xObH2BNhA7eJjBOB3+LXe4+PGWcLQvQ0ZhvQ45F2Mcb1VJFJyv+JjkzIpPCHct80iRVDnJm1wJhBxfzhB-mDLJeyw3GQEk7UPwSOOGNCaYvUWuk8iPequSTekmhGJat+MPxHbB-KCh2SEkG0uMy+OawtyPyKjTnMUA5VscmGDpEvZDw+4ymfI8uh02WyPR+QEiWU8rBsYOOGDUGR212UEj5aFRSVWDwVCIL5ANhC2QO+yLFRSqOAMY53wGwyLkD0d3ypiTvXu+E3W1KueDX0yumQC1j1gqFliCin8l9EfJOFW3vWhEDqhfmKkT4oItD30JbDZI7yLYa7YgF6irkt+t1StaL5PLsdFglWQdX-yJd2lWoGLLO3mHAOpmjVwBuKZi8JEMCmTHCsA9HvBSB0lx1CIOIPGl6oMVG7Qzfk3K3ZTIxrJQ0xRqAfOBfDHxWL3zJ+2yoCKeETsmzXaSCJO3kBQkLsZORkkp+2VhRuF7JvhCnh8bTWW0mA-EGwh+8rlzj+0WCM46OA6ebi1i8jpA5YJEQxmxC3eGeshE4WuyGStGJpqZSMkiD90hCskUTqTT3zc0KxLuRILGxtXylecKIlOOlAHQiZCi2Rdyz+TWN9McUSnkhH1M02JJCiqHi5crjHAuB+CMeiK0FGnsMFwB8QrgX+Pi6u5g6SogOT8rwkm+zPmUSsJLkucBWrI8+DTE67Sje3MSKIEKiZJl7C1IlBgSko5JTk43jCIgpCFRyQXTRvhnJi9cMxOH2OJcRCNE0Keim2fdzA0avBfGOxPwyZNCow+iRGYQwl0kRihJhxeMpBBfS4R3URJ8tGTwa2imBcsr0PESWxr0xDjr0h5I9U7PVHCoBBLoLqCtM3whKkTAl7CUA2AC-BS76FTCEKNOBEK7sL8GAFGNopLG9et3yHMvMKbe6rTx+keQHCL0JNBEChOyIYweOloLfJZ3yMYF32RcHvh08dtSHMkXkDBhAX0wF5PDqRvzbk1ASHJExSuGwHmTwlPmuRPtRzKJGFeKb+LEBacU3xYrwFERONh4sbmoC0xVoCvrgjufGg9KZNHsKef2WhtW3Bp6MlskjhlgcyClg4DpP9WPL31wDexTITewyeHn3wEQfCIExHELMZHCwUAXw40v2S6orxjvE4eDiwF2yvOh00nap2JDxqzVZIkhEpCdOx50sNT7mrzANQZJFOI9rgS40+gac3xXvOHaCnWNhFcW7i044NEkDJJ1ivRVahZp-TTlxXW13x3cN-0oaLmBxxF0RUzkMwTcLSB03W8I89x7Reu0-k9n2gkiMjcOOyP+8Bi0KiLOE6oRdlJJ6t0hRWMNxsOMOmCyGzMJamlHsZlL6KfaANp6zEgM-eih0k5EluogSxy4IXyMnYOQqTPR7Bx-0b01dWb0Z-0Ien-DJGku0pGjnA5alb1maFFNH+W+PwmN6jxsajx7hRaBDRxXEfsDgP0B6C2CxZqhxBJfzbJZGl5wMYky+4MKkEhaF3CsRHOypGQlJ6oJGJgSR1IgJldc6RN28TjSvkvniBaidX3czvEPcltGpxNnFi8eFgeRy9ImJ-D1UemAxwUFfy+GtTAY4reJe+zbxrpNRTrp2a1awgyhtGr00HsrVOBwauiIqwXk3Q1tzBGLTHm6XhFRU5yjV6vI3uEeYMbQSRimMwnE-Y8wN+MT2jUi9mHm6eoKO4vGDm6ZBAPw-9LCU3ahbieVzwZBBEWBlIiHuAHgLIL1JLJ9CT7eQCONekPXixmrAgB82iXuSyilJzgPuhEOTo+UXRZx+xOKBhUWGwo2C16E2Axh5AXek202GMGXnHUq-DUiuXi2R8-w7MWDy0wm+MgZAHB5CkfiEuSFgoO0Ciz+T8OLR9aWKWUKJbswHDeIKhAcy1BmJAyPyJIgVNCaGXVyOkGnyOPeKWy1jUsEgP1LSRvGwxo-zQyJ0QzkTd2KCCCS8CyCWtSTvWVuLaJHpunkIeAlKWpvmMryhaGVqSRPMeN0wNc90ybJvd1hBvwIPxmS1b8+aBv2thKcBXHVl4vN09Idu1c6FVIc+NHRSZ9UjhGNv1s4mjLlkdhzty8JylGY5WNsjVy0UyE0uw9vlyRduE-COZLG42jMfRJpFJcYojoY8VScyjAMQum3RToGMPIWcg3h84mJrOtwxwEM51-pFomqy3DjgBPsK04NbhCJTzJfikqB582XHlu9EAM+mPg6EbnH2WLfmgB9Ak7RCE1e4eAinYttO50EfDVW2y286Xp2vymiMEkHqhDBBzOZJu6EN834i0mg2l9ElFRmCCd28hdwFc0dayym0nGrKIsMgyBU2k89s3H0X9HgCOmUpEZFgjxUg25JNXDTEjeTnMYcPu45LFH0GBgf0NDSaZ-bVQBqyOKCWmTfKumXOUOZXc4LziWJBGM8wWlVqBwhjGejBw2ZuGX8U2BxIUqKSFJtYi1IDYmO4QT2hy3LKWw35hN6bGPnxnGPoqSaHQs85mMUlQLOBZaB+ZdDBDQzTDGmeoI+408kaUDGKhKqOAvebdCeEpW2f05Anl0ZmnIBYOzq4SZLEcp8NrILKUaMguAFmStJY6slIh0UHAKexKmAIC2Buk6YmhpgPEmieukvh0rN7CX7UySeoIlw9DxRIALDTsM1l1UTAxYGj-VpwD-wrQ1pApw6IKCxOUPKuE3EquVqn-6lyKB+ZTmIi1ZQFebRlJaX9CKuHKHzkjIFVZE+CWhEDKOZtXwqulqmYYjMOXeUtAtxElGPKOkQOq6gjYEJ10NO8YPJEmgJZRZYRlKzCPxRfNJioIX3Rir4XWUD5E7+xpm3EPf07mFLxpC6+Q1cQqHyJzjEZkMz3IsDSWva-uFxKRZj+Y+YmEEsrE7oNyIawOHRwMM6TSyqkmdJ+uldJdOMUhlCUxW4f3-Wa4yR0NN3gGWqHaqV3VZe0rHZeigXAoAr3HZsihZusOKvcVH1VaeoPE0v3GHaFK38wMsRfYWDKnB8xOrsp72+WkCN-i5dK649ZIDCjZNuCcMlNEedIAGosgBoJeVvMexI8AJ30KUwTKrpxQQ1pvnC1pi-TuABEQWY8tNHOJGX4YzNMtenN2YhEA1caPqUpuWx29ONzHQq5NRRoxd3i4cTKrUIS1hRgqzZOFGBXhnDHzQRKQVy3HDk4kwwfmidChOVbKgEJEN3I5lA-pVCSBs-zHRGORJVJILk7QR1jJwo-E3xzYJGm9NSL0ujlcY4D3-GpaC1MoiKQIg+ESJDUATgJYAgAHsA2A+kETghUHvgFoFTgiIB3A6AGqSMgH8Q-4GUQLYFq5ZoHq5BgEa5lgGa5C8ja5HSC0QDEE65SKE2g1ICgApQFxQbSH-A70DLANwEUAbADQAMwHpAPsAogSAH0ABCBG5dwDqQ6ME7A7IFfgLcH-Ah3IpAUiDuAb0Gm50wGqAS3JW5qADW5lgCyAKKAcAW3I0Ax3KIQDEDPgVIDO5GiF256AA7AS0C7A53Mm53wGxQ+kAW5e3Mh5L3OQA23NQA3sBkAAIBrQRSEWAMoEJAfXPQAqwGwAvMER5dABu5LXIXQu8BmQLFFx5mAGmABQCQAgwH0AZADQATSFJ50wBRABgGR5NwFp5kIFNAkPIB50PLkAr3Lh5CPITgJEH+AzPPZAImGlgNAAa56AGmAcwAsgePNdA6+BrQoQASAEQFSI0yCxQJPP55ZPLgAFPKp5wwFQAdPPV5DPNRAQvNZ5yKH15R3ITgnXPQA8vLCAcACV5xAHhgOREJ5KvJyARQDYATAHhAacDoAYvIZAuUEN5KPIOAcOFkAwQDQAMsIEAVIE2Qs3LDgHID+Q2yF6Q0UACgl8GCACAFWATsER5PMGUgKgC9gs8BAAMwGW51XNAAqKHDgMyAZAVIDKQGfPh5WfOdAGYHlABgAKgoADyglgB8ATgCgAdACD58cDWg5mFD5c8CPAbfJAghEEMAiICYAakHO5siH3AoAC+ABCGu5RYC55bwCa5dwGqSVIFb59sEn5l3On5YvIG5PMjN0VIEuALPL+5b8FG52-NuAlIA6AByHjg9fI+A7fOz5NAGSATAGUAUAH+Aw-LdAKQE9Ap-LTgWiFAAF-LiAPkEsAg-JnAR-OpAiIDgAsYD+5O-I+5+-JR5MgAQA84D2AofJ+5AAqxQO3JAFhSD357SBZ5lICP5gMHj55vPvgeoHTAZvItAtfMv53fPf5vfIhAzSHRgUsAx5xAqH5lmixA9gFLShkDF5igBJAZIHIFvvMQF7fKFQVIFkA4-J751ArUIXAsX5LEBIFqAA4Cf-L0QoAHGAJgBb55-OEFfKCpQVIB4FoAvaQe-Jj5WcHH5C-J25yAsvgI3NUFmAEoAQfKpAhgGwFu-JO5SAsdgugseQTvMP5aAvzAQAA}{https://vega.github.io/}. It allows zooming, panning, and hovering over points for specific disease names.
Finally, we provide chronic disease label probabilities, full visit level code probability plots, probability densities underlying the aggregated statistics, and a discussion of the various failure modes of our baseline methods for that evaluation in our supplementary information. \method again outperforms the baseline methods in each evaluation.\\

\begin{table*}[]
\small
\centering
\caption{Aggregate statistics regarding the shape of training and compared synthetic datasets}
\resizebox{0.65\textwidth}{!}{
\begin{threeparttable}
\begin{tabular}{c|cc|cc}
\toprule
& \multicolumn{2}{c|}{Outpatient EHR} & \multicolumn{2}{c}{Inpatient EHR} \\
& \makecell{Record Length\\Mean (Std. Dev.)} & \makecell{Visit Length\\Mean (Std. Dev.)}  & \makecell{Record Length\\Mean (Std. Dev.)} & \makecell{Visit Length\\Mean (Std. Dev.)}  \\ \midrule
EVA & 29.49 (28.88) & 3.35 (1.71) & 1.20 (0.723)  
 & 11.92 (3.665)            \\
SynTEG   & 93.00 (2.30)   & 3.70 (4.10)   & 27.55 (3.34)   & 5.93 (10.96)  \\
LSTM      & 32.04 (27.14)   & 3.22 (1.64)    & 1.30 (0.56)   & 9.53 (2.91) \\
GPT      & 95.72 (3.37)   & 2.70 (1.73)    & 1.26 (0.73)   & 9.67 (5.45) \\
\method $-$ Coarse        & 35.26 (31.87)   & 3.77 (2.23)   & 1.13 (0.39)   & 11.21 (3.91) \\
\method      & 36.19 (33.41)   & 3.93 (2.72)   & 1.31 (0.84)   & 11.93 (6.45)  \\ \hline
Train Data   & 34.18 (32.35)   & 3.52 (2.18)   & 1.27 (0.92)   & 11.68 (5.70) \\ \bottomrule
\end{tabular}
\begin{tablenotes}[flushleft]
\item \footnotesize Aggregate statistics on the number of visits per record and the number of codes per visit. The values are mean (std). \method outperformed all the baselines while closely approximating the distribution of the true training data. Source data are provided as a Source Data file.
\end{tablenotes}
\end{threeparttable}}
\label{table:AggregateStats}
\end{table*}

\begin{table*}[]
\small
\centering
\caption{Code probability correlations $R^2$ between training and synthetic datasets}
\begin{threeparttable}
\begin{tabular}{c|ccc|ccc}
\toprule
& \multicolumn{3}{c|}{High-Dimensional Outpatient EHR} & \multicolumn{3}{c}{Low-Dimensional Outpatient EHR} \\
& \makecell{Unigram\\Code Probabilities} & \makecell{Sequential Visit\\Bigram Probabilities}  & \makecell{Same Visit\\Bigram Probabilities} & \makecell{Unigram\\Code Probabilities} & \makecell{Sequential Visit\\Bigram Probabilities}  & \makecell{Same Visit\\Bigram Probabilities}  \\ \midrule
EVA   & 0.910   & 0.082   & 0.128   & 0.957   & 0.134   & 0.225  \\
SynTEG   & \textbf{0.915}   & 0.355   & 0.082   & 0.784   & 0.315   & 0.211  \\
LSTM   & 0.900   & 0.077   & 0.127   & \textbf{0.962}   & 0.135   & 0.225  \\
GPT   & 0.743   & 0.382   & 0.262   & 0.924   & 0.626   & 0.515  \\
\smaller \method $-$ Coarse   & 0.794   & 0.357   & 0.176   & 0.882   & 0.503   & 0.247  \\
\method   & 0.914   & \textbf{0.508}   & \textbf{0.362}   & 0.949   & \textbf{0.686}   & \textbf{0.562}  \\ \bottomrule
\end{tabular}
\begin{tablenotes}[flushleft]
\item The values are $R^2$ values to measure the correlations of the three types of code probabilities for different synthetic datasets against the training data in both high-dimensional and low-dimensional settings. Bold values denote the best results. Although the results showed a drop in performance for each method in the high-dimensional setting, \method was able to maintain strong performance with minimal decline. Overall, our proposed method achieved state-of-the-art performance, outperforming the baselines in both bigram  evaluations in low and high dimensional settings. Source data are provided as a Source Data file.
\end{tablenotes}
\end{threeparttable}
\label{table:VisitLevel}
\end{table*}

\noindent{\bf Key findings:} We observe that besides the GPT baseline struggling with the complexity of the outpatient EHR dataset in terms of stopping the record generation (as is common to many language models in the text generation domain as their overall quality decays for long sequences, and the lack of visit level grouping in its data representation causes its sequences to be considerably longer), the language model architectures (GPT, LSTM, \method $-$ Coarse, and \method) can model both the shape of the synthetic records and the temporal dependencies much better on average than the VAE and especially GAN-based baselines. While each of the compared methods models the unigram code probabilities relatively well, the temporal modeling is better shown in the overall synthetic record and visit lengths, the generation of chronic disease labels, and the sequential bigram evaluation. SynTEG, EVA, and the LSTM baseline thus struggle with these evaluations (with the LSTM baseline struggling largely due simply to overall weakness).

The LSTM and \method $-$ Coarse language model baselines then falter with respect to same-visit bigram probabilities due to their lack of intra-visit dependency modeling while the GPT baseline which models each code individually and so offers that intra-visit modeling can maintain relatively stronger performance there. \method can combine and build on each baseline's strengths without any weaknesses, using the compact multi-hot representation to offer a powerful model that does not struggle with any length or feature of data while simultaneously maintaining the intra-visit modeling in an even more powerful and structured way. As such, it can best maintain performance in this high-dimensional setting and produces state-of-the-art results that closely model the true training data in all settings from record and visit lengths, label probabilities, and all combinations of code probabilities. This signifies that \method is capable of generating data that looks realistic.

\subsection*{Accurate Disease Phenotyping Using Synthetic EHRs} \label{sec:supplementing}
The final evaluation explores the utility of the synthetic datasets for training disease classifiers. To this end, we utilize two different synthetically-supplemented data setups and machine learning classifiers to predict chronic disease labels based on patients' visits. In each of the two setups, we use a simple bidirectional LSTM with a single-layer fully connected head classifier to predict chronic disease label(s) based on a patient's visits. \\

\noindent\textbf{Accurate Disease Phenotyping:} In the first data setup, we assess model performance in real-world scenarios using synthetic training data. We conduct experiments for each of the 11 chronic disease labels in the outpatient EHR dataset, sourced from the Centers for Medicare and Medicaid Services and the SynPUF dataset \cite{SynPUF}. Additionally, we perform experiments for each of the 25 chronic diseases in the inpatient EHR dataset, based on the benchmark proposed in \cite{MimicBenchmark}.

For each chronic disease, we randomly extract 2,500 records for training, ensuring a balanced distribution of positive and negative labels (50-50). This process is repeated across our six synthetic datasets (one for each method) and one real training dataset, resulting in a total of seven balanced training datasets. The selected size of 2,500 records strikes a balance between having enough training data for machine learning models and maintaining sufficient positive labels for each disease.

We train classifiers on each of these datasets and select the best model for each dataset using a validation set of 250 records, equally representing both classes. Finally, we evaluate the models on test sets consisting of 500 records, equally representing both classes, from the original test set comprising real patient data. 

We display the average accuracy and F1 score for each synthetic dataset from each of the compared models as well as the real training data across each of the chronic disease labels in Table \ref{table:SyntheticResults}. Note that we provide the standard deviations of each metric in either table as well, but most of that deviation stems from differences between tasks rather than inconsistent performance within each model.

We provide a full set of results by chronic disease label and also additional synthetically augmented outpatient results in our supplementary information. In both datasets, we can see that each synthetic data of GPT, \method $-$ Coarse, and \method largely maintain the performance of real data and offer large improvements over the SynTEG, EVA, and LSTM baselines. \method's synthetic data offers the best prediction results .\\

\begin{table}[]
\centering
\caption{Chronic disease classification model performance trained on synthetic data}
\resizebox{0.95\columnwidth}{!}{
\begin{threeparttable}
\begin{tabular}{c|cc|cc}
\toprule
& \multicolumn{2}{c|}{Outpatient EHR} & \multicolumn{2}{c}{Inpatient EHR} \\
& Avg. Accuracy  & Avg. F1 Score   & Avg. Accuracy  & Avg. F1 Score 
\\ \midrule
EVA & 0.508 $\pm$ 0.02   & 0.283 $\pm$ 0.26  
        & 0.5356 $\pm$ 0.05  & 0.580 $\pm$ 0.05 \\
SynTEG  & 0.507 $\pm$ 0.03   & 0.514 $\pm$ 0.20 
        & 0.539 $\pm$ 0.06  & 0.438 $\pm$ 0.06 \\
LSTM & 0.506 $\pm$ 0.02   & 0.467 $\pm$ 0.28 
        & 0.522 $\pm$ 0.04  & 0.565 $\pm$ 0.04\\
GPT & 0.851 $\pm$ 0.03   & 0.854 $\pm$ 0.03 
        & \textbf{0.877 $\pm$ 0.05}  & \textbf{0.881 $\pm$ 0.05} \\
\method $-$ Coarse & 0.867 $\pm$ 0.03   & 0.863 $\pm$ 0.03  
        & 0.863 $\pm$ 0.05  & 0.865 $\pm$ 0.05\\ 
\method & \textbf{0.879 $\pm$ 0.03}  & \textbf{0.878 $\pm$ 0.03}    
        & \textbf{0.882 $\pm$ 0.04}  & \textbf{0.884 $\pm$ 0.04} \\ \hline
Real Data & 0.891 $\pm$ 0.03   & 0.895 $\pm$ 0.03   
        & 0.938 $\pm$ 0.04 & 0.937 $\pm$ 0.04 \\ \bottomrule
\end{tabular}
\begin{tablenotes}[flushleft]
\item We compared the average performance in terms of accuracy and F1 Score for each of the 11 chronic disease labels in our outpatient dataset and 25 chronic disease labels in our inpatient dataset. The models were trained on each of our synthetic datasets and tested on real data. The reported values represent the mean and standard deviation across the tasks, with bold values indicating the best results. GPT, \method $-$ Coarse, and \method's data offer large improvements over the other baselines and perform similarly to real training data. \method's synthetic data performs the best with the highest average performance of all synthetic methods. Source data are provided as a Source Data file.
\end{tablenotes}
\end{threeparttable}}
\label{table:SyntheticResults}
\end{table}

\noindent\textbf{Phenotyping of Rare Conditions:} We evaluate the utility of synthetic EHR data in identifying uncommon conditions. We created a highly imbalanced dataset of patients labeled with cancer chronic disease from the outpatient EHR dataset. The dataset comprised 50,000 EHR records without the cancer chronic disease label and only 1,000 records with the label.

Using this imbalanced data, we trained a classifier and compared its performance to classifiers trained on balanced datasets. For balancing, we added 49,000 positively labeled synthetic records and also used another classifier trained on a dataset balanced using real records.

The evaluation results are summarized in Table \ref{table:RareDiseaseResults}. Notably, \method outperformed all baselines, exhibiting significant improvements on the original unbalanced dataset as well as the synthetically augmented datasets. It approached the upper bound performance of the ideal balanced dataset.

This experiment underscores the potential of synthetic EHR data in supporting the identification of uncommon conditions. 

\begin{table}[]
\small
\centering
\caption{Rare disease detection performance on synthetic balanced datasets}
\begin{threeparttable}
\begin{tabular*}{0.9\columnwidth}{c|ccccccc}
\toprule
&\footnotesize BCE Loss  &\footnotesize Accuracy &\footnotesize F1 Score  &\footnotesize AUROC     
\\ \midrule
\footnotesize Original imbalanced & 0.693  & 0.497  & 0.013  & 0.417 \\
Balanced with real data  & 0.127  & 0.951  & 0.951  & 0.989 \\ \hline
EVA & 0.615  & 0.695  & 0.705  & 0.730 \\
SynTEG & 0.598  & 0.735  & 0.758  & 0.786 \\
LSTM & 0.593  & 0.702  & 0.714  & 0.743 \\
GPT & 0.472  & 0.880  & 0.869  & 0.956 \\
\method $-$ Coarse & 0.265  & 0.918  & 0.916  & 0.959 \\
\method   & \textbf{0.192} & \textbf{0.931}  & \textbf{0.931} & \textbf{0.976} \\ \bottomrule
\end{tabular*}
\begin{tablenotes}[flushleft]
\item We present the classification results on the test set for the simulated rare-disease detection task. We compare models trained on datasets balanced using each synthetic dataset against models trained on the original imbalanced data (representing the rare disease dataset). Additionally, we compare the results against an upper-bound ideal dataset balanced using real data. The best results are highlighted in bold.
Among the evaluated models, EVA and SynTEG exhibit limited utility, while the language model architectures LSTM, GPT, and \method $-$ Coarse offer substantial value. \method achieves state-of-the-art performance, closely approaching the results of a true, balanced dataset. The source data can be found in the provided Source Data file.
\end{tablenotes}
\end{threeparttable}
\label{table:RareDiseaseResults}
\end{table}

\subsection*{Realistic Continuous Variables in Synthetic EHRs}
\label{method:generate-lab}
We conclude with a brief exploration to demonstrate the viability of our discretized representation of continuous values, and \method's effectiveness in using it to model those variables. We build new training datasets including visit gaps in the outpatient EHR dataset and lab values in the inpatient EHR dataset. We use these dataset to train a new version of our model and generate another synthetic dataset of 250,000 and 45,000 records respectively. 

\begin{figure*}
        \centerline{\includegraphics[scale=0.35]{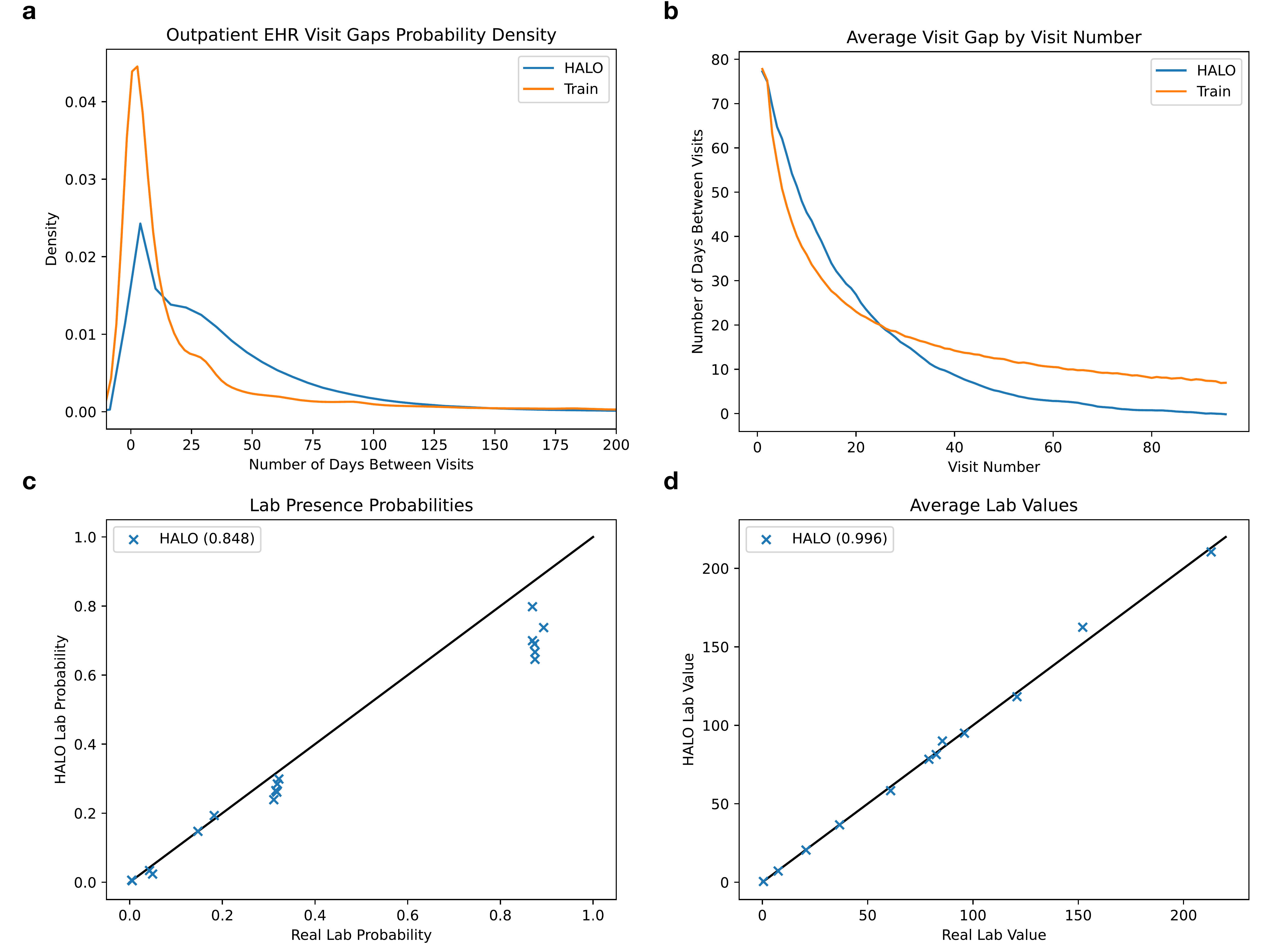}}
    \caption{Continuous Variable Generation Performance: \method effectively captures the distribution of continuous variables through its discretization approach, as demonstrated in four scenarios.
a) Inter-visit gap probability density: The probability density of inter-visit gaps indicates that \method closely approximates the true shape of real data.
b) Inter-visit gap by visit number: The mean visit gap, as per visit number, across both real and synthetic datasets reveals that \method accurately captures the pattern of patients with many records, showing shorter gaps in their subsequent visits.
c) Lab presence probabilities: The probability of binary lab presence demonstrates that \method accurately generates lab variables, even when discretized across multiple variables.
d) Mean lab values: The average value of labs, when present, confirms that \method's synthetic labs closely resemble those of the real dataset. Values in parentheses are $R^2$. }
    \label{fig:ContinuousStats}
\end{figure*}

We then show that the distributions of those variables match the real values. In Figure \ref{fig:ContinuousStats}a and Figure \ref{fig:ContinuousStats}b, we show that \method accurately replicates the distribution of gaps between patient visits and the pattern of shorter gaps for longer records, respectively. These captured nuanced patterns are on top of the aggregate mean gaps being very similar as well. There are 33.53 days between visits on average within the real outpatient EHR data and 35.77 days on average for \method's synthetic data.

Using the inpatient dataset, we then demonstrate that \method replicates not only the presence (in Figure \ref{fig:ContinuousStats}c) but also the average values (in Figure \ref{fig:ContinuousStats}d) of performed lab tests. Specific labs included (corresponding to points in those two plots) are included in our supplementary information. Overall, \method's approach to continuous variables is effective, and it has the potential to generate comprehensive synthetic patient records with multiple variables of different types.

\subsection*{Privacy Evaluation of Synthetic EHRs} \label{sec:securing}
In addition to demonstrating the high fidelity of synthetic EHRs generated by \method, we want to ensure that the privacy of the patients within the original training dataset is protected. To that end, we conduct a commonly used membership inference attack to test its identification risk, and we provide the results of two more evaluations in our supplementary information. 

\noindent\textbf{Membership Inference Attack:} The evaluation is the ability to thwart a membership inference attack. These attacks aim to determine whether any specific real patient record was used in the training dataset to generate the synthetic records. Membership inference attacks are a well-known privacy test in the field of synthetic EHR generation, and addressing them is crucial to ensure the privacy and confidentiality of patient identities.

To demonstrate that \method is not susceptible to such an attack, we show that we can prevent two different attempts at a membership inference attack based on the synthetic data generator and the synthetic dataset itself. We generate an attack dataset by first selecting 100,000 records from each real dataset used for training and assigning them a positive label. Then we select 100,000 records from the remaining records not used for training as the negative label set.

Next, we conduct two attacks: 
\begin{itemize}
\item 
In the Model Attack, we label the 100,000 records with the highest log probability from the model as positive, predicting that they were part of the training dataset.
\item In the Dataset Attack, we label the 100,000 records with the lowest hamming distance to the closest record in synthetic dataset as positive. We pick hamming distance (equivalent to Manhattan Distance in our binary setting) as our distance metric between patient records throughout our privacy evaluations in accordance with \cite{yan2022multifaceted}, but any distance metric could be substituted interchangeably.
\end{itemize}
These two attacks allow us to test the ability of the synthetic dataset to prevent an attacker from inferring whether a real record was used in the training dataset.

We show the results of the classifications from the attacks in Table \ref{table:membershipAttack}. The accuracy of both attacks on both datasets is approximately 50\%, which is similar to a random guess. This shows that neither the model nor the synthetic dataset reveals any meaningful or compromising information about the patient identity in the training dataset. We also perform the dataset attack with each of our baseline datasets and see that each similarly accomplishes it, achieving a similar probability at around 50\%. 
Note that we do not perform the model attack with the baseline models because most of them cannot offer a probability output of input patient records, and the dataset-based attack is the standard one used throughout literature in this domain.\\

\begin{table}[]
\centering
\caption{Membership inference attack results}
\resizebox{0.95\columnwidth}{!}{
\begin{threeparttable}
\begin{tabular}{c|ccc|ccc}
\toprule
& \multicolumn{3}{c|}{Outpatient EHR} & \multicolumn{3}{c}{Inpatient EHR} \\
&\footnotesize Acc.       &\footnotesize Precision     &\footnotesize Recall  
&\footnotesize Acc.       &\footnotesize Precision     &\footnotesize Recall  
\\ \midrule
\method Dataset Attack   & 0.501   & 0.501   & 0.501  
& 0.492   & 0.491   & 0.477 \\
\method Model Attack     & 0.509    & 0.509   & 0.509
& 0.515    & 0.515   & 0.515 \\\hline
EVA Dataset Attack    & 0.498    & 0.498   & 0.496
& 0.493    & 0.493   & 0.477    \\
SynTEG Dataset Attack    & 0.500    & 0.500   & 0.500
& 0.491    & 0.491   & 0.467    \\
LSTM Dataset Attack    & 0.499    & 0.499   & 0.496
& 0.494    & 0.494   & 0.481    \\
GPT Dataset Attack    & 0.500    & 0.500   & 0.500
& 0.492    & 0.491   & 0.455    \\
\method $-$ Coarse Dataset Attack    & 0.500    & 0.500   & 0.499
& 0.491    & 0.491   & 0.462 
\\ \bottomrule
\end{tabular}
\begin{tablenotes}[flushleft] 
    \item For each record in the attack dataset, we find both the log probability of the record from the trained model (Model Attack) and the hamming distance to the closest record in the synthetic dataset (Dataset Attack). The attacks then label the half of the records with the highest probability or lowest distance records, respectively, as in the training set. We see that the accuracy for either attack is right around 50\%, which is similar to a random guess. This indicates that the synthetic dataset and the model do not reveal any patient-identifying information about the original training datasets. We also find that each baseline synthetic dataset similarly thwarts the dataset attack. Source data are provided as a Source Data file.
\end{tablenotes}
\end{threeparttable}}
\label{table:membershipAttack}
\end{table}

Beyond membership inference attack, we also show that \method passes attribute inference attack and nearest neighbor adversarial accuracy \cite{yale2020generation} evaluations in our supplementary information.

\section{Discussion}
In this paper, we proposed a method \method for generating high-dimensional synthetic longitudinal EHR data. Our method is specifically designed to handle the sequential, multi-granular, and high-dimensional nature of electronic health records by generating an explicit probability distribution over the codes, visits, and records, and \method can generate realistic data so without needing to aggregate or remove any codes as past approaches have unanimously done. We then showed that \method can produce incredibly realistic synthetic EHR data. Specifically, we showed that \method can capture the probability distribution underlying the records better than other language model baselines and then produce a synthetic dataset that both looks similar to and offers the utility of real patient records as measured by medical code occurrence probabilities and machine learning classification tasks augmented with synthetic data. Finally, we also show that our method offers this performance without compromising privacy through several privacy evaluations. 

In conclusion, one of the key advantages of \method is its ability to generate binary sequences that are over a million variables in length. Its impressive performance makes it a promising avenue for developing and sharing realistic but synthetic EHR datasets that can support diverse applications. This represents an exciting opportunity to expand the use of synthetic data in the healthcare field and could help to address some of the challenges associated with data privacy and security.

While we have shown the impressive performance of \method in both producing high-quality, high-fidelity, and privacy-preserving, we now briefly discuss some remaining limitations. First, the architecture is designed in the model of a large language model. While the multi-modal setup allows the model to condition on more patterns per data point and learn more efficiently, our high-performing generator still requires relatively large training datasets which might not be available in some settings. 

Another important aspect of our model is that it generates synthetic records through a probabilistic process. While it learns real-world patterns during training, there is still a chance that some generated records may not be clinically meaningful. However, this risk can be mitigated through postprocessing with clinical rules that validate the synthetic records. If our model is deployed in the real world, it is important to consider implementing such postprocessing steps to ensure that only clinically relevant synthetic records are produced.

Finally, our \method model focuses on generating longitudinal EHR data, such as medical codes and lab results. However, other crucial data modalities, such as clinical notes and medical images, are not yet covered by the model. To generate fully comprehensive patient records that include all modalities, it will be necessary to use diverse training data and develop multiple models to handle each modality. This exciting avenue of research is a promising future direction.

\section{Methods}
Our study is acquired exempt status from Institutional Review Board (IRB) approval. This study has been found to be exempt pursuant to 45CFR46.104(d)(4) "Secondary research for which consent is not required: Secondary research uses of identifiable private information, if (i) The identifiable private information is publicly available; AND (ii) Information is recorded by the investigator in such a manner that the identity of the human subjects cannot readily be ascertained directly or through identifiers linked to the subjects, the investigator does not contact the subjects, and the investigator will not re- identify subjects."

\subsection*{Background and Related Work}
Of all the EHR generation methods, rule-based approaches, such as Synthea \cite{Synthea} or SynPUF~\cite{SynPUF}, have proven to be the most effective in delivering practical value. These simple approaches either offer de-identification of real records by combining data across multiple patients in a sufficiently privacy-preserving way \cite{SynPUF}, simulation of patients within a complex yet constrained rule-based system \cite{Synthea}, Bayesian probabilistic modeling of aggregated, non-temporal patient records \cite{Bayesian}, or proprietary method without detailed explanation \cite{EMRBots, MDClone, Syntegra}. Many of these systems can only produce synthetic patient data with limited capacity in realism and utility. We focus instead on ML methods that have the potential to generate realistic high-dimensional synthetic patient data.\\

\noindent\textbf{GAN-based Methods}
Many synthetic data generation methods use Generative Adversarial Networks (GANs), which involve  a generator that creates realistic data, and a discriminator that decides if the data is real or fake~\cite{GANs}.
The GANs has been applied to patient record generation first in \cite{MedGAN} followed by many other GAN-based approaches \cite{MedWGAN, EMR-WGAN, HGAN, CorGAN, LongGAN, SmoothGAN, CONAN, EHRMGAN, SynTEG}. However, GANs have limitations when generating sequential data like EHRs. They usually only produce one output (no time connections) and so most EHR generation methods aggregate EHR data into one time step \cite{EMR-WGAN, HGAN, SmoothGAN}, create a representation of EHR data \cite{CONAN}, or do both \cite{MedGAN, MedWGAN}.

GANs also struggle with high dimensional and sparse data like real-world EHR, limiting all existing synthetic EHR GAN approaches to produce relatively low dimensional data through the aggregation of visits and medical codes or removal of rare codes. For example, there are a few methods in this category which do generate longitudinal data. LongGAN \cite{LongGAN} and EHR-M-GAN \cite{EHRMGAN} both focus only on dense lab time series of under a hundred dimensions. CorGAN \cite{CorGAN} generates records with 1,071 distinct codes, and the current state of the art GAN approach that we baseline against, SynTEG \cite{SynTEG}, both combines and removes rare codes before arriving at a final dimensionality of 1,276.

While GANs have the potential to be conditioned on external factors and labels, such as demographics or disease phenotype labels, the ability to do so has not been extensively explored in existing works on EHR generation. Moreover, there are only a limited number of approaches that can generate synthetic EHR data tailored to specific diseases. For example, SmoothGAN \cite{SmoothGAN} focuses on aggregated lab and medication information and does not model individual visits; EHR-M-GAN \cite{EHRMGAN}  offers conditional and sequential capabilities, but for low dimensional (under 100 dimensions) lab time-series information; CONAN and MaskEHR \cite{CONAN, MaskEHR} model only a single rare-disease population for data augmentation; and EMR-WGAN and HGAN \cite{EMR-WGAN, HGAN} can only model low-dimensional (both under 1000 dimensions) aggregated EHRs.\\

\noindent\textbf{Deep Sequential Methods}
Accurately modeling the longitudinal nature of EHRs is crucial for realistic EHR generation. In recent years, two methods have shown progress in generating sequential EHRs by using either a GAN or a VAE to condition on representations of past patient visits to generate current visits \cite{SynTEG, EVA}. Specifically, SynTEG \cite{SynTEG} models the time between visits, and EVA \cite{EVA} offers a conditional variant. In our experiments, we compare \method to these two models. However, both SynTEG and EVA often need to perform preprocessing steps to reduce the dimensionality of the vocabulary by aggregating medical codes and removing rare codes.\\

\noindent\textbf{Language Models}
Our objective is to develop an improved method for generating realistic and high-dimensional EHR data by drawing inspiration from natural language generation. Language generation models predict the next word based on the preceding words, thereby learning a probability distribution of languages. Similarly, EHR models predict the next visit based on past visits. Also our proposed method  provides an explicit probability output that allows for direct modeling and evaluation of the underlying data distribution. This approach is particularly beneficial in accurately capturing the complex and high-dimensional nature of EHR data.

The Transformer architecture, introduced in \cite{Attention}, has revolutionized natural language processing and enabled the development of large, attention-based models like BERT \cite{BERT} and GPT \cite{GPT1, GPT2, GPT3}. Among these models, we draw inspiration from GPT, which relies on a stack of Transformer decoder blocks that use masking to predict the next set of probabilities in parallel, allowing for fast training and scalability.
However, applying language models directly to EHR data poses unique challenges. Unlike natural language sequences, EHR data exhibits a hierarchical structure that must be captured, with medical codes associated with specific patient visits, and visits associated with individual patients. Additionally, EHR data contains heterogeneous elements, including demographic variables, structured medical codes, and numeric lab measures, not all of which are discrete tokens. Addressing these challenges requires approaches that leverage the strengths of language models while adapting them to the peculiarities of EHR data.

\subsection*{Hierarchical Autoregressive Language Model (\method)}
We model the probability of patient record  $\mathbf{R}$, $P(\mathbf{R})$, via a hierarchical autoregressive model, which utilizes both visit- and code-level structures of a patient record. First, it factorizes the probability along the visit level using the autoregressive identity by
\begin{align}\begin{split}
    P(\mathbf{R}) &= P(\mathbf{v}_s, \mathbf{v}_l, \cdots, \mathbf{v}_T, \mathbf{v}_e) \\
           &= P(\mathbf{v}_s)P(\mathbf{v}_l|\mathbf{v}_s)P(\mathbf{v}_1|\mathbf{v}_s,\mathbf{v}_l) \cdots P(\mathbf{v}_{e}|\mathbf{v}_s, \mathbf{v}_l, \cdots, \mathbf{v}_{T}) \\
\end{split}\end{align} 
to produce what we call our coarse autoregressive sequence. We then continue to factorize the probability of visits further along the code level by converting
\begin{align}\begin{split}
    P(\mathbf{v}_t|\mathbf{v}_s, \mydots\,, \mathbf{v}_{t-1}) &=  P(c_t^1|\mathbf{v}_s, \mydots\,, \mathbf{v}_{t-1}) P(c_t^2|\mathbf{v}_s, \mydots\,, \mathbf{v}_{t-1},c_t^1)\\
           &\;\;\;\;\mydots\, P(c_t^C|\mathbf{v}_s, \mydots\,, \mathbf{v}_{t-1}, c_t^1, \mydots\,, c_t^{C-1})
\end{split}\end{align} 
into what we call our fine autoregressive sequence. This final probability is then rewritten as the product
\begin{equation}\label{eq:probability}
    P(\mathbf{R}) = \prod_t \prod_i^C P(c_t^i|\mathbf{v}_s, \cdots, \mathbf{v}_{t-1}, c_t^1, \cdots, c_t^{i-1})
\end{equation}
where the probability of each code is based on each of the previous visits and each of the previous codes in the current visit. Our multi-granularity approach enables the modeling of high-dimensional sequences of many binary variables per record. This is achieved by grouping prior information into significantly fewer multivariate time steps for previous visits, while retaining the full autoregressive modeling capability for each current visit.
Our \method architecture is designed to reflect this powerful yet compact model, with a powerful and efficient structure divided into two distinct granularity levels: visit level and code level. This allows for each code to be conditioned on all previous visits and the past codes of the current visit.\\

\noindent\textbf{Visit-Level Module}
We begin with the coarse, visit-level granularity. We use a stack of $M$ transformer decoder blocks, which have shown to be effective in the high-dimensional domain of natural language processing, to generate a sequence of visit-level histories, where the $t$-th element in the sequence, $\mathbf{h}^{(M)}_t \in \mathbb{R}^{n_{\text{emb}}}$, is an embedding that represents all of a patient's medical history through their $t$-th visit. Those histories then combine to form $\mathbf{H}^{(M)} \in \mathbb{R}^{(T+3) \times n_{\text{emb}}}$ (where the $3$ in $T+3$ includes the start, label, and end visits), the output of the first module which serves of the purpose of the $\mathbf{v}_s, \mathbf{v}_l, \mathbf{v}_1, \cdots \mathbf{v}_{t-1}$ priors in Equation \ref{eq:probability}.

To encode each of the multi-hot visit representations $[\mathbf{v}_1 \cdots \mathbf{v}_n]$ into a fixed-length vector in $\mathbb{R}^{n_\text{emb}}$, we employ an embedding layer that includes two trainable parameter matrices: a code embedding matrix $\mathbf{W}_c$ and a positional embedding matrix $\mathbf{W}_p$. The code embedding matrix maps each visit code to a dense vector representation, while the positional embedding matrix captures the relative position of each visit in the sequence. Next, we use a decoder model consisting of $M = 12$ transformer decoder blocks to generate a series of visit history representations, which summarize the information contained in all previous visits in the coarse, visit-level sequence. The transformer decoder blocks employ masked multi-head self-attention, which allows the model to attend to all previous visits while preventing information leakage from future visits.
This process is written more formally as
\begin{align}\begin{split}
    \mathbf{H}^{(0)} &= \mathbf{R}\mathbf{W}_e + \mathbf{W}_p \\
    \mathbf{H}^{(m)} &= \text{transformer\_block}(\mathbf{H}^{(m-1)}) \;\;\;\forall \:m \in [1,M]
\end{split}\end{align}
where $\mathbf{R} \in \mathbb{R}^{(T+3) \times C}$ is the patient record matrix representation, $\mathbf{W}_e \in \mathbb{R}^{C \times n_{\text{emb}}}$ is the code embedding matrix, $\mathbf{W}_p \in \mathbb{R}^{(T+2) \times n_{\text{emb}}}$ is the positional embedding matrix (to recapture the position and order of the sequence of visits), and each transformer block is based on a decoder block from the original transformer architecture \cite{Attention} which we describe in more detail in our supplementary information.

Thus, having processed the multi-hot patient visits through the initial, coarse visit-level module of our architecture, we obtain a sequence of visit history representations $\mathbf{H}^{(M)}$, which capture the collective information of all previous visits up to each time step. These representations provide a compressed summary of the patient's visit history, enabling downstream modules to make predictions based on the patient's medical trajectory.\\

\noindent\textbf{Code-Level Module}
However, we still need to add in the code-level priors and generate output probabilities.
To construct the input for the fine, code-level module, we offset and concatenate the previous module's visit history embedding outputs with the original record input, $\mathbf{R}$. Specifically, we append the first $T+2$ visit histories with the last $T+2$ visit representations $[\mathbf{v}_l, \mathbf{v}_1, \cdots, \mathbf{v}_{T}, \mathbf{v}_e]$ to create $\mathbf{H}'^{(0)}$. Each of the $T+2$ inputs in $\mathbf{H}'^{(0)}$ has a representation of the history of all the previous visits and the codes of the current visit, mirroring both the visit and code priors in Equation \ref{eq:probability}. The final input representation $\mathbf{H}'^{(0)}$ has size $\mathbb{R}^{(T+2) \times (n_{\text{emb}}+C)}$

To model the distribution of each $P(c_t^i)$, this $\mathbf{H'}^{(0)}$ is then fed through $N = 2$ masked linear layers which maintain the same dimensionality and use upper triangular masking of the weight matrix to ensure that they preserve the autoregressive property of the probabilities (and have a ReLU activation function between layers). These linear layers are able to efficiently model the high-dimensional, intra-visit patterns where other sequential approaches such as additional recurrent or transformer modules would run out of memory. The probabilities are generated formally by
\begin{align}\begin{split}
    \mathbf{H'}^{(0)} &= \text{offset\_and\_concat}(\mathbf{H}^{(M)}, \mathbf{R}) \\
    \mathbf{H'}^{(n)} &= \text{masked\_linear}(\mathbf{H'}^{(n-1)}) \;\;\;\forall \:n \in [1,N] \\
    \mathbf{O} &= \text{sigmoid}(\mathbf{H'}^{(N)}[\,:\;,\,n_{\text{emb}}:\,])\\
\end{split}\end{align}
where the submatrix indexing at the end removes the visit-level history embedding portions of each vector to extract just the code probabilities, and the masked linear layers are achieved by
\begin{align}\begin{split}
    \mathbf{H'}^{(n)} &= \max(0,\mathbf{H'}^{(n-1)}(\mathbf{W}^{(n)} \odot \mathbf{M}) + \mathbf{b}^{(n)})
\end{split}\end{align}
where the $\max$ function is omitted for the final fine layer (sigmoid is used instead), $\odot$ is element-wise matrix multiplication, $\mathbf{M} \in \mathbf{R}^{(n_{\text{emb}}+C) \times (n_{\text{emb}}+C)}$ is the upper triangular masking matrix (with ones in the upper triangular portion and zeros in the lower portion) to preserve the autoregressive property, and $\mathbf{W}^{(n)} \in \mathbb{R}^{(n_{\text{emb}}+C)\times(n_{\text{emb}}+C)}$ 
and $\mathbf{b}^{(n)} \in \mathbb{R}^{n_{\text{emb}}+C}$ are the trainable parameters of the module.

The output $\mathbf{O} \in \mathbb{R}^{(T+2) \times C}$ is then a matrix of probabilities of each code for each visit after the start visit built from the visit histories and each previous code in the same visit.  Each code corresponds to a conditional probability in the product from Equation \ref{eq:probability}. 

We train our model using the binary cross-entropy loss function over each medical code (treating the problem as a multi-label classification problem) with masking applied such that the start visit as well as any padded visits (of all zeros) do not contribute to the loss. The architecture of our model is shown in Figure \ref{fig:MediSynArchitecture}.

\subsection*{Additional Features and Considerations}
Finally, We discuss different variants and add-on features of \method. \\

\noindent\textbf{Conditional Generation}
Our method generates electronic health record (EHR) data by using demographics $\mathcal{S}$ and chronic disease phenotypes $\mathcal{D}$ as labels, which are represented in our label vocabulary and applied to individual visits, as shown in Figure \ref{fig:DataRepresentation}. We selected these labels based on their relevance to downstream use cases. Each label is represented as a binary variable in $\mathbf{v}_l$, indicating the presence of the corresponding disease or demographics group indicator. These indicators are defined by concepts such as specific categories of genders, races, ethnicity, age groups, and more. We can easily extend this strategy to include other labels of interest, such as various biomarkers, patient outcomes, or even abstract patient embeddings.\\

\noindent\textbf{Unconditional Generation}
Our setup generates electronic health record (EHR) data with conditional labels by incorporating a "label visit" in the data format, as illustrated in Figure \ref{fig:DataRepresentation}. This format enables easy generation of labeled and conditional data, which are highly valuable for using synthetic data in machine learning tasks and as an augmentation tool, particularly for rare cohorts. However, it's important to note that this formatting is optional. If desired, the "label visit" component can be removed from the EHR representation, and the architecture can be trained to generate unconditioned EHRs without any modification.\\

\noindent\textbf{Generation of Continuous Variables}
Our model can generate not only medical codes  but also continuous variables, such as lab values and temporal gaps between visits. However, the availability of these additional variables in the generated data depends on their presence in the original dataset used for training. For example, the outpatient EHR dataset used in our study includes the time between visits, while the inpatient EHR dataset includes lab values.

In previous models, continuous values were typically generated using either GANs, which lack the autoregressive probabilistic modeling that we employ, or value predictors (such as time series analysis models), which we often found to produce average values with insufficient variance. To overcome these limitations, we model continuous variables within the healthcare domain by discretizing lab values and temporal gaps into clinically equivalent buckets. The resulting binary variables are included in the model's context, denoted as $\mathcal{C}$, before being converted back to continuous values through random uniform sampling within the corresponding bucket range. By using this approach, our model generates more realistic and diverse continuous variables than previous methods.

More specifically, to generate discrete versions of continuous variables, such as lab values and temporal gaps, we divide the range of each variable into several "buckets", as represented by the values $b_1, b_2, \cdots, b_{|l^{(t)}_j|}$, where $|l^{(t)}_j|$ refers to the number of buckets required. We determine the bucket ranges by either seeking advice from clinicians on practical ranges,  creating granular but equivalent groupings, or using a histogram construction algorithm \cite{guha2006approximation}. The same approach is applied to temporal gaps as well.

For example, the heart rate lab test with possible values ranging from 0 to 400 beats per minute down could be broken down into twenty different buckets splitting the overall span into smaller ranges which offer the same medical meaning for all their contained values. This breakdown could have $b_1 = (0,40)$ and $b_7 = (90, 100)$. These buckets then convert the single continuous variable into many binary variables. Whenever the continuous variable is present in the original EHR, a single one of those variable representing the corresponding bucket is set to 1 with the rest remaining 0. For instance, if a patient has a heart rate lab measurement of 93 bpm in their seventh visit, the seventh of the new heart rate variables would be 1 and the rest would remain 0. If there was no such lab measurement in the visit, they would all be 0. 

These new binary variables are added into the wider code vocabulary $\mathcal{C}$ and treated in the same way as all of the other medical codes in the vocabulary by our \method model during learning and generation. After generation, the specific lab values and inter-visit gaps are converted back into a continuous value by uniformly sampling from the corresponding bucket range at the very end.

This discretization allows us to maintain the same powerful and probabilistic modeling process, matching the probabilistic variance of real continuous values in the same way we match the variance of medical code presences. However, by building appropriately granular buckets, we can avoid losing  meaningful information and maintain a full representation of a patient. We explore the performance of this approach further in our experiments.

\section{Data Availability}
The MIMIC-III inpatient EHR dataset \cite{MIMIC} that we use is publicly available and may be downloaded and used freely after performing training and applying on \url{physionet.org}. Furthermore, we also released the synthetic data for each of our compared methods for both the inpatient and outpatient datasets at \url{https://figshare.com/articles/dataset/HALO_Synthetic_Data/23811162}. 
These datasets can then be used to reproduce the results and data statistics. 

\section{CODE AVAILABILITY}
We make our code for the inpatient dataset experiments, including dataset construction, modeling building, training, and evaluation, available at \url{https://github.com/btheodorou99/HALO_Inpatient} \cite{btheodorou99_2023_8041405}. Between this and public availability of that dataset, all inpatient results can be fully reproduced. Furthermore,  \method  is also included in the open-source machine learning package for healthcare PyHealth \cite{pyhealth2022github}, where it is available for easy use in concert with various machine learning tasks.

\section{Acknowledgements} This work is in part supported by National Science Foundation award SCH-2014438, IIS-1418511, CCF-1533768, IIS-2034479, the National Institute of Health award NIH R01 1R01NS107291-01 and R56HL138415, all to J.S. 

\section{Author Contributions} B.T. and J.S. proposed the method, B.T. and conducted all the experiments, B.T., C.X. and J.S. wrote the manuscript.

\section{Competing Interests} The authors declare no competing interests.

\end{document}


\bibliographystyle{naturemag}
\title{Supplementary Information for ``Synthesize High-dimensional Longitudinal Electronic Health Records via Hierarchical Autoregressive Language Model''}
%

%


\author{Brandon Theodorou$^{1,2}$, Cao Xiao$^{2}$,  Jimeng Sun$^{1,2*}$}
\affiliation{University of Illinois at Urbana-Champaign, 201 North Goodwin Avenue,
Urbana, IL 61801$^{1}$}
\affiliation{Medisyn Inc., Las Vegas, NV$^{2}$}
\affiliation{$^{*}$ To whom correspondence should be addressed: jimeng@illinois.edu}

\maketitle

We aim to make our main paper self-contained, comprehensive, and easy to understand by providing a detailed account of our task, the proposed \method method, and the experimental results. In order to ensure greater completeness and clarity, we have included additional model details, dataset information, and results in this section.

\section{Supplementary Note 1}
One of the main contributions of \method is its ability to provide strong performance in the high-dimensional setting found with real-world EHRs. In our main paper, we provide a comparison with the relatively low dimensionalities of some popular previous works in this domain. Here we provide a comprehensive comparison to all previous works we could find a dimensionality count for in Supplementary Table \ref{table:OtherDimensionalities}.

\begin{table}
\centering
\begin{tabular}{l|c} \toprule
Method   & Dimensionality \\ \midrule
CONAN \cite{CONAN}          & $128^*$ \\
CorGAN \cite{CorGAN}          & $1,071^*$ \\
EHR-M-GAN \cite{EHRMGAN}          & 98 \\
EMR-WGAN \cite{EMR-WGAN}          & $944^*$ \\
EVA \cite{EVA}          & $-\char`\^$ \\
HGAN \cite{HGAN}          & $926^*$ \\
MedGan \cite{MedGAN}          & $615^*$ \\
MedWGAN \cite{MedWGAN}          & $1,651^*$ \\
SynTEG \cite{SynTEG}          & 1,276 \\
\method          & $9,882$ \\
\bottomrule
\end{tabular}
\caption{Dimensionalities for previous ML approaches for generating synthetic EHR Data. * signifies a non-longitudinal output (producing either a patient embedding or a single aggregated vector instead of a series of visits) while \char`\^ signifies the special case of one-hot vector output that can only generate a limited number of common code combinations per visit predefined based on patterns from the training EHR data. No past approaches have ever produced synthetic health record data matching the high-dimensionality (on the order of 10,000+ medical codes).}
\label{table:OtherDimensionalities}
\end{table}

\section{Supplementary Methods}

\subsection{Notation}
We provide a table of notations for reference in Supplementary Table \ref{tab:Notation}

\begin{table}
\small
\centering
\caption{Table of Notations}
\begin{tabular}{c|p{0.3\textwidth}} \toprule
Notation   & Description \\ \midrule
$\mathcal{R}$          & A patient's EHR medical record \\
$\mathcal{V}^{(t)}$          & The $t$-th visit in $\mathcal{R}$ \\
$m^{(t)}_i$          & The $i$-th medical code in $\mathcal{V}^{(t)}$ \\
$l^{(t)}_j$          & The $j$-th lab value in $\mathcal{V}^{(t)}$ \\
$g^{(t)}$          & The gap between the $t-1$ and $t$-th visits \\
$\mathcal{S}$          & A patient's static demographic information \\
$\mathcal{D}$          & A patient's chronic disease information \\
$\mathcal{L}$          & The set of all labs \\\hline
$T \in\mathbb{N}$  & The number of visits in $\mathcal{R}$ \\
$\mathcal{C}$          & The set of all medical codes \\\hline
$\mathbf{R}\in\mathbb{R}^{(T+3) \times |\mathcal{C}|}$          & The matrix representation of $\mathcal{R}$, $\mathcal{S}$, and $\mathcal{D}$ \\
$\mathbf{v}_t\in\mathbb{R}^{|\mathcal{C}|}$          & The vector representation of the $t$-th visit in $\mathbf{R}$ \\
$c_t^i\in \{0,1\}$          & The binary presence of the $i$-th code in $\mathcal{C}$ in $\mathbf{v}_t$ \\ \bottomrule
\end{tabular}
\label{tab:Notation}
\end{table}

\subsection{Transformer Decoder Block}
The visit-level module described in the main paper makes use of a stack of Transformer Decoder blocks from the original Transformer paper \cite{Attention}. 
We provide additional details on those blocks here. Each block is defined mathematically by
\begin{align}\begin{split}
    \mathbf{H}^{(m)}_1 &= \mathbf{H}^{(m-1)} + \text{Masked Multi-Head Self-Attention}(\mathbf{H}^{(m-1)}) \\
    \mathbf{H}^{(m)}_2 &= \text{Layer Normalization}(\mathbf{H}^{(m)}_1) \\ 
    \mathbf{H}^{(m)}_3 &= (\mathbf{H}^{(m)}_2 + (\max(0,\mathbf{H}^{(m)}_2 \mathbf{W}^{(m)} + \mathbf{b}^{(m)}) \mathbf{V}^{(m)} + \mathbf{c}^{(m)})) \\
    \mathbf{H}^{(m)} &= \text{Layer Normalization}(\mathbf{H}^{(m)}_3)
\end{split}\end{align}
where Masked Multi-Head Self-Attention (MMSA) is then defined by
\begin{align}\begin{split}
    &\text{MMSA}(\mathbf{V}) = \text{Concat}(head_1, \cdots, head_h)\mathbf{W}^O \\
    &head_i = \text{Masked Attention}(\mathbf{V}\mathbf{W}_i^Q, \mathbf{V}\mathbf{W}_i^K, \mathbf{V}\mathbf{W}_i^V)\\
    &\text{Masked Attention}(\mathbf{Q}, \mathbf{K}, \mathbf{V}) = \text{softmax}\left(\frac{\mathbf{Q}\mathbf{K}^T}{\sqrt{d_k}} + \mathbf{M}\right)\mathbf{V}
\end{split}\end{align}
with $\mathbf{M}$ in the final line being a triangular matrix of $-\infty$ values ensuring that the softmax calculation only allows elements in the sequence to attend to themselves and elements before them. 

Layer Normalization is then a regularization technique that ensures that all neurons or variables in the corresponding intermediate layer have the same distribution across all features in a given input, which has been shown to produce smoother gradients and more effective training.

\subsection{Padding, Masking, and Truncating}
Each patient may have a different number of visits in their medical record. We first limit the records to a maximum of 96 visits as an arbitrary number which requires a small percentage of truncations and results in sequences that are just under 100 visits after adding the first visit, label visit, and end visit. From there, to support batch input to a variety of models, we need to maintain a uniform size. So, we pad each record $\mathbf{R}$ of real length $n$ to a constant number of columns, $N$, by appending the requisite number of empty vectors to $\mathbf{R}$. However, these padded visits as well as the initial start token visit do not contain any actual patient information and so should not be learned by any models. So, when calculating losses during training, both these ground truth visits and any outputs in these visits are zeroed out in order to ensure that they match and no loss arises from these meaningless visits.

\subsection{Chronic Diseases Labels}
Here we provide the specific list of chronic diseases we use to label patients in either dataset and also provide counts of how many patients possess each label.\\

\noindent The diseases for the outpatient dataset are:
\begin{itemize}
    \item Alzheimer 
    \item Kidney Disease 
    \item Heart Failure 
    \item Cancer 
    \item Depression 
    \item Arthritis 
    \item COPD 
    \item Stroke 
    \item Heart Disease 
    \item Diabetes 
    \item Osteoporosis\\
\end{itemize}

\noindent The diseases for the inpatient dataset are:
\begin{itemize}
    \item Renal Failure 
    \item Cerebrovascular Disease 
    \item Myocardial Infarction 
    \item Cardiac Dysrhythmias 
    \item Kidney Disease 
    \item COPD 
    \item Surgical Complications 
    \item Conduction Disorders 
    \item Congestive Heart Failure 
    \item Heart Disease 
    \item Diabetes With Complications 
    \item Diabetes Without Complications 
    \item Lipid Metabolism Disorders 
    \item Essential Hypertension 
    \item Fluid and Electrolyte Disorders 
    \item Gastrointestinal Hemorrhage 
    \item Hypertension with Complications 
    \item Other Liver Diseases 
    \item Other Lower Respiratory Diseases 
    \item Other Upper Respiratory Diseases 
    \item Pleurisy 
    \item Pneumonia 
    \item Respiratory Failure 
    \item Septicemia 
    \item Shock \\
\end{itemize}

\noindent The patient counts for the outpatient and inpatient datasets can be found in Supplementary Table \ref{table:OutpatientCounts} and \ref{table:InpatientCounts} respectively.

\begin{table*}[]
\smaller
\centering
\begin{tabular}{c|ccccccccccc}
\toprule
 & \footnotesize Alzheimer & \footnotesize Kidney Disease & \footnotesize Heart Failure & \footnotesize Cancer & \footnotesize Depression & \footnotesize Arthritis & \footnotesize COPD & \footnotesize Stroke & \footnotesize Heart Disease & \footnotesize Diabetes & \footnotesize Osteoporosis       \\ \midrule
Patients  & 6,325 & 30,357 & 50,242 & 83,117 & 117,839 & 1,997 & 105,547 & 58,305 & 23,926 & 155,463 & 23,325  \\
\bottomrule
\end{tabular}
\caption{Patient counts by disease label for the outpatient EHR dataset}
\label{table:OutpatientCounts}
\end{table*}

\setlength{\tabcolsep}{2.1pt}
\begin{table*}[]
\footnotesize
\centering
\resizebox{0.95\textwidth}{!}{
\begin{tabular}{c|ccccccccccccccccccccccccc}
\toprule
 & \rot{Renal Failure} & \rot{Cerebrovascular Disease} & \rot{Myocardial Infarction} & \rot{Cardiac Dysrhythmias} & \rot{Kidney Disease} & \rot{COPD} & \rot{Surgical Complications} & \rot{Conduction Disorders} & \rot{Congestive Heart Failure} & \rot{Heart Disease} & \rot{Diabetes With Complications} & \rot{Diabetes Without Complications} & \rot{Lipid Metabolism Disorders} & \rot{Essential Hypertension} & \rot{Fluid and Electrolyte Disorders} & \rot{Gastrointestinal Hemorrhage} & \rot{Hypertension with Complications} & \rot{Other Liver Diseases} & \rot{Other Lower Respiratory Diseases} & \rot{Other Upper Respiratory Diseases} & \rot{Pleurisy} & \rot{Pneumonia} & \rot{Respiratory Failure} & \rot{Septicemia} & \rot{Shock}       \\ \midrule
Patients  & 9,549 & 3,679 & 5,032 & 13,702 & 5,039 & 5,090 & 10,853 & 2,976 & 10,432 & 13,114 & 3,413 & 8,061 & 12,248 & 17,921 & 12,730 & 3,431 & 4,889 & 4,156 & 2,732 & 1,852 & 4,652 & 6,558 & 9,883 & 6,925 & 4,002  \\
\bottomrule
\end{tabular}}
\caption{Patient counts by disease label for the inpatient EHR dataset}
\label{table:InpatientCounts}
\end{table*}

\subsection{Lab Values}
In our evaluation of \method's ability to handle and produce continuous values, we explored including lab measurements from an expanded inpatient dataset. Here we provide the specific list of those labs:
\begin{itemize}
    \item Capillary Refill Rate
    \item Glascow Coma Scale Eye Opening
    \item Glascow Coma Scale Motor Response
    \item Glascow Coma Scale Verbal Response
    \item Glascow Coma Scale Total
    \item Diastolic Blood Pressure
    \item Systolic Blood Pressure
    \item Mean Blood Pressure
    \item Fraction Inspired Oxygen
    \item Glucose Percentage
    \item Heart Rate
    \item Height
    \item Weight
    \item Oxygen Saturation
    \item Respiratory Rate
    \item Temperature
    \item pH
\end{itemize}

\section{Supplementary Discussion}

\subsection{Code Orderings}
As mentioned in our main paper, we explore different orderings of code variables within a visit but find limited effect, so we settle on a random ordering in our experiments. However, we provide the full test set modeling (F1 Score and Perplexity) results here for four different orderings on the inpatient dataset. Note that these orderings do not affect any baselines as they all model each code independently rather than with any intra-visit modeling which could be affected by the order, so we only include results from \method here. We display results using our original ordering from the main paper, a second random ordering, alphanumeric ordering, and orderings low-to-high and high-to-low with respect to code prevalence in Supplementary Table \ref{table:OrderingResults}. There we see relatively little impact from any of the different orderings considered, with the random ordering and pseudo-random alphanumeric orderings performing the best, validating our decision to proceed with a random order in our experiments.

\begin{table}[]
\centering
\begin{tabular}{c|cc}
\toprule
Ordering & F1 Score & PP Per Code \\ \midrule
Main Paper (Random)        & 0.414 & 24.664 \\
Another Random             & 0.419 & 23.120 \\
Alphanumeric               & 0.417 & 23.680 \\
High-to-Low Prevalence     & 0.404 & 24.680 \\
Low-to-High Prevalance     & 0.400 & 27.243 \\ \bottomrule
\end{tabular}
\caption{The test set modeling results in terms of F1 Score and Perplexity achieved by \method in the context of our original and four new code variable orderings. Here we see very limited impact from the different orderings, with the random ordering and pseudo-random alphanumeric orderings performing the best, validating the decision to proceed with a random ordering within the main paper.}
\label{table:OrderingResults}
\end{table}

\subsection{Code Probabilities}
While we capture the unigram, sequential visit bigram, and co-occurrence bigram probabilities normalized at both the visit and record levels, we presented only those normalized at the record level (the probability of a given patient having that code or pair of codes) in our main text beyond a table of just the $R^2$ values for the visit-level due to the large size of the figure and the redundancy of further similar results. However, we feel that the visit level results are not only important but also offer insight into one of the key failure modes of the synthetic baselines and so provide them here. We show the visit level code probabilities in Supplementary Figure \ref{fig:codeProbs_visit}. There we see that at the visit level, each model is able to show strong correlation in the unigram evaluations, and the generally weaker performing baselines of SynTEG, EVA, and LSTM actually outperform the large language model baselines. SynTEG even improves upon \method's $R^2$ correlation by 0.001 with those two offering state of the art results in the evaluation. However, those three baselines then struggle and the gap between \method and all baselines widens in the sequential visit bigram probabilities. There SynTEG, EVA, and LSTM show basically no correlation while the GPT and \method $-$ Coarse baselines offer significantly less correlation than \method. This signifies a significant lack of temporal coherence in each of the baselines compared to \method. This incoherence is then a major failure mode and explains the highly sloped lines in many of the record-level code plots. Instead of repeating codes and maintaining consistency from one visit to the next over the course of a patient's medical history, many of those baselines produce individual visits which are reasonable but which are generated largely independently of another. This then causes the distinct number of codes and pairs of codes to be much larger than in real, temporally coherent patient data (and also explains why those steeply sloped plots are problematic and not easily fixed despite seemingly signaling correlation). The co-occurrence bigram results then largely mirror those in the record level plots albeit with slightly less correlation across the board due to the greater difficulty of the task. Again, \method easily outpaces each of the baselines with SynTEG and GPT's inter-visit modeling placing them second and third with all others lagging significantly behind. So, in looking at the visit level code probabilities we are able to not only reconfirm \method's state of the art performance but also gain insights into one of the major failure modes of the synthetic data baselines.

\begin{figure*}
    \centerline{\includegraphics[scale=0.59]{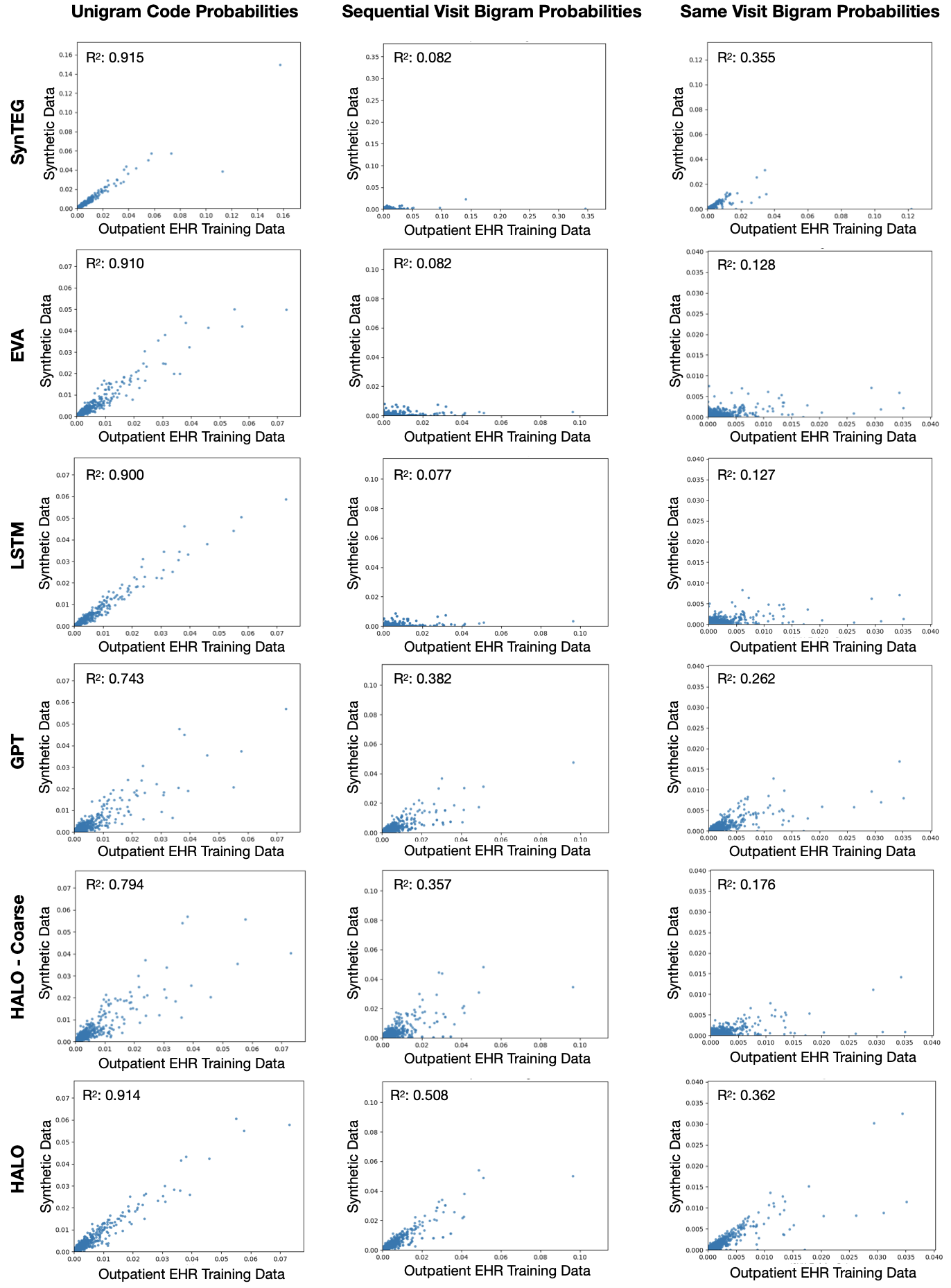}}
    \caption{Per Visit Unigram, Bigram, and Sequential Visit Bigram code probabilities for each synthetic dataset consisting of roughly 10,000, 1,500,000, and 5,000,000 points for each of the three types of plots respectively. These are analogous to the plots in the main paper but normalized at the probability of each code or pair of codes for individual visits rather than individual patients. Each model is able to show strong correlation in the unigram evaluations, and the generally weaker performing models of SynTEG, EVA, and LSTM perform very well. However, all of the baselines perform significantly worse in the sequential visit bigram probabilities, signaling their weaker temporal coherence which acts as a major failure mode for them. \method is able to largely avoid that problem and maintain not only its temporal coherence but achieve state of the art results in all settings.}
    \label{fig:codeProbs_visit}
\end{figure*}

\subsection{Low-Dimensional Setting}
To offer a more ready-made comparison to the settings that past works were proposed within and to demonstrate how baseline performance degrades upon moving to the high-dimensional setting that we examine, we repeat some of our experiments on the outpatient EHR dataset converted to a lower dimensional setting. Specifically, we aggregate codes into code phenotypes and remove any phenotypes which show up in our dataset less than 1000 times in the same way as \cite{SynTEG} did. This results in a low-dimensional version of our outpatient EHR dataset containing 1,349 distinct codes. We then train a new set of models and generate synthetic datasets for each of our compared methods of the same size as the training dataset. We extract a set of statistics and specifically examine a new set of per visit unigram, per visit bigram, and per visit sequential visit bigram code probabilities for each new low-dimensional synthetic dataset. 

We present the plots of these probabilities compared to those in the low-dimensional training dataset in Supplementary Figure \ref{fig:codeProbs_visit_lowDimensional}, and we compare the corresponding $R^2$ scores to those from the high dimensional setting in our main paper. We see that every model performs better and shows a stronger correlation for each probability than in the high-dimensional setting. The only exception to this universal improvement is SynTEG which performs slightly worse in the unigram and same-visit bigram code probabilities but with an extra boost in the more complex temporal patterns that it was specifically deficient in before as seen in its improved sequential visit bigram probabilities, chronic disease label probabilities, and record shapes, making for much better performance overall. In this low-dimensional setting, some of the weaker baselines are even able to approach or slightly surpass \method with respect to their per visit unigram code probabilities, demonstrating the strong performance that is showcased in their respective papers. However, those same baselines are then the ones that have the sharpest drop-off in performance when translating to the high-dimensional setting while \method, \method $-$ Coarse, and GPT are able to maintain stronger performance there. Furthermore, those weaker baselines, \method $-$ Coarse, and to a lesser degree GPT all perform significantly worse in both of the more complex and difficult bigram probabilities even at this lower dimensionality. \method is able to largely avoid that problem and maintain its performance there. So, \method achieves very strong unigram code probabilities and state of the art results in the two bigram probabilities by a wide margin in this low-dimensional setting as well.

\begin{figure*}
    \centerline{\includegraphics[scale=0.59]{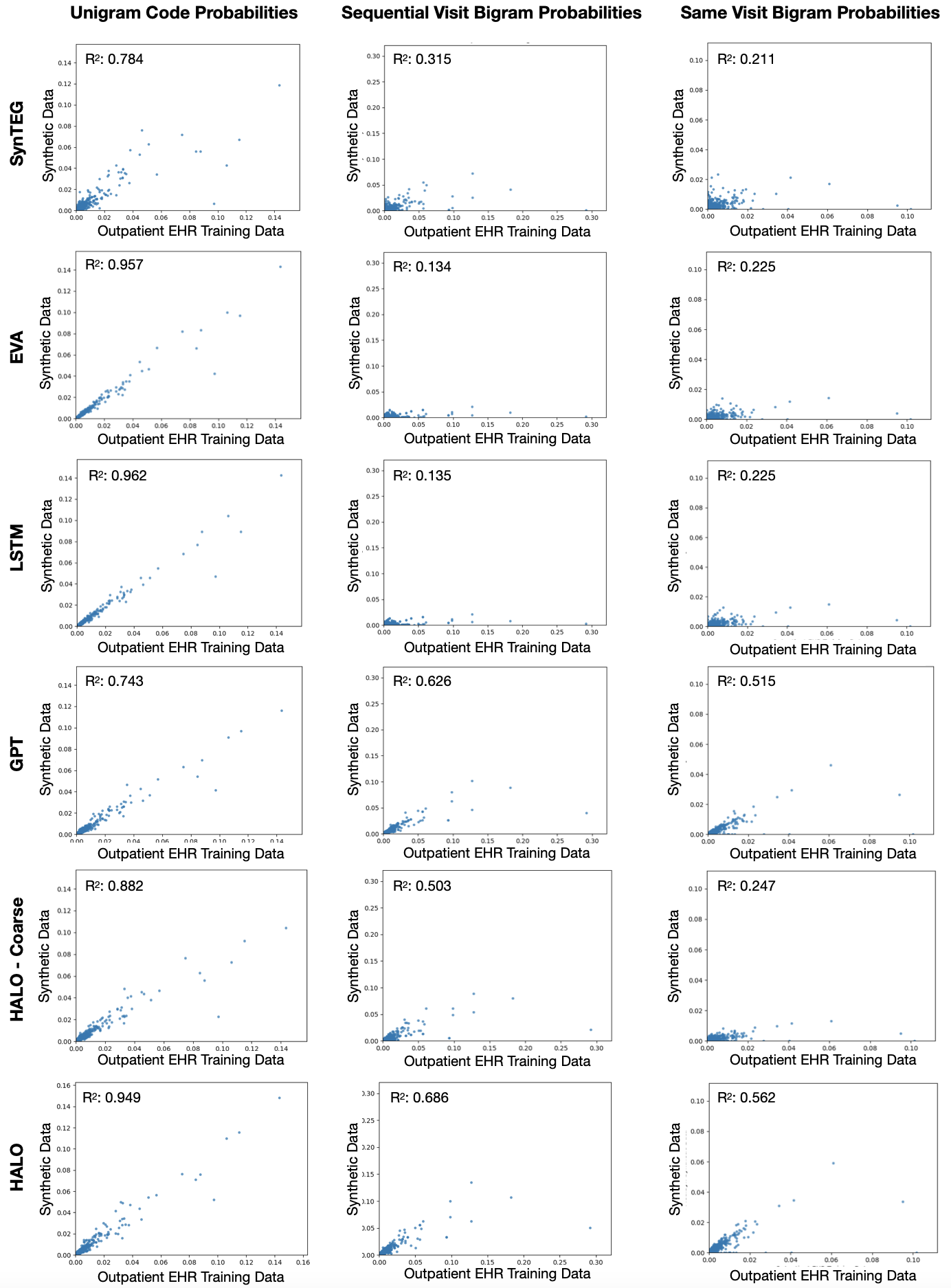}}
    \caption{\small Per Visit Unigram, Bigram, and Sequential Visit Bigram code probabilities for each synthetic dataset in the lower-dimensional setting consisting of just around 1,300 code phenotypes after code aggregation and rare code removal. These are the same plots as Supplementary Figure \ref{fig:codeProbs_visit} but in a low-dimensional setting to offer a comparison with the original setting of past works. Each model performs better and shows stronger correlation for each probability than in the high-dimensional setting. Some of the weaker baselines are even able to approach or slightly surpass \method with respect to their unigram code probabilities. However, all of the baselines perform significantly worse in both bigram probabilities even at this lower dimensionality. \method is able to largely avoid that problem and maintain its performance for those more complex probabilities. So, \method achieves very strong unigram code probabilities and state of the art results in the two bigram probabilities in this low-dimensional setting as well.}
    \label{fig:codeProbs_visit_lowDimensional}
\end{figure*}

\subsection{Comparison to ChatGPT}
While we focus on comparing \method to other leading baselines designed specifically for synthetic EHR generation, we feel that it is also worthwhile to compare its effectiveness to general purpose large language models such as OpenAI's ChatGPT. Such models have gained a lot of recent renown for offering state of the art performance in a wide variety of tasks. So, we explored ChatGPT's ability to generate patient records in a variety of formats and were impressed by its plausibility and ability to produce a variety of types of records (both in terms of conditional generation and output format). However, it is notable that despite this impressive ability, the generation process is designed to generate realistic responses overall rather than sample from the underlying distribution of patient records (let alone the distribution of records within a specific training dataset). This difference is then notable in preventing downstream usage for data analysis or machine learning tasks. To demonstrate the divide, we generate 1,000 patient records by feeding ChatGPT the prompt ``Generate a realistic patient record as a sequence of hospital visits with a set of ICD-9 codes at each visit in the form [[Code1, Code2, ...], [Code3, ...], ...]''. We then process and compare this synthetic dataset's statistics to those of our inpatient dataset. We show Per Visit code probabilities for both our \method and ChatGPT in Supplementary Figure \ref{fig:ChatGPTProbs}. While the fact that ChatGPT offers any correlation is an impressive feat, it is nonetheless clear that such language models can not be reasonably compared to synthetic EHR generation models such as \method for the types of downstream use cases that are desired.

\begin{figure*}
    \centerline{\includegraphics[scale=0.45]{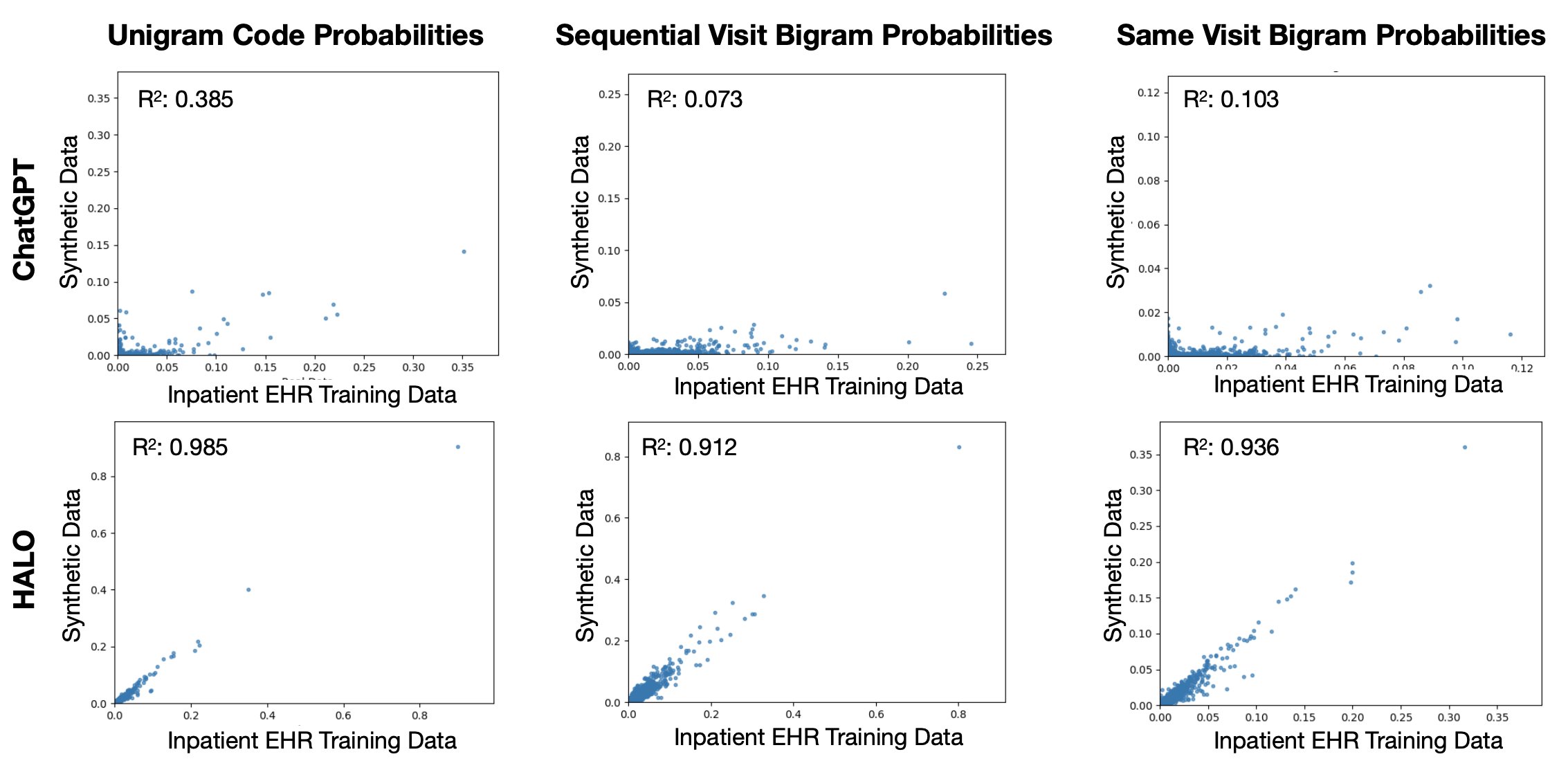}}
    \caption{Per Visit Unigram, Bigram, and Sequential Visit Bigram code probabilities for \method and ChatGPT's synthetic inpatient datasets. While ChatGPT's responses and data appear reasonable, they are not data-driven in terms of sampling from the underlying distribution of patient records and offer little correlation. \method is alternatively able to closely mirror the true distribution and statistics.}
    \label{fig:ChatGPTProbs}
\end{figure*}

\subsection{Record Shapes}
We presented aggregate statistics regarding mean number of visits per record and codes per visit for both the real and synthetic datasets in our main paper. We now present the more detailed corresponding probability density plots for each statistic in Supplementary Figure \ref{fig:ProbDensities}. There we see that \method and \method - Coarse outperform the other baselines to accurately mimic the training dataset's shape not just on average but throughout the distribution. 

We also present the chronic disease label probabilities for the outpatient dataset in Supplementary Figure \ref{fig:ConditionProbabilities}. There EVA and \method perform very well, effectively capturing the patterns of those codes found in the second, label visit.

\begin{figure*}
    \begin{subfigure}[b]{0.49\textwidth}
        \centerline{\includegraphics[scale=0.55]{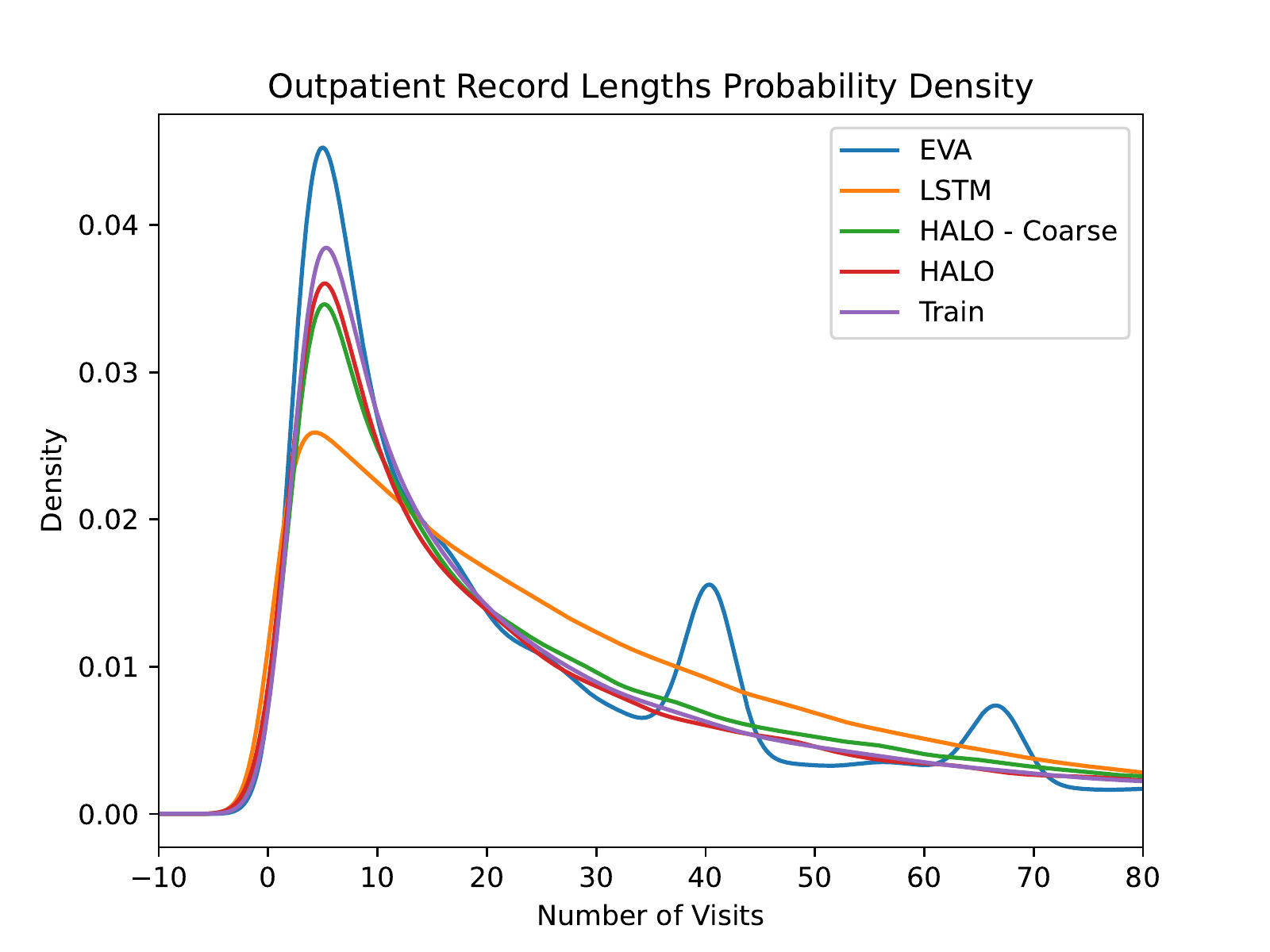}}
    \end{subfigure}
    \begin{subfigure}[b]{0.49\textwidth}
        \centerline{\includegraphics[scale=0.55]{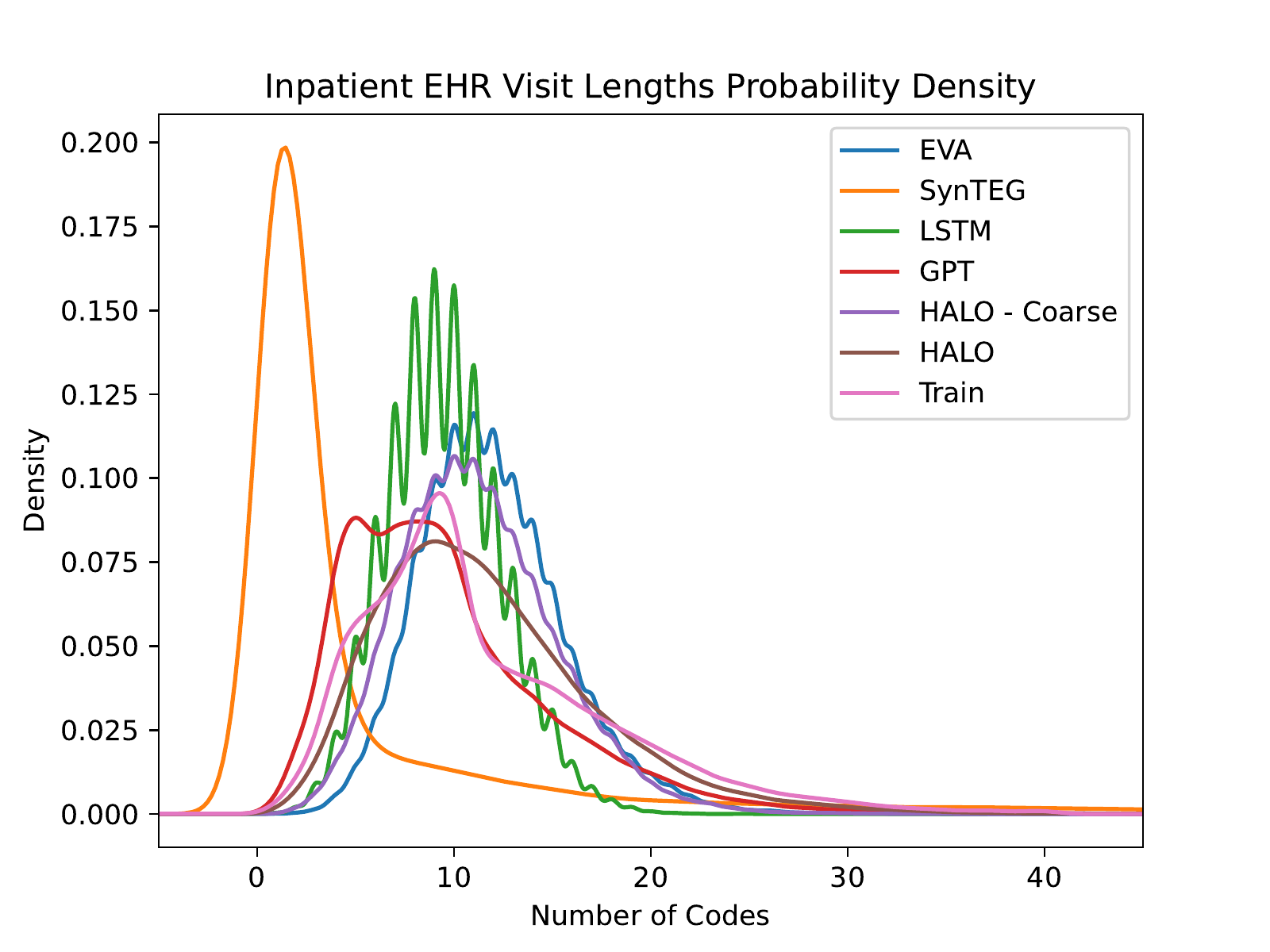}}
    \end{subfigure}
    \caption{Probability densities for the number of visits per patient record for the outpatient EHR dataset and corresponding synthetic datasets and the number of codes per visit for the inpatient EHR dataset and corresponding synthetic datasets respectively. We see that the language model architectures such as \method and \method $-$ Coarse outperform the other baselines to accurately mimic the training data distribution.
    }
    \label{fig:ProbDensities}
\end{figure*}

\begin{figure}
    \centerline{\includegraphics[scale=0.37]{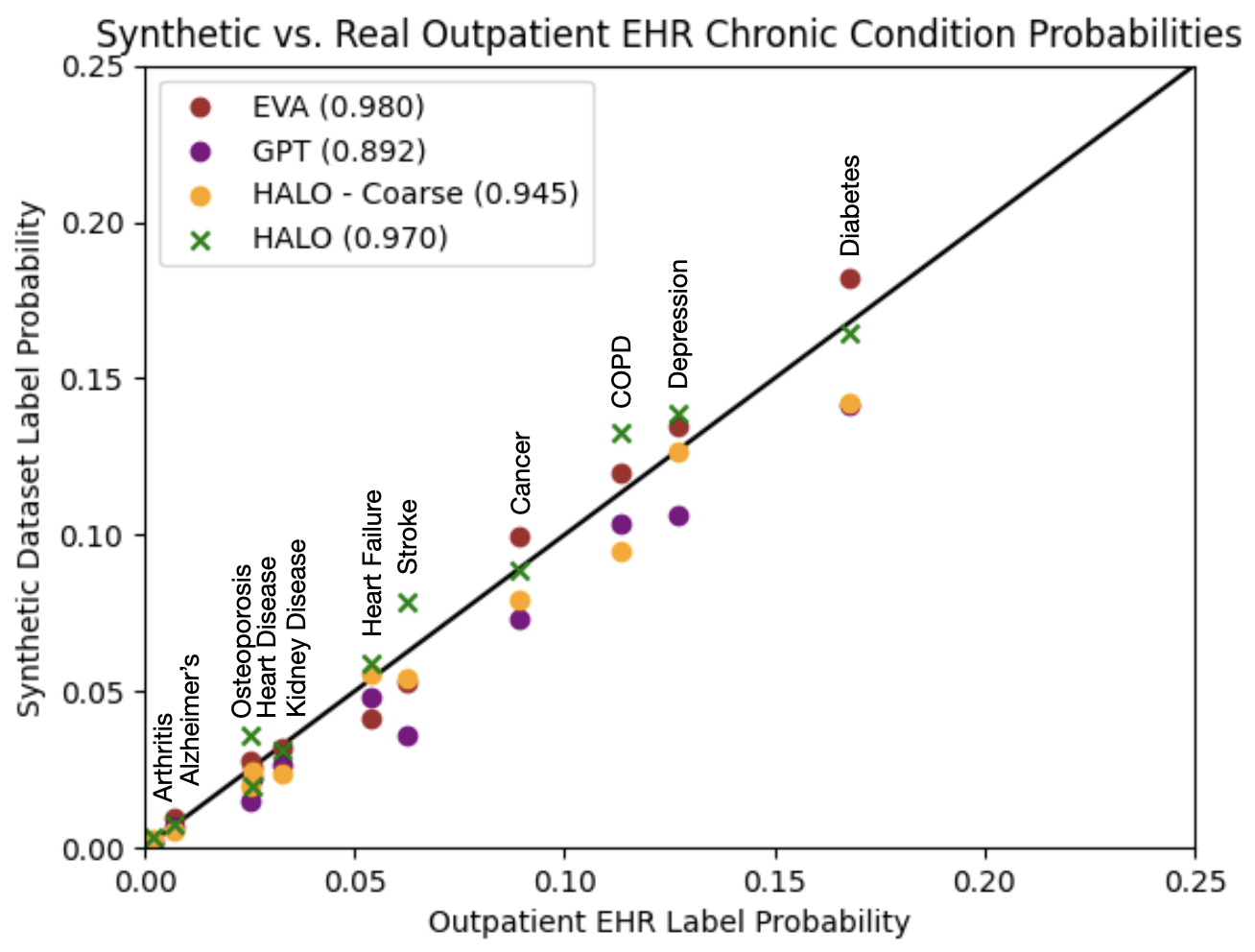}}
    \caption{We plotted the probabilities of each chronic disease label in the original outpatient EHR training dataset against their corresponding probabilities in each synthetic dataset. The $R^2$ value is shown in parentheses in the legend. The SynTEG and LSTM baselines both struggle with temporal consistency as manifested through their weak ability to create these chronic disease labels in the label visit, so they are omitted from the plot. In contrast, the EVA, \method $-$ Coarse, and \method architectures all closely mirror the training data with \method and EVA performing the best overall on average.}
    \label{fig:ConditionProbabilities}
\end{figure}

\subsection{Synthetic Training}
Finally, we presented aggregated results across the chronic disease labels for both the outpatient and inpatient EHR dataset in our main paper as they are more concise and easier to understand. However, for the sake of completeness we also provide accuracy by individual label for models trained on each of our compared synthetic datasets and the real training dataset in Supplementary Table \ref{table:FullOutpatientSynthetic} for our outpatient EHR dataset and Supplementary Table \ref{table:FullInpatientSynthetic} for our inpatient EHR dataset.

\begin{table*}[]
\smaller
\centering
\begin{tabular}{c|ccccccccccc}
\toprule
 & \footnotesize Alzheimer & \footnotesize Kidney Disease & \footnotesize Heart Failure & \footnotesize Cancer & \footnotesize Depression & \footnotesize Arthritis & \footnotesize COPD & \footnotesize Stroke & \footnotesize Heart Disease & \footnotesize Diabetes & \footnotesize Osteoporosis       \\ \midrule
EVA  & 0.518 & 0.533 & 0.5 & 0.549 & 0.449 & 0.5 & 0.559 & 0.5 & 0.526 & 0.5 & 0.5  \\
SynTEG  & 0.494 & 0.516 & 0.580 & 0.498 & 0.438 & 0.500 & 0.572 & 0.460 & 0.512 & 0.500 & 0.502  \\
LSTM  & 0.574 & 0.492 & 0.499 & 0.508 & 0.445 & 0.475 & 0.5 & 0.499 & 0.499 & 0.5 & 0.500  \\
GPT  & 0.908 & 0.903 & 0.904 & 0.857 & 0.829 & 0.844 & 0.861 & 0.846 & 0.911 & 0.951 & 0.851  \\
\method $-$ Coarse  & 0.915 & 0.905 & 0.909 & 0.891 & \textbf{0.854} & \textbf{0.930} & 0.937 & 0.885 & 0.914 & 0.951 & 0.888  \\
\method  & \textbf{0.924} & \textbf{0.909} & \textbf{0.912} & \textbf{0.908} & 0.852 & 0.926 & \textbf{0.948} & \textbf{0.889} & \textbf{0.921} &\textbf{ 0.963} & \textbf{0.896}  \\ \hline
Real  & 0.953 & 0.936 & 0.940 & 0.931 & 0.901 & 0.952 & 0.945 & 0.924 & 0.928 & 0.966 & 0.922  \\
\bottomrule
\end{tabular}
\caption{Full accuracy results by compared method and chronic disease label for the outpatient EHR dataset}
\label{table:FullOutpatientSynthetic}
\end{table*}

\setlength{\tabcolsep}{2.1pt}
\begin{table*}[]
\smaller
\centering
\begin{tabular}{c|ccccccccccccccccccccccccc}
\toprule
 & \rot{Renal Failure} & \rot{Cerebrovascular Disease} & \rot{Myocardial Infarction} & \rot{Cardiac Dysrhythmias} & \rot{Kidney Disease} & \rot{COPD} & \rot{Surgical Complications} & \rot{Conduction Disorders} & \rot{Congestive Heart Failure} & \rot{Heart Disease} & \rot{Diabetes With Complications} & \rot{Diabetes Without Complications} & \rot{Lipid Metabolism Disorders} & \rot{Essential Hypertension} & \rot{Fluid and Electrolyte Disorders} & \rot{Gastrointestinal Hemorrhage} & \rot{Hypertension with Complications} & \rot{Other Liver Diseases} & \rot{Other Lower Respiratory Diseases} & \rot{Other Upper Respiratory Diseases} & \rot{Pleurisy} & \rot{Pneumonia} & \rot{Respiratory Failure} & \rot{Septicemia} & \rot{Shock}       \\ \midrule
EVA  & 0.52 & 0.50 & 0.59 & 0.49 & 0.58 & 0.58 & 0.52 & 0.60 & 0.55 & 0.49 & 0.50 & 0.63 & 0.52 & 0.46 & 0.50 & 0.52 & 0.50 & 0.56 & 0.58 & 0.41 & 0.50 & 0.50 & 0.51 & 0.59 & 0.58  \\
SynTEG & 0.56 & 0.61 & 0.51 & 0.46 & 0.65 & 0.49 & 0.51 & 0.63 & 0.54 & 0.50 & 0.49 & 0.56 & 0.73 & 0.50 & 0.52 & 0.57 & 0.58 & 0.50 & 0.54 & 0.52 & 0.44 & 0.60 & 0.55 & 0.44 & 0.50  \\
LSTM  & 0.54 & 0.50 & 0.55 & 0.53 & 0.58 & 0.49 & 0.49 & 0.48 & 0.57 & 0.58 & 0.46 & 0.56 & 0.50 & 0.56 & 0.49 & 0.52 & 0.55 & 0.49 & 0.47 & 0.48 & 0.53 & 0.50 & 0.57 & 0.38 & 0.56  \\
GPT  & 0.90 & 0.86 & \textbf{0.91} & 0.90 & \textbf{0.93} & 0.88 & 0.85 & \textbf{0.83} & \textbf{0.95} & \textbf{0.93} & 0.87 & 0.89 & \textbf{0.93} & 0.91 & 0.83 & 0.84 & 0.90 & 0.85 & 0.72 & 0.73 & \textbf{0.86} & \textbf{0.87} & \textbf{0.86} & \textbf{0.92} & 0.88  \\
\method $-$ Coarse  & \textbf{0.91} & 0.86 & 0.81 & 0.90 & 0.87 & \textbf{0.89} & 0.84 & 0.82 & \textbf{0.95} & 0.90 & 0.87 & \textbf{0.91} & 0.86 & \textbf{0.94} & \textbf{0.84} & \textbf{0.86} & 0.89 & 0.86 & 0.69 & 0.73 & 0.85 & 0.86 & 0.81 & 0.91 & 0.83  \\
\method  & 0.90 & \textbf{0.89} & 0.90 & \textbf{0.91} & 0.91 & 0.88 & \textbf{0.87} & 0.82 & 0.92 & \textbf{0.93} & \textbf{0.89} & \textbf{0.91} & 0.92 & 0.93 & \textbf{0.84} & 0.84 & \textbf{0.93} & \textbf{0.87} & \textbf{0.74} & \textbf{0.80} & 0.85 & 0.85 & 0.85 & 0.90 & \textbf{0.89}  \\ \hline
Real  & 0.95 & 0.95 & 0.96 & 0.96 & 0.95 & 0.96 & 0.91 & 0.93 & 0.96 & 0.95 & 0.94 & 0.96 & 0.96 & 0.97 & 0.89 & 0.94 & 0.96 & 0.90 & 0.81 & 0.89 & 0.90 & 0.94 & 0.89 & 0.97 & 0.95  \\
\bottomrule
\end{tabular}
\caption{Full accuracy results by compared method and chronic disease label for the inpatient EHR dataset}
\label{table:FullInpatientSynthetic}
\end{table*}

We then also provide additional aggregated results on the outpatient data demonstrating the effect of using synthetic data as an augmentation technique to supplement real data. Specifically, we add additional models for each chronic disease classification task trained on the same real data but augmented with each synthetic datasets in turn as well. We show those aggregated results of mean test set classification performance across the 11 label-based tasks in Supplementary Table \ref{table:OutpatientSyntheticResults}. These results mirror those from the original setting of replacing real training data with synthetic data, with \method performing the best. However, it here offers the most gain over real-only training datasets rather than the least dropoff. Thus, we show that \method's synthetic data is able to be effectively used as an augmentation technique to produce better results than can be achieved with real data alone.

\begin{table}[]
\smaller
\centering
\begin{tabular}{c|ccc}
\toprule
& Avg. Accuracy  & Avg. F1 Score       & Avg. AUROC
\\ \midrule
EVA & 0.508 $\pm$ 0.02   & 0.283 $\pm$ 0.26   & 0.471 $\pm$ 0.08 \\
SynTEG  & 0.507 $\pm$ 0.03   & 0.514 $\pm$ 0.20   & 0.506 $\pm$ 0.07 \\
LSTM & 0.506 $\pm$ 0.02   & 0.467 $\pm$ 0.28   & 0.495 $\pm$ 0.06 \\
GPT & 0.851 $\pm$ 0.03   & 0.854 $\pm$ 0.03   & 0.914 $\pm$ 0.03 \\
\method $-$ Coarse & 0.867 $\pm$ 0.03   & 0.863 $\pm$ 0.03   & 0.920 $\pm$ 0.03 \\ 
\method & 0.879 $\pm$ 0.03  & 0.878 $\pm$ 0.03   & 0.938 $\pm$ 0.02 \\ \hline
Real Data & 0.891 $\pm$ 0.03   & 0.895 $\pm$ 0.03   & 0.943 $\pm$ 0.02 \\ \hline
EVA + Real & 0.844 $\pm$ 0.03   & 0.852 $\pm$ 0.03   & 0.921 $\pm$ 0.02 \\
SynTEG + Real & 0.846 $\pm$ 0.03   & 0.850 $\pm$ 0.02   & 0.915 $\pm$ 0.02 \\
LSTM + Real & 0.853 $\pm$ 0.02   & 0.857 $\pm$ 0.03   & 0.923 $\pm$ 0.02 \\
GPT + Real & 0.904 $\pm$ 0.02   & 0.906 $\pm$ 0.02   & 0.953 $\pm$ 0.01 \\
\method $-$ Coarse + Real & 0.910 $\pm$ 0.02   & 0.910 $\pm$ 0.02   & 0.958 $\pm$ 0.01 \\
\method + Real & \textbf{0.912 $\pm$ 0.02}   & \textbf{0.912 $\pm$ 0.02}   & \textbf{0.959 $\pm$ 0.02} \\ \bottomrule
\end{tabular}
\caption{Chronic disease classification models trained on different types of training data in the outpatient setting - real data, synthetic data generated by different methods, and real data augmented by synthetic data. Values are mean and standard error over 11 tasks. Bold values denote the best results. GPT, \method $-$ Coarse, and \method's synthetic data perform better than the other methods, and are comparable to using real data as training data. Augmenting real data with \method's synthetic data leads to better performance than just using real data. \method has the best results, with little drop-off in performance compared to real data and the largest gain when used to augment the training set.}
\label{table:OutpatientSyntheticResults}
\end{table}

\subsection{Additional Privacy Evaluations}
In our main paper we use a pair of membership inference attacks to evaluate the privacy preservation of \method and our other compared methods. Here we provide two other attacks from literature and show that \method thwarts them as well.

\noindent\textbf{Attribute Inference Attack:} The first of the two additional evaluations is the ability to thwart a typical attribute inference attack. This attack determines whether the synthetic dataset leaks specific and sensitive patient attributes based on correlations from demographic and other more common, less sensitive attributes of the patient. Consequently, it tests whether the synthetic dataset can be used to learn individual attributes of real patient data.

To demonstrate that \method is not susceptible to such an attack, we show that it thwarts the nearest neighbor-based attribute inference attack.  In this attack, we use subsets of the synthetic dataset and the original training dataset, randomly sampled to match the size of the original test dataset. We define demographic information, chronic disease labels, and the binary presence of the 500 most common medical codes (determined by the training dataset) as the conditional attributes. The sensitive attributes to be identified are the binary presence of the remaining uncommon medical codes.

To conduct the attack, we find the closest patient in the synthetic dataset for each patient in the training set based on having the most shared conditional attributes. We then predict each of the uncommon attributes to be the same as that closest synthetic patient. Those predicted attributes are compared with the ground truth sensitive patient attributes and graded using F1 Score. We then repeat this attack with real patients from the test dataset in place of the synthetic dataset and use the results as a baseline for acceptable attribute inference.

We show the results of the classifications from the nearest neighbor attacks in Supplementary Table \ref{table:attributeAttack}. There we see that not only are the prediction F1 Scores incredibly low on both datasets (4.7\% for the outpatient dataset and 3.3\% for the inpatient dataset), they are crucially lower than the baseline attack from the test set. This attack, labeled Real Data Attack in the table, sets the threshold for the amount of information revealed by the patterns of real data. So, staying below that level means incurring only an acceptable amount of attack success. So, we see that the synthetic dataset does not reveal any meaningful insight into the attributes of real patient data. We then see that each of the baseline synthetic datasets pass the test as well by having lower F1 Scores than the real data attack. GPT and \method $-$ Coarse allow similar F1 Scores to \method while all of the rest have much lower scores, likely because they do not capture the real patterns as effectively.\\

\begin{table}[]
\smaller
\centering
\begin{tabular}{c|c|c}
\toprule
& \multicolumn{1}{c|}{Outpatient EHR} & \multicolumn{1}{c}{Inpatient EHR} \\
& F1 Score        & F1 Score      \\ \midrule
Synthetic Data Attack   & 0.0397   & 0.0335 \\
Real Data Attack     & 0.0503    & 0.0473 \\ \hline
EVA     & 0.0108    & 0.0078 \\
SynTEG     & 0.0162    & 0.0094 \\
LSTM     & 0.0119    & 0.0068 \\
GPT     & 0.0447    & 0.0324  \\
\method $-$ Coarse     & 0.0330    & 0.0202 \\
\bottomrule
\end{tabular}
\caption{The results of the nearest neighbor attribute inference attack. The results showed that the F1 Score on both the inpatient and output datasets was below 0.05, and crucially lower than the baseline attacks using real data from the test set. This baseline attack sets the threshold for the amount of information revealed by the patterns of real data and so staying below it means incurring only an acceptable amount of attack success. This suggests that the synthetic dataset does not reveal any significant insights into the attributes of real patient data, and that \method is effective in preventing an attacker from inferring sensitive information. We then see that each of the baseline synthetic datasets pass the test as well by having lower F1 Scores than the real data attack. GPT and \method $-$ Coarse allow similar F1 Scores to \method while all of the rest have much lower scores, likely because they do not capture the real patterns as effectively. Source data are provided as a Source Data file.}
\label{table:attributeAttack}
\end{table}

\noindent\textbf{Nearest Neighbor Adversarial Accuracy Risk:}
The final evaluation, proposed in \cite{yale2020generation}, measures the degree to which a model overfits to its training dataset by looking at the relative likelihood of a patient's nearest neighbor being in the same or different datasets. As such, passing this test ensures that a generative model is generating wholly new synthetic patients rather than copying or performing simple augmentation on real training patients.

The evaluation is performed by calculating the metric Nearest Neighbor Adversarial Accuracy (NNAA). Let $S_T$, $S_S$, and $S_E$ be random subsets of $n$ records (we use $n$ = 5,000 records in our experiment) from the training, synthetic, and evaluation datasets respectively. NNAA risk is then the difference 
\begin{equation}
    AA_{ES} - AA_{TS}
\end{equation}
where
\begin{align}\begin{split}
    AA_{ES} &= \frac{1}{2}\left(\frac{1}{n}\sum_{i=1}^n 1\left(d_{ES}(i) > d_{EE}(i)\right) + \frac{1}{n}\sum_{i=1}^n 1\left(d_{SE}(i) > d_{SS}(i)\right)\right) \\
    AA_{TS} &= \frac{1}{2}\left(\frac{1}{n}\sum_{i=1}^n 1\left(d_{TS}(i) > d_{TT}(i)\right) + \frac{1}{n}\sum_{i=1}^n 1\left(d_{ST}(i) > d_{SS}(i)\right)\right)
\end{split}\end{align}
where the $E$ subscript throughout refers to the evaluation (test) dataset, $S$ refers to the synthetic dataset, and $T$ refers to the training dataset. $1(\cdot)$ is then the indicator function and $d_{ES}(i)$ is the distance from the $i$-th record in the evaluation dataset to its closest record (as determined by hamming distance in accordance with \cite{yan2022multifaceted}
) in the synthetic dataset. Each of $E$ and $S$ in $d_{ES}(i)$ can also be replaced interchangeably with any of $E$, $S$, and $T$, where the calculation just omits the record in question if the two datasets are the same. So, each $\frac{1}{n}\sum_{i=1}^n 1(d_{AB}(i) > d_{AA}(i))$ component is the probability of a record in dataset A being closer to another record in its own dataset than any record in dataset B. If they are randomly drawn from the same or similar distributions, we would expect that probability to be $\frac{1}{2}$, but it could be much lower if one of the datasets were copying from the other. We baseline this likelihood of the synthetic dataset copying from both its training and testing datasets, comparing the two to produce our overall risk.

\cite{yale2020generation} set 0.03 as the threshold for an acceptable NNAA risk. We show in Supplementary Table \ref{table:NearestNeighbor} that the NNAA values for both our inpatient and outpatient datasets are easily below that mark. Furthermore, we show that as more data is added as with the outpatient EHR dataset, the risk decreases to an even smaller value. So, we show that our \method method is not overfitting to or copying from its training dataset and instead is producing wholly new synthetic records. We repeat the evaluation with each of the baseline synthetic datasets and show that they pass as well. 

\begin{table}[]
\smaller
\centering
\begin{tabular}{c|c|c}
\toprule
Method & \multicolumn{1}{c|}{Outpatient NNAA} & \multicolumn{1}{c}{Inpatient NNAA} \\ \midrule
\method   & 0.0104   & 0.0211 \\ \hline
EVA     & 0.0040    & 0.0018     \\
SynTEG     & -0.0002    & -0.0080  \\
LSTM     & 0.0178    & 0.0082     \\
GPT     & 0.0045    & 0.0221    \\
\method $-$ Coarse     & 0.0047    & 0.0301  \\ 
\bottomrule
\end{tabular}
\caption{The results of the nearest neighbor adversarial accuracy risk evaluation. These values are calculated through the likelihood of data in the synthetic dataset being overly similar to records in the training set, normalized by their baseline likelihood of being close to unseen test set data. The metric was proposed in \cite{yale2020generation} where they set 0.03 as the acceptable risk threshold, a value that both the inpatient and outpatient synthetic datasets are well below. \method and other baselines all achieve much lower NNAA risk.}
\label{table:NearestNeighbor}
\end{table}

So, \method succeeds in passing both of our additional privacy evaluations, further reinforcing that its strong performance does not come at the expense of the privacy of the underlying patient records.

\bibliography{refs}